\documentclass[onecolumn]{robbyant}
\usepackage{tabularx}
\usepackage{overpic}
\usepackage[normalem]{ulem}
\newcolumntype{L}{>{\raggedright\arraybackslash}X}

\newcommand{\R}{\mathbb{R}}

\newcommand{\Loss}{\mathcal{L}}

\newcommand{\releaseTrainScenes}{475}
\newcommand{\releaseEvalScenes}{36}
\newcommand{\releaseScenes}{\the\numexpr\releaseTrainScenes+\releaseEvalScenes\relax}
\newcommand{\captureCollectionScenes}{1,550}
\newcommand{\goatTrainScenes}{145}
\ExplSyntaxOn
\newcommand{\trainingSceneRatio}{\fp_eval:n { round(\releaseTrainScenes / \goatTrainScenes, 1) }}
\ExplSyntaxOff

\newcommand{\releaseBenchmarkEpisodes}{360}
\newcommand{\releaseBenchmarkRequests}{1,497}
\newcommand{\relocationRequests}{1,105}
\newcommand{\goatMsgNavSubsetSR}{58.24}
\newcommand{\goatAstraSubsetSPL}{36.62}
\newcommand{\goatHarnessSubsetSR}{51.97}
\newcommand{\goatHarnessSubsetSPL}{39.28}
\newcommand{\goatHarnessArrivalSeconds}{16.7}
\newcommand{\goatHarnessTokensPerSubtask}{384,025}
\newcommand{\goatHarnessCostPerSubtask}{0.8}
\newcommand{\lingbotHarnessSubsetSR}{26.67}
\newcommand{\lingbotHarnessSubsetSPL}{17.74}
\newcommand{\goatGoalSR}{68.60}
\newcommand{\goatGoalSubsetSPL}{38.04}
\newcommand{\goatGoalSubsetSoftSPL}{33.02}
\newcommand{\goatGoalSubsetEpisodeSR}{22.22}
\newcommand{\lingbotGoalSR}{47.78}
\newcommand{\lingbotGoalFullSR}{47.76}
\newcommand{\goatMapTwoSubsetSR}{47.92}
\newcommand{\goatMapTwoSubsetSPL}{23.76}
\newcommand{\lingbotMapTwoSubsetSR}{26.39}
\newcommand{\lingbotMapTwoSubsetSPL}{14.46}
\newcommand{\goatAssistedMapTwoFullSR}{40.76}
\newcommand{\lingbotAssistedMapTwoFullSR}{26.28}
\metadata[Website]{\url{https://technology.robbyant.com/lingbot-navi}}
\metadata[GitHub]{\url{https://github.com/Robbyant/lingbot-navi}}
\metadata[Checkpoints]{\url{https://huggingface.co/robbyant/lingbot-navi}}
\newcommand{\methodname}{LingBot-Navi\xspace}
\newcommand{\method}{\texttt{\methodname}\xspace}
\newcommand{\datasetname}{\methodname}
\newcommand{\architectureName}{Scene-aware Navigation Transformer with Spatial Memory}
\newcommand{\reporttitle}{Scene-aware Navigation with Spatial Memory}

\newcommand{\bylinemark}[1]{\texorpdfstring{\textsuperscript{#1}}{}}
\newcommand{\bylinebreak}{\texorpdfstring{,\\}{, }}
\newcommand{\reportauthors}{%
    Huanhang~Hu, Xiao~Li, Lingyuan~Yang, Tong~Wang, Tianlong~Sun, Jie~Ding,
    Bo~Chen, Ge~Mu\bylinebreak
    Zhenhang~Guo, \href{https://ly-kc.github.io/}{Yuzheng~Liu},
    \href{https://pluswave.top/}{Jiawei~Wang},
    \href{https://scholar.google.com/citations?user=Vn4kwtcAAAAJ}{Zixuan~Wang},
    \href{https://scholar.google.com/citations?user=7rf6Bw4AAAAJ}{Zihan~Wang},
    Shizhe~Zang\bylinebreak
    Xing~Zhu, Yujun~Shen,
    Yangyan~Li\bylinemark{\textdagger}, Xinghe~Li\bylinemark{\textdaggerdbl}
}
\hypersetup{pdftitle={\reporttitle},pdfauthor={\reportauthors}}
\title{\reporttitle}

\author{%
\begin{center}
    Huanhang~Hu \quad
    Xiao~Li \quad
    Lingyuan~Yang \quad
    Tong~Wang \quad
    Tianlong~Sun \quad
    Jie~Ding \quad
    Bo~Chen \quad
    Ge~Mu
    \\[2pt]
    Zhenhang~Guo \quad
    Yuzheng~Liu \quad
    Jiawei~Wang \quad
    Zixuan~Wang \quad
    Zihan~Wang \quad
    Shizhe~Zang
    \\[2pt]
    Xing~Zhu \quad
    Yujun~Shen \quad
    Yangyan~Li$^{\dagger}$ \quad
    Xinghe~Li$^{\ddagger}$
    \\[12pt]
    $^{\dagger}$Corresponding Author \qquad
    $^{\ddagger}$Project Lead
\end{center}%
}

\begin{document}

\abstract{%
Memory is crucial for intelligent navigation systems, especially when a robot operates in the same environment multiple times.
Intuitively, a return visit should be considerably more efficient than the first, since the robot is expected to have become more familiar with its surroundings after the initial run.
To this end, we introduce a memory mechanism that encodes a scene as a point cloud of 3D visual tokens, each associating a spatial location with a semantics-aware feature descriptor.
Paired with this representation is a navigation system that operates in two regimes: (1) autonomous initial exploration guided by an occupancy map when no memory is yet available, and (2) \textbf{\textit{rapid retrieval, localization, and route planning}} by the \architectureName{} once the memory is in place.
The system additionally supports goal-directed exploration when retrieval finds no match and \textbf{\textit{online memory updates}} from continuously incoming observations.
It offers preliminary support for \textbf{\textit{disambiguation of user requests}} by describing differences between matching objects so the user can choose the intended instance.
Completing the system, which we term \method, a local planner executes the predicted route, keeping the full pipeline fast enough to run locally on edge compute.
To facilitate model training, we collect a dataset comprising \releaseScenes{} captured scenes across seven domains, covering over 103,000~m$^2$ of mapped free floor area, from which \releaseEvalScenes{} sites are held out to constitute a comprehensive evaluation benchmark.
Extensive evaluation on both the widely used GOAT-Bench and our self-constructed benchmark demonstrates the capability of our approach, which even outperforms a navigation agent powered by frontier models such as GPT-6 Astra.
The model, dataset, and benchmark are made publicly available to serve as a foundation for future studies on memory-augmented robot navigation.

}

\maketitle
\justifying

\section{Introduction}
\label{sec:introduction}

In his 1948 account of cognitive maps, Edward Tolman~\cite[pp.~195--196]{tolman1948}
described an experiment by Spence and Lippitt. Rats first visited both
arms of a maze while neither hungry nor thirsty. Tested later, hungry
rats tended to choose the arm where food had been and thirsty rats the
arm where water had been.
What they had learned about the maze served a need that arose later.
A robot that works in the same building day after day also has evidence
about what lies beyond its current view. Restricting each new request
to that view discards much of what the robot has already seen. In
\emph{scene-aware navigation}, past observations help determine both
where the user wants the robot to go and how to get there from its
present position.

A long line of work in vision-and-language navigation (VLN) studies how a
robot follows a route described in natural language~\cite{r2r,vlnce,duet,ivln,mapnav,navila}.
A visitor can follow turn-by-turn directions from someone who knows
the building. A person asking a robot to find an object may know
exactly what they want without knowing which room it is in. The robot
must then locate the object the user means, a task called
\emph{grounding}, and find its own route there. Several systems choose a destination
or a place to search and pass it to a separate path planner~\cite{msgnav,threedmem,scope}.
\method learns to ground the request and predict the route from the
same \emph{spatial memory} of its surroundings.

We represent this memory as a featurized point cloud. Each sampled surface
position, paired with a feature from a pretrained vision-language model
(VLM), forms a \emph{3D visual token}. This choice serves two purposes.
(i)~It supports open-vocabulary grounding, so a user can describe
a target in terms beyond a predefined list of object categories.
A scene graph records objects
and their relationships and can answer requests about the instances it
contains. Surface features can retain visual detail that nobody labeled
as an object or relation when the robot first saw it. Their 3D
positions let the model relate details seen from different viewpoints
within a common reconstruction. (ii)~We retain one token per occupied
3D cell (voxel) in each reconstruction, so seeing the same cell again
need not increase its token count. A frame archive grows with time spent in a room. A short frame window
can lose sight of rooms the robot visited earlier.

The \architectureName{} uses this memory to plan routes at the scale
of the scene (\cref{fig:architecture}).
The \emph{spatial memory encoder}
(a) encodes the appearance and 3D arrangement of the surface points.
The \emph{vision-language decoder} (b) uses those features to retrieve
objects matching a description or to localize an image by estimating
its camera's position and orientation in the remembered scene.
\textbf{Spatial memory makes navigation scene-aware by linking
the user's description to a remembered location, even when several
destinations are equally plausible from the current view alone.}
When several objects match, the decoder can describe differences
in their appearance or surroundings. This gives the user a way to
clarify which object they meant (\cref{fig:clarification_example}). An object's center
identifies the candidate but may lie inside it. The decoder therefore
pairs each center with a predicted \emph{observation viewpoint}, a pose
from which the robot might see the object on arrival. The robot's current
position and the selected viewpoint's position fix the two ends of the route.
The \emph{trajectory decoder} (c) reads the appearance and 3D arrangement
of remembered surfaces to infer where the path can pass between them.
Starting from a straight line between the endpoints, its flow-matching
head jointly refines the intermediate waypoints. A local planner uses
fresh camera views to follow the predicted route.

Errors in camera motion estimates accumulate over a long traversal.
Two views of the same surface can then appear to belong in different
cells. LingBot-Map~\cite{lingbotmap,chen2026reconsolidatable} reconstructs short sequences of color images. Each sequence forms a \emph{memory block} whose local geometry
is consistent enough to combine repeated observations. On a first visit,
an occupancy map of free space and obstacles guides the robot's autonomous
\emph{initial exploration} as it builds its first memory blocks (\cref{fig:system}). The robot
updates memory online as it travels. The transformer reads the blocks in
their own coordinate frames, so the robot can use new observations
without waiting for a globally aligned reconstruction. When the system selects blocks by coverage, it favors newer views of
revisited areas while retaining older blocks that cover other rooms.

When retrieval yields no match, the transformer uses the same memory
for \emph{goal-directed exploration}. Given the request, memory, and
current position, the vision-language decoder predicts where to search
next along the \emph{frontier} between observed free space and unexplored
space. The trajectory decoder predicts a route there. The robot adds
the new views to memory and tries retrieval again.

\begin{figure}[!htp]
\centering
\includegraphics[width=\linewidth]{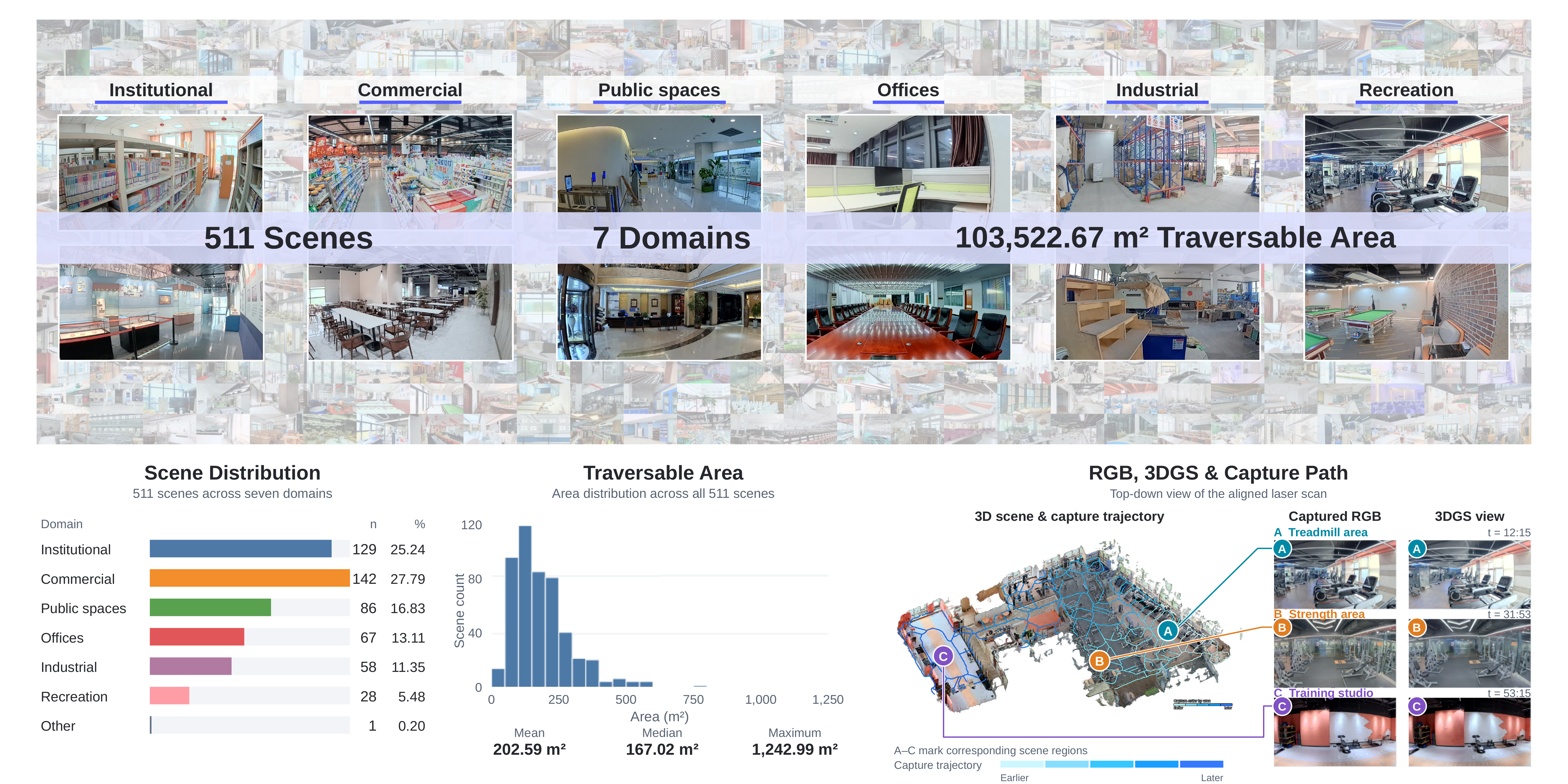}
\caption{\datasetname{} Dataset. The photographs at the top show examples
from six domains. Below, the domain counts and the distribution of mapped
free-floor area cover all \releaseScenes{} released scenes across seven
domains. At lower right, A--C link three regions of a gym scan to the
treadmill area, strength area, and training studio shown in the paired
views. Each row compares a held-out photograph with a 3D Gaussian
rendering from the same camera pose. Blue shading along the capture
path indicates elapsed time.}
\label{fig:dataset}
\end{figure}

GOAT-Bench~\cite{goatbench} tests whether a robot can use observations
from one request to answer the next. The \datasetname{} Dataset spans
\releaseScenes{} captured scenes across seven domains, from shops and
libraries to industrial facilities (\cref{fig:dataset}). Its
\releaseTrainScenes{} training scenes constitute the largest reported
training split among the successive-goal benchmarks in
\cref{tab:dataset_context} (\trainingSceneRatio{} times GOAT-Bench's count).
We release the remaining \releaseEvalScenes{} sites as a benchmark for
navigation across successive requests. Five commercial sites each have more than 500\,m$^2$ of
mapped free floor (\cref{sec:real_capture}). Across all eight systems
evaluated on both benchmark subsets, subtask success and success
weighted by path length are lower on \datasetname{}
(\cref{tab:standard_navigation}).

A robot can choose a good destination and still fail to reach it.
To distinguish these abilities, we evaluate goal prediction, where the
simulator plans and follows a path to the selected destination, and
action generation, where the navigator controls its own movement.
We test each with simulator geometry and with geometry reconstructed
from color images. All four configurations use the same four-billion-parameter navigator.
The simulator guides travel during the initial exploration pass.

With simulator depth, pose, and pathfinding, \method reaches
\goatGoalSR{}\% subtask success on the matched GOAT-Bench subset,
compared with \goatMsgNavSubsetSR{}\% for MSGNav~\cite{msgnav} with GPT-4o,
the strongest of six external goal predictors on this metric.
Among systems that generate actions with simulator depth, \method has higher
subtask success on both subsets than GPT-6 Astra~\cite{gpt6astra}
operating through NavHarness~\cite{navharness}, which scores higher when success
is weighted by path length. On \datasetname{}, their reported success rates are
32.36\% and \lingbotHarnessSubsetSR{}\%, respectively. The systems differ
in initial exploration and access to pose. \method receives simulator
pose, while NavHarness estimates it from color and depth. Our model runs locally. On GOAT-Bench the hosted agent averaged
\goatHarnessArrivalSeconds{}\,s per arrival check (\cref{sec:resource_profile}).

When \method uses reconstructed geometry and follows its own route,
it achieves \goatAssistedMapTwoFullSR{}\% subtask success on full GOAT-Bench
and \lingbotAssistedMapTwoFullSR{}\% on full \datasetname{} with LingBot-Map~2.0.
These scores are 4.15 and 2.35 percentage points higher, respectively,
than those obtained with Map~1.0. Both map versions use the same navigation
weights, trained with Map~1.0 reconstructions.

In this work, we introduce the \method model and the \datasetname{} Dataset.
The model uses remembered surfaces to locate the robot, ground the user's
request, and plan a route to the target. The benchmark asks the robot to use an earlier sighting to answer a
later request. Its relocation tasks test what happens when the robot
returns to the remembered location and finds the target has moved (\cref{tab:object_relocation}).

\section{Related Work}
\label{sec:related}

Prior work offers several ways to connect what a robot has seen to where
it should go. We examine how systems keep and query past observations, then how they
turn a selected destination into a route through the scene.

\subsection{Spatial memory across navigation tasks}
Room-to-Room~\cite{r2r} asks an agent to follow a route described in
words. In REVERIE~\cite{reverie}, the instruction refers to an object
the agent must reach and identify. With several requests in the same
scene, the path to one goal may already have brought a later target
into view. MultiON~\cite{multion} studies ordered object goals, and
iterative vision-and-language navigation~\cite{ivln} carries memory
across a tour of instruction-following episodes. GOAT~\cite{goat}
stores observations of individual object instances.
GOAT-Bench~\cite{goatbench} evaluates navigation over successive goals
specified by category, description, or reference image.
DynaMem~\cite{dynamem} studies how to update the scene memory when
later observations reveal changes in the environment. OpenIN~\cite{openin}
tests navigation to household objects that have moved since the initial
map was built. We test whether checking a remembered location helps the robot
respond when an object moves (\cref{sec:memory_eval}).

How a navigator stores its past observations shapes what it can retrieve
for the next request. Scene Memory
Transformer~\cite{scenememory} embeds successive observations for an
action policy to attend to. StreamVLN~\cite{streamvln} prunes visual
tokens using spatial information, while AstraNav-Memory~\cite{astranavmemory}
compresses them to retain a longer history. An agent can also retrieve
entire views. 3D-Mem~\cite{threedmem} selects snapshots from different
viewpoints. RAVEN~\cite{raven} indexes visual embeddings by place and time,
and Compass~1~\cite{compass1} uses selected views and a map of camera
positions and headings to return to previously observed targets.
In NavHarness~\cite{navharness}, GPT-6 Astra~\cite{gpt6astra}
directs its search using saved views, search notes, and a map built
from sensor observations. It chooses which area
to visit, issues movement commands, and requests a stop when it judges
that the goal has been reached.

A spatial representation also determines how a model connects a
description to a location. VLMaps~\cite{vlmaps} places vision-language
features on a map to locate language-specified goals and derive
obstacle maps for path planning. ConceptFusion~\cite{conceptfusion} fuses features at
corresponding pixels across views into a 3D map. Object-level graphs
can also support open-vocabulary queries. ConceptGraphs~\cite{conceptgraphs}
keeps visual embeddings and captions for segmented objects, together
with their relations. This is useful when a request identifies a mapped
object by its name or its relation to a landmark. MSGNav~\cite{msgnav}
uses a scene graph for navigation, and HOV-SG~\cite{hovsg} organizes
evidence by floor, room, and object. MapNav~\cite{mapnav} presents an
annotated top-down map to a VLM. In \method, visual features remain
addressable at surface locations even when the mapper has not
identified an object instance there. The same point features can
therefore support both matching an object's appearance to a phrase and
interpreting the surrounding layout when predicting a route.

When memory contains no match, the navigator must choose where to
search next. Existing approaches differ in how they select a frontier.
VLFM~\cite{vlfm} scores frontiers using the goal description.
MTU3D~\cite{mtu3d} scores both objects and frontiers,
SCOPE~\cite{scope} estimates the value of further exploration, and
HGR~\cite{hgr} revises hypotheses about where to search as new
observations arrive. When \method finds no match in memory, it uses
the request and the observed part of the scene to predict a frontier
for further search (\cref{sec:cold_start}).

\subsection{Spatial reasoning and conversational grounding}
A spatial question can call for a verbal answer even when the evidence
is geometric. SQA3D~\cite{sqa3d} studies spatial question answering
(QA) from a described situation in a scene. OpenEQA~\cite{openeqa}
evaluates answers grounded in recorded experience or active exploration.
LEO~\cite{leo} trains a shared agent for grounding, reasoning, and action
in 3D environments. Qwen-3D~\cite{qwen3d} connects a vision-language
backbone to scene tokens in world coordinates to answer questions and
predict object masks for grounding and instance segmentation. During navigation, a question about how
two objects differ can turn several plausible matches into choices
a user can recognize.

Navigation dialogue can resolve uncertainty about the route or about
the destination. Vision-and-Dialog Navigation~\cite{cvdn} studies exchanges in which a
human navigator asks a partner for route guidance. In
DialFRED~\cite{dialfred}, an agent can ask about a target's appearance or
location. SDA-LLM~\cite{sdallm} addresses ambiguous navigation requests
directly. It uses annotated camera views to describe how matching
objects differ in appearance or in their surroundings, then links the
user's choice to a location in a laser-scanned map. In \method, the
vision-language decoder describes the candidates using visual features
stored at the retrieved 3D locations. In our current interaction, the
user chooses among these descriptions by letter. The controller uses
the viewpoint paired with that object as the route goal.

\subsection{Reconstruction, route generation, and execution}
Reconstructed geometry makes observations from different viewpoints
comparable. Both versions of LingBot-Map~\cite{lingbotmap,chen2026reconsolidatable}
provide local geometry for our memory blocks. Our encoder adapts the attention pattern of the
Visual Geometry Grounded Transformer (VGGT)~\cite{vggt}. VGGT
alternates attention within and across images for 3D reconstruction.
We apply attention within and across blocks so that the navigation
decoders can read observations from several local coordinate frames.
The camera and block pose predictors draw on the multi-view geometry
model $\pi^3$~\cite{pi3}.

Navigation tasks differ in who chooses the route and who executes it.
For instruction following, DUET~\cite{duet} relates directions to a
growing map of connected places and to features of the current view.
Vision-and-Language Navigation in Continuous Environments~\cite{vlnce}
also requires the agent to execute movements along the instructed route.
When the user specifies a destination, the system must choose how to
reach it. In their released GOAT-Bench implementations,
MSGNav~\cite{msgnav} and 3D-Mem~\cite{threedmem} use the simulator's
pathfinder to reach a selected destination. Modular GOAT~\cite{goat}
instead plans on a map it builds from observations, and the
SenseAct-NN baselines~\cite{goatbench} predict movement actions.
NaVILA~\cite{navila} issues language-conditioned movement commands
to a locomotion policy.

Route prediction can also be viewed as generating a geometric shape
subject to constraints. Goren et al.~\cite{goren2026geometric} showed
that visual diffusion can approximate geometric constructions in 2D,
including Steiner trees that connect a given set of points. In maze
navigation, Diffuser~\cite{janner2022diffuser} conditioned trajectory
generation on the start and goal to produce paths around walls.
DualVLN~\cite{dualvln} and OmniNav~\cite{omninav} use flow matching
to generate local waypoints. We use flow matching for scene-level routes,
conditioning the waypoints on spatial memory.

\section{Approach}
\label{sec:system}\label{sec:model}

The navigation transformer reads the request, the robot's current view,
and spatial memory to choose a destination and predict a route there.
We first describe how it reads a memory whose surface points already
share a metric coordinate frame (\cref{sec:scene_planner}). We then
show how its encoder accepts short reconstructions with independent
frames (\cref{sec:memory_construction}). The controller selects and
updates these reconstructions across requests and aligns them with
odometry, the robot's estimate of its own motion (\cref{sec:cold_start}).
The local planner (\cref{sec:local_planner}) then uses fresh camera
views to follow the route and steer around nearby obstacles.

\subsection{\architectureName{}}
\label{sec:scene_planner}

\begin{figure}[!t]
\centering
\includegraphics[width=\linewidth]{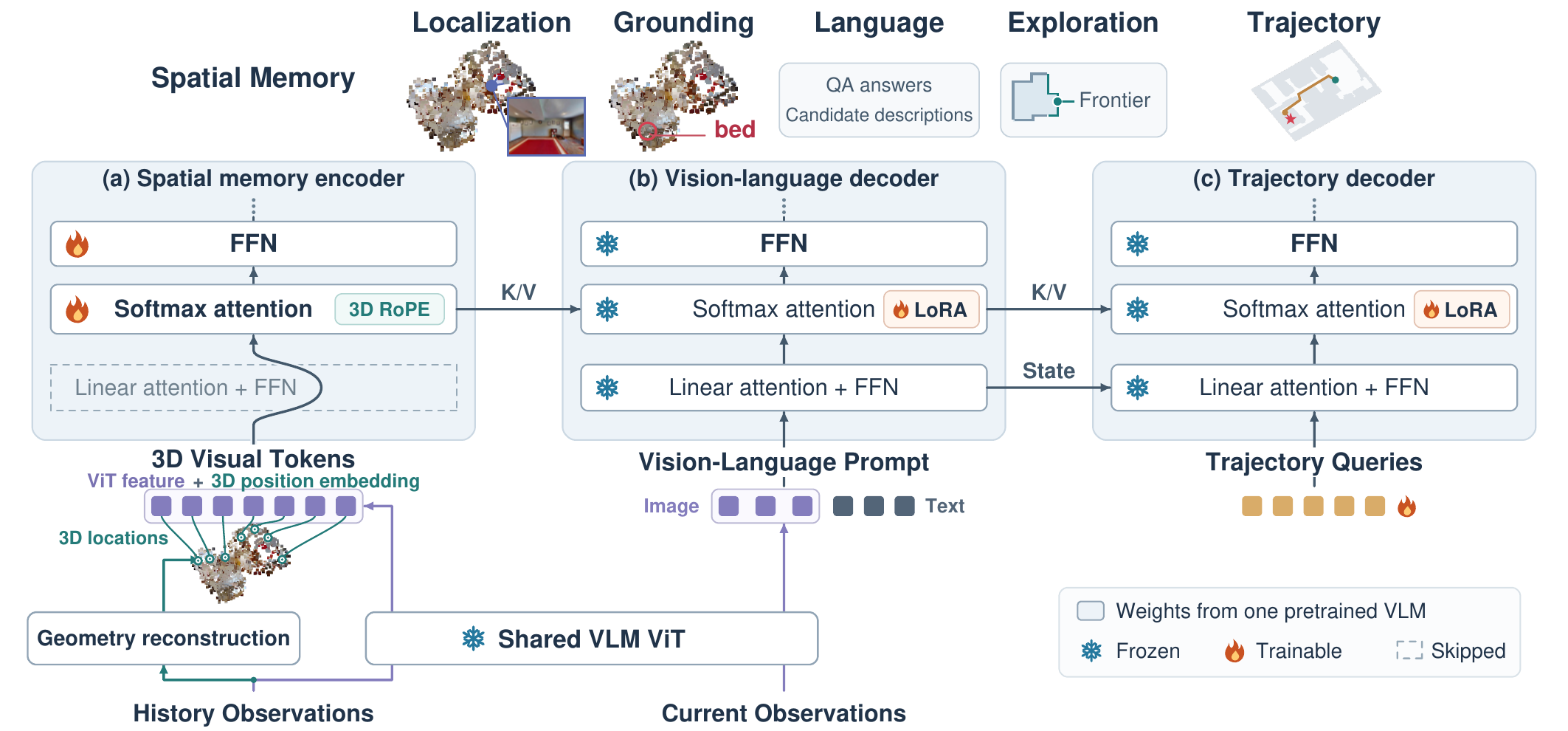}
\caption{The \architectureName{}. \textbf{(a) Spatial memory encoder.}
Teal lines connect reconstructed surface locations to patch features
from the shared frozen vision transformer (ViT). Each location supplies
a learned position embedding (\cref{eq:spatial_token_embedding}) and
the rotation used in 3D rotary position embedding (3D RoPE).
\textbf{(b) Vision-language decoder.} Image and text tokens attend to
the memory through its keys and values (K/V). The top row illustrates
image localization, grounding a bed, answering questions or describing
candidates, and predicting a frontier for an unresolved request. \textbf{(c) Trajectory decoder.}
At softmax-attention layers, learned queries read K/V from memory and
the prompt. At linear-attention layers, they inherit the prompt's
recurrent state. Their features condition the flow-matching head that
produces the route at the top right. Each panel shows one group of
backbone layers. Snowflakes and flames mark frozen and trainable
parameters. The dashed box in (a) marks the skipped layers.}
\label{fig:architecture}
\end{figure}

The three modules in \cref{fig:architecture} build on
\href{https://huggingface.co/Qwen/Qwen3.5-4B}{Qwen3.5-4B}, a
four-billion-parameter vision-language model (VLM) with publicly
available weights. Its frozen vision transformer (ViT) extracts patch
features from both the stored views and the current view. Reconstruction
assigns 3D surface positions to the stored features. Using the same ViT
for both inputs preserves the visual representation learned during
pretraining. The architecture can also use a different source of 3D
surface positions or a VLM that exposes patch features and supports
attention to memory.

The spatial memory encoder processes surface tokens, the vision-language
decoder processes the request and its images, and the trajectory decoder
uses learned queries to represent positions along the route. Following
the mixture-of-transformers design~\cite{mot}, each stream adapts
parameters to its inputs. We fine-tune copies of selected backbone layers
in the spatial memory encoder. The two decoders share frozen backbone
weights, each with its own low-rank adaptation (LoRA)~\cite{lora} of
the query, key, value, and output projections in softmax attention.
Related designs include the modality-specific feed-forward networks
(FFNs) in VLMo~\cite{vlmo} and
the separate action network conditioned on image and language tokens
in $\pi_0$~\cite{pi0}.

\paragraph{(a) Spatial memory encoder.}
For each retained patch $i$, the teal lines in \cref{fig:architecture}a
link its ViT feature $\mathbf f_i$ to the reconstructed 3D position
$\mathbf x_i\in\R^3$ of the surface visible in that patch. These position--feature
pairs form the point cloud. Its metric
coordinate frame $F$ is also the frame in which the model predicts
object centers and observation viewpoints, camera poses, exploration
frontiers, and route waypoints.

The encoder uses each position twice. First, it adds position information
to the visual feature itself. A learned position encoder $\phi_{\rm pos}$ takes the
coordinates and their Fourier features $\gamma(\mathbf x_i)$, which
evaluate sines and cosines at several spatial frequencies~\cite{fourfeat}.
Writing $[\cdot\,;\,\cdot]$ for concatenation, we obtain the spatial
memory token $\mathbf z_i$ for patch $i$ as

\begin{equation}
  \mathbf z_i = \mathbf f_i +
  \phi_{\rm pos}\!\left([\mathbf x_i\,;\,\gamma(\mathbf x_i)]\right).
  \label{eq:spatial_token_embedding}
\end{equation}

In our implementation, $\phi_{\rm pos}$ projects the raw coordinates with an
affine layer and the Fourier features with a two-layer perceptron. It
normalizes their sum and scales it with a learned scalar. Second, the
encoder uses the coordinates when two tokens attend to one another.
We adapt rotary position embedding (RoPE)~\cite{roformer} to metric 3D
coordinates. This 3D RoPE rotates the queries and keys using the same
spatial frequencies as $\gamma$. The relative rotation in their
attention score encodes the 3D displacement between the two surface points.

Surface tokens have no natural sequence order. The encoder uses
bidirectional softmax attention so every point can attend to the other
points, and it fine-tunes these layers together with their FFNs and
normalization weights. The backbone's linear-attention layers accumulate
a state in sequence order, so a different ordering of the same points
could change their representation. The curved arrow in panel~(a)
shows that the encoder skips these layers and their FFNs. The two
decoders retain them for their ordered inputs. At each stage, the
memory encoder projects the surface tokens into attention keys and
values (K/V) for both decoders. At inference, the system can reuse
these K/V for a new request or camera view because the encoder reads
only the memory. A change to the encoder's memory input requires it to
recompute the K/V.

\begin{table}[!t]
\centering\small
\caption{Tasks that share spatial memory. For coordinate navigation
and exploration, the output column follows the query sequence from the
vision-language decoder (b) to the trajectory decoder (c) in
\cref{fig:architecture}.}
\label{tab:task_interfaces}
\begin{tabularx}{\linewidth}{@{}p{3.3cm}LL@{}}
\toprule
Task & Query evidence in addition to memory & Model output\\
\midrule
Multi-object grounding & Category, description, or visible text to match &
Center and observation viewpoint for each match, or no match\\
\addlinespace[3pt]
View localization & Current or reference camera image &
Estimated camera position and orientation, or no match\\
\addlinespace[3pt]
Candidate choice & Request, lettered candidate centers, comparison prompt, and user's letter choice &
Candidate descriptions and the selected center--viewpoint pair\\
\addlinespace[3pt]
Spatial QA & Question about the scene & Text answer\\
\addlinespace[3pt]
Coordinate navigation & Current view and destination in the memory frame &
Estimated start pose and route. During navigation, aligned odometry can supply the start.\\
\addlinespace[3pt]
Goal-directed exploration & Unresolved request and current location in partial memory &
Frontier coordinate from (b), then a route to it from (c)\\
\bottomrule
\end{tabularx}
\end{table}

\paragraph{(b) Vision-language decoder.}
Image and text tokens read the spatial memory through the encoder's
keys and values. At softmax-attention layers, patches in the first
prompt image attend to one another bidirectionally. When a prompt
contains both a live view and a reference photograph, the live view
comes first. Later tokens, including patches in subsequent images,
use causal attention and can read their own and earlier positions.
The decoder uses multimodal rotary position embedding (MRoPE) to
encode a patch's frame, row, and column. Text uses the sequence index on each axis.

For object grounding (\cref{tab:task_interfaces}), the decoder emits
two special tokens for each match. Learned output layers, or
\emph{readouts}, map their features to a 3D object center and an
\emph{observation viewpoint}. The center identifies the instance but
may lie inside the object. The viewpoint gives a proposed position
from which to inspect it, together with the cosine and sine of a
horizontal heading. Appendix~\ref{app:training_batches} gives the losses for labeled viewpoints.

For localization, a pose readout estimates camera position and
orientation from an image. Estimates from live views help the controller
align odometry with spatial memory (\cref{sec:cold_start}). The pose
of a reference photograph supplies an approach point for an image goal.
For a spatial question, the decoder answers in text. When a target
cannot be retrieved, an exploration prompt combines the unresolved
request with the robot's current position. The decoder predicts a frontier coordinate in the memory frame,
which becomes the next destination for exploration.

\phantomsection\label{sec:dialogue}%
When several objects match, the interactive system assigns each candidate
a letter and supplies its center coordinates to the vision-language
decoder. A comparison prompt asks the decoder to describe differences
visible in spatial memory. The user can choose a letter from these
descriptions, as illustrated in \cref{fig:clarification_example}. The
choice identifies the object and its paired observation viewpoint for
planning. Appendix~\ref{app:retrieval_prompts} gives the retrieval and
clarification prompts.

\paragraph{(c) Trajectory decoder.}
Once the robot has identified a destination, the trajectory decoder plans
a route from the robot's estimated camera position. Let $\mathbf s$ and
$\mathbf g$ denote the start and destination, both 3D positions in the
memory frame $F$. For a selected object, $\mathbf g$ is the position of
its predicted observation viewpoint. During goal-directed exploration $\mathbf g$ is
the frontier predicted by panel~(b). A localized reference image or a
supplied coordinate can also provide the destination. The trajectory
decoder represents a route of $n$
segments by $n+1$ waypoints in meters. If $\widehat{\mathbf p}_k$
denotes the predicted position of waypoint $k$, indexed from start
($k=0$) to destination ($k=n$), the route is

\begin{equation}
 (\widehat{\mathbf p}_k)_{k=0}^{n}
 =(\mathbf s,\widehat{\mathbf p}_1,\ldots,\widehat{\mathbf p}_{n-1},\mathbf g).
 \label{eq:scene_planner}
\end{equation}

For training and evaluation, we pair each prediction with the reference
waypoint $\mathbf p^\star_k$ at fraction $k/n$ of the
reference route's arc length (Appendix~\ref{app:training_routes}).
Each interior index $k=1,\ldots,n-1$ has a trainable \emph{trajectory query}.
We use $n=79$, giving 78 queries and two fixed
endpoints. The queries follow the prompt at consecutive MRoPE positions.
A fixed sinusoidal encoding adds the route index to each learned query
embedding.

The arrows in \cref{fig:architecture} show how the trajectory queries
receive the earlier context. At softmax-attention layers, each query
reads the keys and values from the spatial memory and the encoded
prompt. The queries also attend to one another bidirectionally, so
earlier and later waypoints can inform one another. At linear-attention
layers, the State arrow carries the state accumulated while encoding the
prompt. The trajectory queries update that state in route order.

The interior waypoints must describe a single path through the scene.
For example, passing through a doorway constrains the approach to the
opening as well as the path beyond it. We refine the waypoints jointly
with a conditional flow-matching head~\cite{lipman2023flow}. The
trajectory decoder computes the features $H$ of all 78 queries once.
The head learns a velocity field $v_\theta$, with parameters $\theta$,
conditioned on $H$ and the endpoints (Appendix~\ref{app:training_batches}).

At inference, interior waypoint $k$ starts at
$\mathbf c_k=\mathbf s+(k/n)(\mathbf g-\mathbf s)$ on the straight
line from start to destination. We express its displacement from
$\mathbf c_k$ as a dimensionless offset, using a length scale of
$\ell=5$\,m. The matrix $U^{(r)}\in\R^{(n-1)\times 3}$ collects all
offsets after step $r$, with row $[U^{(r)}]_k$ for waypoint $k$.
Starting from $U^{(0)}=0$, we integrate the field over flow time
$t\in[0,1]$ in $S=16$ Euler steps. At step $r=0,\ldots,S-1$, we
evaluate the velocity at $t=r/S$. The update and final positions are
\begin{equation}
 U^{(r+1)} = U^{(r)} + \frac{1}{S}v_\theta\!\left(U^{(r)},r/S\mid H,\mathbf s,\mathbf g\right),
 \qquad
 \widehat{\mathbf p}_k = \mathbf c_k+\ell\,[U^{(S)}]_k .
 \label{eq:trajectory_flow}
\end{equation}
All $S$ updates reuse the same query features. The local planner follows
the resulting route (\cref{sec:local_planner}).

\begin{figure}[!t]
\centering
\includegraphics[width=\linewidth]{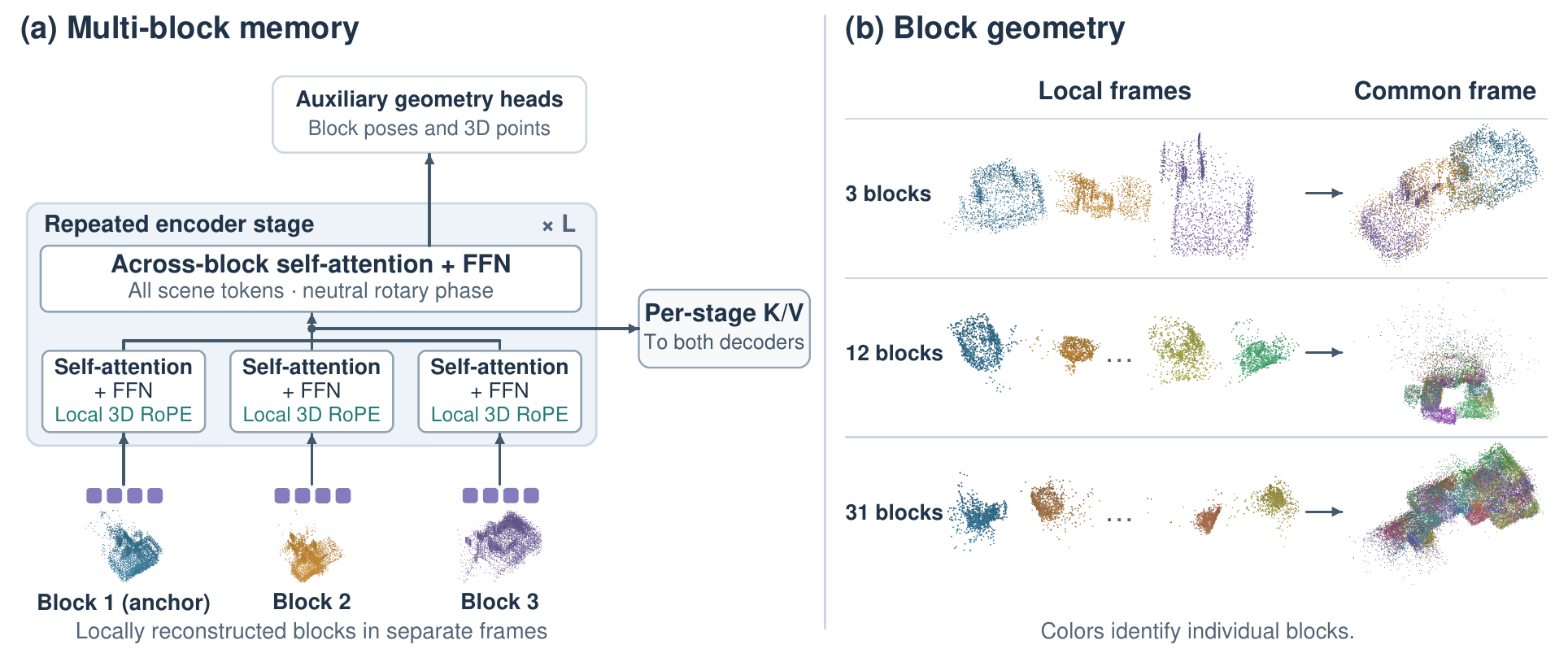}
\caption{Encoding short reconstructions with independent coordinate
frames. \textbf{(a) Multi-block memory.} Block~1 defines the reference
frame. Each of the $L=8$ stages applies attention within each block
using local 3D RoPE, then across blocks with a neutral rotary phase.
The across-block layer projects the within-block outputs into the K/V
that both task decoders read. Two auxiliary heads use the final point
features to predict block poses and surface positions in the reference
frame. \textbf{(b) Block geometry.} Each color identifies a block.
The rows show reconstructions with 3, 12, and 31 blocks in local frames
(left) and a common frame (right). For the larger reconstructions, the
left column shows selected blocks.}
\label{fig:memory_build}
\end{figure}

\subsection{Building and encoding spatial memory from local reconstructions}
\label{sec:memory_construction}

In a common metric frame, the encoder in \cref{fig:architecture}a
can relate any two observed surface points directly. Over a long
traversal, however, errors in estimated camera motion accumulate.
Two observations of the same surface can then occupy different voxels
and leave duplicate features in memory. We consolidate observations
within short reconstructions and keep each resulting \emph{memory block}
in its own coordinate frame. The encoder in \cref{fig:memory_build}
accepts these blocks before registration, the process of aligning their
coordinate frames. It supplies both decoders with spatial keys and
values in the same format as the encoder operating in a single coordinate frame.

\paragraph{Building locally consistent blocks.}
We reconstruct short sequences of color (RGB) images with
LingBot-Map~1.0~\cite{lingbotmap} or LingBot-Map~2.0~\cite{chen2026reconsolidatable}.
Version~1.0 supplied the reconstructed training memories. At inference,
the trained encoder accepts geometry from either version.
Both backends process frames as they arrive. Each block takes up to
250 new frames. To initialize the next block, the mapper can replay up
to 20 recent frames and store their tokens only once. During navigation,
streaming reconstruction can also supply current-view depth for recovery
(\cref{sec:benchmark_comparisons}).

Within each block, we pair the shared ViT's features with reconstructed
surface points and retain one pair per occupied voxel. The points have
metric coordinates local to that block. The first block presented to the
encoder is the \emph{reference block}. Its coordinate frame $F_1$
serves as the output frame $F$ in \cref{sec:scene_planner}. In our
simulator runs, odometry provides metric scale for reconstruction, and
the controller maintains a transform between odometry and $F_1$
(\cref{sec:cold_start}). Appendix~\ref{app:training_memories} describes
how we calibrate the training segments using reference depth. A
simultaneous localization and mapping (SLAM) system could also supply
locally consistent surface positions. Registering those positions across
the scene would yield a spatial memory in a single coordinate frame.

\paragraph{Reading across blocks.}
Coordinate differences measure physical displacements within a block.
Across blocks, the same calculation mixes different origins and axes.
Following VGGT~\cite{vggt}, which alternates attention within and
across images, each of our eight stages applies bidirectional softmax
attention first within each block, then across all blocks
(\cref{fig:memory_build}a). We initialize the two layers from
separate copies of the corresponding pretrained VLM decoder layer and
fine-tune their attention, FFN, and normalization weights independently.
All blocks share each local layer.

Within-block attention uses metric 3D RoPE to encode displacements
between surface points. Across-block attention assigns every token the
rotary position $(0,0,0)$. This gives an identity rotation, allowing the
layer to compare the features computed within each block even when the
coordinate frames differ. A learned token marks the start of each
block, and a distinct token marks the reference block. With only one
block, both layers attend to the same tokens.

At each stage, the across-block layer projects the within-block outputs
into keys and values \emph{before} updating the scene tokens. Both task
decoders read these K/V, one entry per retained point and one per block
token. Two additional heads use the final point features to predict
geometry in $F_1$. A rigid-pose head based on $\pi^3$~\cite{pi3} pools
the features of each nonreference block to predict the rotation and
translation that place it in $F_1$. The reference transform is the
identity. \Cref{fig:memory_build}b illustrates local and common
frames, and Appendix~\ref{sec:block_alignment_cases} compares predicted
and reference block poses. The controller uses the predicted poses to
compare the areas covered by different blocks (\cref{sec:cold_start}).
The second head predicts individual surface locations in $F_1$, with
an auxiliary training loss wherever a reference point is available.

\subsection{Using and maintaining spatial memory across requests}
\label{sec:cold_start}
\begin{figure}[!t]
\centering
\begin{overpic}[width=\linewidth,rel=1056]{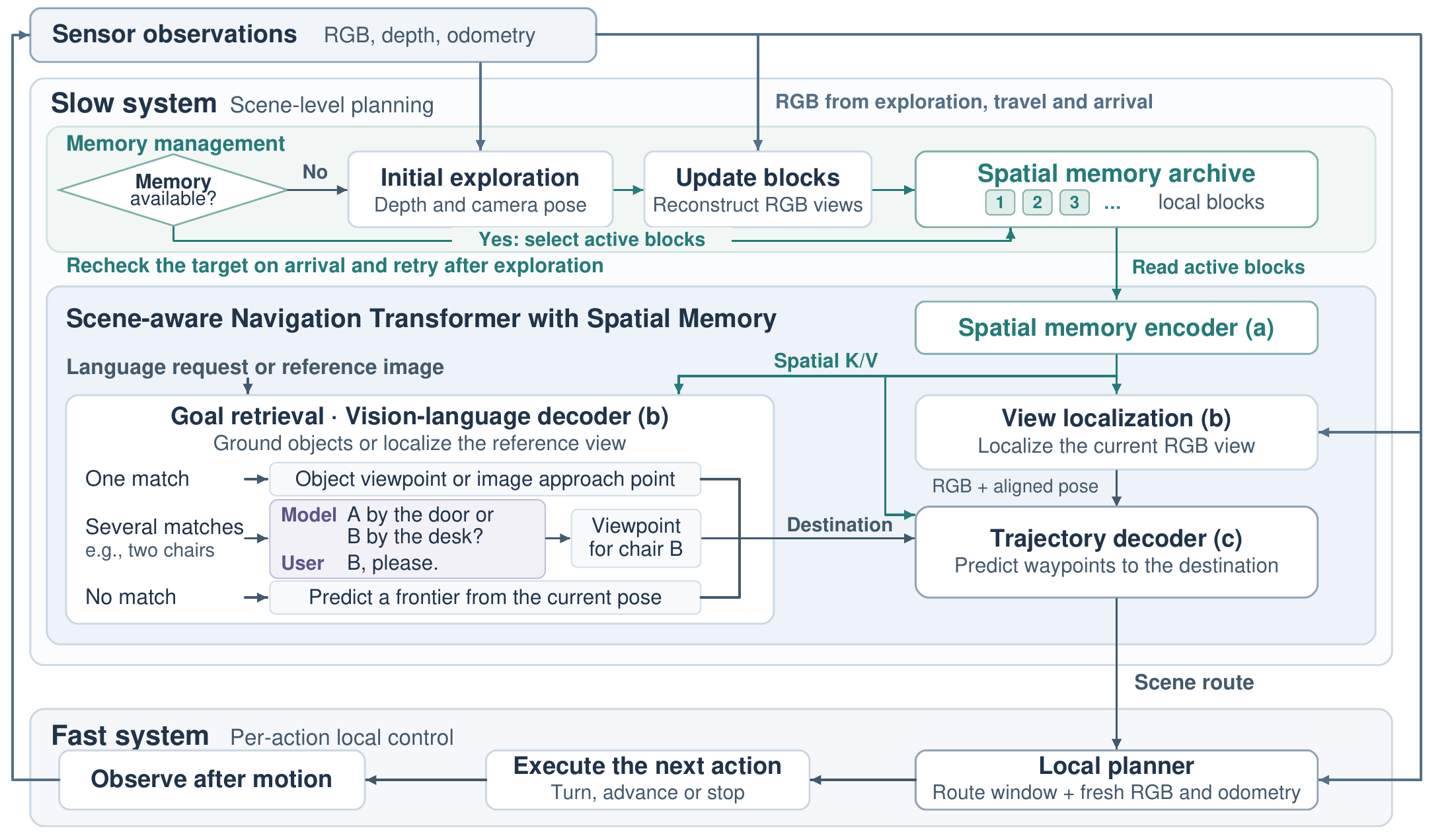}
  \put(577,372){\setlength{\fboxsep}{1.3pt}\sffamily\bfseries\scriptsize
    \hyperref[fig:architecture]{\colorbox[HTML]{455C72}{\color{white}Fig.~\ref*{fig:architecture}}}}
\end{overpic}
\caption{Navigation across requests. Initial exploration supplies the
first memory blocks. For each request, the slow system reads the active blocks.
Retrieval supplies the viewpoint of a matching object or an approach
point for a localized reference image. If retrieval finds no match,
the vision-language decoder predicts a frontier for goal-directed exploration.
When both chairs match in the illustrated exchange, the user selects
chair~B by the desk.
The trajectory decoder combines the destination with the current view
and aligned pose to predict a route. The fast system uses a window of
that route, fresh images, and odometry to choose the next action.
Observations from exploration, travel, and arrival feed back into
memory. In the relocation experiment, eligible targets can trigger the
pre-stop recheck in the slow system. Letters (a)--(c) identify the modules in \cref{fig:architecture}.}
\label{fig:system}
\end{figure}

The slow system in \cref{fig:system} selects a destination and predicts
a scene-level route whenever the robot needs a plan. Between plans,
the fast system chooses each movement from that route and fresh
observations. Each planning request therefore needs both a set of
memory blocks to read and the robot's current pose in their reference frame.

\paragraph{Initial exploration and memory selection.}
Before the first request, the robot builds its first memory blocks through
autonomous initial exploration guided by an occupancy map. It updates a
2D grid from metric depth and camera poses, then selects a reachable
frontier viewpoint by the area of unknown space it expects to reveal
per action. A reference pathfinder guides travel between viewpoints in
our benchmarks (\cref{sec:benchmark_comparisons}). Exploration ends when
the budget runs out or no reachable frontier remains. The resulting
views also serve to align odometry with memory.

The system keeps a chronological archive and adds observations from
travel to a target, from later exploration rounds, and, when needed,
before a retry. If the last block spans less than
1\,m of travel or contains fewer than ten frames, the next segment can
extend that block.

The blocks the controller presents for the current request form its
\emph{active memory}. For coverage-based selection, the controller uses predicted
relative poses to project each block's surface points onto a common
horizontal grid with 0.25\,m cells. It considers blocks from newest
to oldest and keeps a block if its previously uncovered cells add more
than 2\% of the full archive's footprint. This gives priority to the
newest view of a revisited area while preserving older blocks that
cover other parts of the scene. Among the selected blocks, the newest
with more than 30 frames becomes the reference. If none qualifies, the
controller uses the one with the most frames. It places the reference
first and retains it as new blocks arrive during the request.

GOAT-Bench uses coverage-based selection with simulator geometry and
with either LingBot-Map version. On \datasetname{}, the simulator-geometry
configuration also uses this rule. The two LingBot-Map configurations
read all available episode blocks, with the initial exploration blocks
first and the first of these as the localization reference.

\paragraph{Aligning odometry with spatial memory.}
The transformer predicts waypoints in $F_1$, and the robot tracks its
motion in the odometry frame $O$. The controller estimates the transform
between these frames from several views because a single view of a
blank wall or a repeated pattern can give an ambiguous pose. It selects
up to 24 near-horizontal views at different positions and headings
during initial exploration. On \datasetname{}, sampling spans
the entire exploration trajectory and favors views with reliable depth.
For each view the controller pairs the visual estimate of its camera
pose in memory with the pose reported by odometry. From mutually
consistent pairs it fits the transform $T_{F_1\leftarrow O}$, which
maps odometry coordinates into the memory frame by a rotation about the
vertical axis and a translation. A Huber loss limits the influence of
large residual errors in this fit.

Between reliable visual fixes, including when the current view is
poor, the controller uses odometry to estimate the robot's pose in
$F_1$. Consistent new visual estimates can correct the alignment. A
switch of reference block transfers the alignment using the predicted
relative block pose. The controller also updates the transform if a
merge shifts the current reference block's origin. The inverse transform
then places the planned waypoints in the odometry frame for execution.

\paragraph{Retrieving a goal and following the route.}
For a category or description, the vision-language decoder pairs each
matching object center with an observation viewpoint. In the exchange in
\cref{fig:system}, the system offers two labeled descriptions and the
user selects the chair by the desk (\cref{sec:dialogue}). In autonomous
benchmark runs, the controller instead ranks the retrieved centers and
retains the chosen center to identify the same object when replanning
(Appendix~\ref{sec:benchmark_scoring}). In either case, the trajectory
decoder receives the position of the paired observation viewpoint.
The executor checks traversability without enforcing the predicted
heading, so the robot can face a different direction at the stop. For a reference image, the controller derives the approach
point from the estimated camera pose.

If retrieval yields no match, the robot begins \emph{goal-directed
exploration}. The frontier query retains the original description or reference
image and the no-match response. Given the active memory and the
robot's current position in $F_1$, the vision-language decoder predicts
a frontier for that request (Appendix~\ref{app:route_prompts}). In
training, its target lies near the point where a path toward the goal
leaves observed free space in a partial memory
(Appendix~\ref{app:training_routes}). The trajectory decoder predicts
waypoints to the frontier, and the local planner follows the route
using fresh views (\cref{sec:local_planner}). Appendix~\ref{sec:exploration_cases}
compares routes to predicted and reference frontiers from the same
start. After each round, the controller adds usable views to memory
and retries retrieval. A match supplies a destination. A further miss
can trigger another round within the remaining budget.

With coverage-based selection, the controller also measures the area
newly covered by the archive after each exploration round. If two
consecutive rounds each add less than 1\,m$^2$, the controller withdraws
the oldest eligible block from the active input. Eligible blocks
predate the reference selected at the start of the request and have
received no new observations. The withdrawn block stays in the archive,
but a previously covered area can now become a frontier for the current
search. If no block is eligible, the controller tries retrieval once
more before ending the search.

\paragraph{Checking the destination.}
At a retrieved destination, the robot scans its surroundings and saves
the new observations. Older views can still place an object at its
former location after it has moved. In the \emph{relocation experiment},
where selected objects move between requests, our baseline permits an
arrival recheck for every target. Before stopping, the controller queries
the updated active memory. If it retrieves a center within 2\,m
horizontally of the robot's estimated position, it clears pre-request
surface points within the same horizontal radius of that center and
keeps the new observations.
It drops a block if fewer than 10\% of its original points remain, but
keeps the reference block intact if pruning would cross that threshold.
It then replans the same request once from the current position.
Retrieval from the revised memory can lead to a new approach or to
exploration. If the recheck yields no nearby center, the evaluator
scores the original stop. A variant with a mobility prior rechecks
only flagged targets. In both cases the evaluator retains the
ground-truth positions and move schedule. Appendix~\ref{sec:benchmark_scoring}
describes how actions count across attempts.

\subsection{Turning a route into actions with the local planner}
\label{sec:local_planner}
The local planner uses a short stretch of the predicted route and
recent camera views to choose movements around nearby obstacles. The
route carries the scene-level guidance from spatial memory. We train
the planner separately on routes from a frozen navigation transformer
and movements from reference trajectories (Appendix~\ref{app:training_batches}).
If progress stalls, the controller can save the new observations and
ask the navigation transformer for another route.

At each action, the controller finds the point on the predicted
route nearest the robot. It samples 16 waypoints along a short
stretch around that point, the \emph{route window}, and expresses them
in the robot's current frame. For each waypoint, a multilayer perceptron
embeds its 3D coordinates and their metric Fourier features, using the
same map $\gamma$ as in \cref{eq:spatial_token_embedding}. The input
also includes the arc length from the start of the window in meters
and the waypoint's fractional index from zero to one.
Even near a bend, these last two inputs distinguish the beginning of
the window from the end and preserve the intended direction of travel.

A frozen vision transformer encodes the current image, the image from
the last route update, and up to four earlier views selected as the
robot moves or turns. Each image patch carries its position in the
image, the frame's age, and indicators marking a current, route-update,
or historical view.

A four-layer transformer with hidden width 384 and eight attention
heads reads the visual and route tokens together with ten learned
\emph{action queries}. Each action query represents a future movement
step and carries an embedding of its index in this ten-step horizon.
Within each layer, visual and route tokens exchange information
bidirectionally. Action queries can read this context and one another,
while attention from the context to the queries is masked. The
three linear output heads use each query's features to predict a waypoint
in the robot's frame, scores for turning left, turning right, and moving
forward, and the probability of being at the
goal after that movement. The controller uses the first query's outputs
to choose the next action and then predicts again from the new
observation.

On reference trajectories, the local planner learns waypoint positions
with a smooth L1 loss $\Loss_{\rm waypoint}$ (transition distance 0.1\,m),
movement classes with cross-entropy $\Loss_{\rm action}$, and arrival at
the goal with binary cross-entropy $\Loss_{\rm arrival}$. We average
each loss over its valid targets and sum them with equal weight to give
the local training objective $\Loss_{\rm local}$,
\begin{equation}
  \Loss_{\rm local}=
  \Loss_{\rm waypoint}+\Loss_{\rm action}+\Loss_{\rm arrival}.
  \label{eq:local_loss}
\end{equation}
An arrival label is available only when the reference trajectory ends
at a known task goal (Appendix~\ref{app:training_batches}).

\section{The \datasetname{} Dataset and navigation experiments}
\label{sec:data}\label{sec:evaluation}

A robot may reach an object by recalling an earlier sighting or by searching
for it again. The successive requests in GOAT-Bench and \datasetname{} let
us study both behaviors. We introduce the captured scenes and how they
yield navigation tasks, then compare methods on the two benchmarks.
We examine how memory affects navigation, test individual model
predictions, and report inference costs and recurring failure modes.

\subsection{The \datasetname{} Dataset and benchmark tasks}
\label{sec:real_capture}

\begin{table}[!t]
\centering\small
\setlength{\tabcolsep}{3.5pt}
\caption{Navigation evaluations and their scene sources. HM3D denotes the
Habitat-Matterport 3D Dataset. Requests count individual goals or route
instructions. A sequence groups successive requests over which the agent
retains experience. A dash denotes a count that is inapplicable or
unreported for the indicated split. For SAGE-Bench and Habitat-GS we list the public navigation splits.
Wanderland evaluates pretrained policies across its full corpus. The
IR2R-CE and GOAT-Bench counts cover validation in unseen scenes. The
\datasetname{} row counts standard episodes and requests at the held-out sites.}
\label{tab:dataset_context}
\begin{tabularx}{\linewidth}{@{}l>{\raggedright\arraybackslash}p{.168\linewidth}rrrrL@{}}
\toprule
& & \multicolumn{1}{c}{Training} & \multicolumn{3}{c}{Evaluation} & \\
\cmidrule(lr){3-3}\cmidrule(lr){4-6}
Benchmark & Scene source & Scenes & Scenes & Sequences & Requests & Navigation protocol\\
\midrule
SAGE-Bench~\cite{sage3d} & InteriorGS (artist-designed) & 929 & 44 & --- & 944 &
\href{https://huggingface.co/datasets/spatialverse/SAGE-3D_VLN_Data}{High-level instructions} for 204 test routes.\\[3pt]
Habitat-GS~\cite{habitatgs} & InteriorGS and real scans & 110 & 19 & --- & 950 &
Instruction following on the \href{https://huggingface.co/datasets/RukawaY/gs_scenes}{VLN validation split}.\\[3pt]
Wanderland~\cite{wanderland} & Urban captures & --- & 530 & --- & --- &
Pretrained policies evaluated on point, image, or language goals.\\[3pt]
IR2R-CE~\cite{ivln} & Matterport3D & 60 & 11 & 36 & 1,824 &
The scene remains fixed throughout each tour of route instructions.\\[3pt]
OpenIN~\cite{openin} & Gibson & --- & 4 & --- & --- &
Objects move before navigation. Each sequence has 4--5 goals.\\[3pt]
FAROS-Bench~\cite{farosbench} & HM3D & --- & 7 & 21 & 210 &
A target moves twice and is requested at each placement.\\[3pt]
LiSoNav-Eval~\cite{lisonav} & HM3D & --- & 28 & 145 & 1,199 &
Objects can move between episodes within a sequence.\\[3pt]
GOAT-Bench~\cite{goatbench} & HM3D & \goatTrainScenes{} & 36 & 360 & 2,669 &
The scene remains fixed across category, description, and image goals.\\[3pt]
\textbf{\datasetname{}} & \datasetname{} & \releaseTrainScenes{} & \releaseEvalScenes{} & \releaseBenchmarkEpisodes & \releaseBenchmarkRequests &
Successive goals, with a separate relocation extension.\\
\bottomrule
\end{tabularx}
\end{table}

The \datasetname{} Dataset contains \releaseScenes{} of the \captureCollectionScenes{}
scenes we captured. The public split assigns \releaseTrainScenes{} scenes
to training and reserves \releaseEvalScenes{} distinct sites for evaluation. Each scene includes a fitted 3D
Gaussian splatting (3DGS) model for rendering color views, a mesh for
rendering depth, and an occupancy map that marks free space. An agent
can choose its own path through the reconstruction and build memory from
what it sees. The gym in \cref{fig:dataset} illustrates how views taken
far apart in time contribute to one reconstruction. The camera visited
the treadmill area (A) and the training studio (C) 41 minutes apart.
The scan connects these regions to pairs of held-out photographs and
rendered views.

Commercial venues (142) and institutional sites (129) are the two
largest release domains, alongside 67 offices and 58 industrial sites.
\Cref{fig:dataset} shows photographs from six domains alongside the
counts for all seven. \Cref{tab:capture_statistics}a also breaks
the counts down for the held-out sites.
SAGE-Bench~\cite{sage3d} builds instruction-following tasks on InteriorGS,
whose 1,000 artist-designed scenes include 752 residential interiors.
Habitat-GS~\cite{habitatgs} combines InteriorGS with real scans, and
Wanderland~\cite{wanderland} evaluates navigation in captured urban
environments. GOAT-Bench draws on the Habitat-Matterport 3D Dataset
(HM3D)~\cite{hm3d}, which contains 1,000 real scans of residential,
commercial, and civic buildings. Among the successive-goal benchmarks
with reported training counts in \cref{tab:dataset_context},
\datasetname{} has the largest training split by scene count,
\releaseTrainScenes{} versus \goatTrainScenes{} for GOAT-Bench.

The release covers 103,523\,m$^2$ of mapped free floor, with a median of
167\,m$^2$ per scene (\cref{fig:dataset}). We count free cells in
horizontal projection and multiply by the area of one cell.
The \releaseEvalScenes{} held-out sites cover 9,007\,m$^2$ before
allowing clearance for the robot's body. They comprise 11 commercial
venues, 11 public facilities, seven offices, four recreation sites,
and three institutional sites. Their median area is 196\,m$^2$ and
their 90th percentile is 596\,m$^2$. Thirteen exceed 250\,m$^2$, and
five commercial sites exceed 500\,m$^2$, including a shopping-centre
floor of 853\,m$^2$. \Cref{tab:capture_statistics}c gives the area
definitions used by other scene collections. The evaluation captures
contain 210,456 registered color views, a median of 5,785 per site,
including images reserved for checking the reconstructions
(\cref{tab:capture_statistics}b).

\paragraph{Successive navigation requests.}
GOAT-Bench~\cite{goatbench} issues a sequence of requests in one
scene. Each request is scored as a subtask, and the robot carries its
pose and memory forward into the next one. \datasetname{} adopts this
episode structure so a view acquired on one task can help with another.
GOAT-Bench evaluates category, description, and reference-image goals
across 36 scenes. The \datasetname{} benchmark uses object descriptions
for all \releaseBenchmarkRequests{} standard and \relocationRequests{}
relocation requests across its \releaseEvalScenes{} held-out sites
(\cref{tab:capture_statistics}).

Each \datasetname{} episode specifies a start pose, camera pitch, and
request order. Each target has a visibility mask, the set of free floor cells
that satisfy its distance and visibility criteria. An explicit stop inside the
mask succeeds. Oracle success credits any visit to the mask, even if the robot
later times out or stops elsewhere (Appendix~\ref{sec:benchmark_scoring}).

\paragraph{From mapping trajectories to supervision.}
For training, we render views along planned camera paths through the
reconstructed scenes. A single mapping trajectory can supply many
labeled requests because object positions are known relative to the
cameras and the navigable floor. \Cref{fig:asset_supervision_pipeline}a
follows a chair from a rendered view to a reviewed 3D instance.
Panel~(b) shows the resulting training examples. On held-out sites,
panel~(c) shows how the evaluator defines valid stops for a target
and updates them when the target moves.

To build training memories in \datasetname{} and HM3D, we reconstruct
short color sequences with LingBot-Map~1.0~\cite{lingbotmap} and attach
features from the frozen vision encoder to the surface points.
Grounding requests use the instances visible from recorded starts in a
region. Category, color, or material can distinguish an instance, and
a sign supplies a text target when its lettering is readable from the
selected memory. All labeled instances provide a center. Where the
annotations permit, they also provide an observation viewpoint with
a position and heading (Appendix~\ref{app:object_labels}). GOAT-Bench
descriptions add grounding examples, and ScanNet supplies spatial
question--answer pairs~\cite{scannet}.

Localization uses rendered starting views with known camera poses.
For route prediction, we sample reachable goals and generate reference
paths through the free space observed in the selected memory. Holding
out the blocks that cover a goal supplies examples of which frontier
to explore and how to reach it (Appendix~\ref{app:training_routes}).
In the training run documented in \cref{tab:data}, \datasetname{}
provided 53,855 eligible object instances, 7,396,283 candidate camera
poses shared by navigation and localization, and 139,663 exploration
starts. These counts describe the pool available before online
sampling. The local planner trains separately on reference trajectories
paired with routes from the frozen navigation transformer. A second
variant conditions on a language prompt and uses observations before
and after recovery maneuvers (Appendix~\ref{app:training_batches}).

\subsection{Navigation results on GOAT-Bench and \datasetname{}}
\label{sec:benchmark_comparisons}

\begin{table}[!t]
\centering\small
\setlength{\tabcolsep}{3pt}
\newcommand{\executionband}[2]{\multicolumn{9}{@{}l@{}}{%
  \makebox[0pt][l]{\colorbox{robbyfontbg}{\makebox[\dimexpr\linewidth-2\fboxsep\relax][l]{%
    \normalsize\sffamily\bfseries #1\quad{\small\normalfont #2}}}}}}
\newcommand{\geometryheading}[1]{\multicolumn{9}{@{\hspace{.65em}}l@{}}{\itshape #1}}
\caption{Navigation on successive requests. Compare methods within each group and split
using subtask success rate (SR) and success weighted by path length (SPL), both in percent.
The groups specify who executes the route and which geometry is available
after initial exploration. All \method runs use simulator-assisted initial exploration.
In the action-generation group with simulator inputs, SenseAct-NN receives pose,
NavHarness receives depth, and Modular GOAT and \method receive both.
All GOAT-Bench reruns use the released evaluator: the robot must stop strictly
within 0.25\,m of an annotated goal viewpoint by shortest navigable path.
\datasetname{} uses target visibility masks (Appendix~\ref{sec:benchmark_scoring}).
Each full split has 360 episodes. Our \datasetname{} subset uses the first episode
at each site (36 episodes, 156 requests). Our full-split scores come from earlier
runs. Episode-level pairing with external methods on the subset is unverified.
Bold marks the best reported score per group and column, a dash an unreported pair,
and an asterisk a published full-split result~\cite{goatbench,msgnav}.}
\label{tab:standard_navigation}
\begin{tabular*}{\linewidth}{@{\extracolsep{\fill}\hspace{1.6em}}lrr@{\hspace{9pt}}rr@{\hspace{15pt}}rr@{\hspace{9pt}}rr@{}}
\toprule
& \multicolumn{4}{c}{GOAT-Bench (Val-Unseen)} & \multicolumn{4}{c}{\datasetname{}}\\
\cmidrule(lr){2-5}\cmidrule(l){6-9}
& \multicolumn{2}{c}{Matched subset} & \multicolumn{2}{c}{Full split}
& \multicolumn{2}{c}{Subset} & \multicolumn{2}{c}{Full split}\\
\cmidrule(lr){2-3}\cmidrule(lr){4-5}\cmidrule(lr){6-7}\cmidrule(l){8-9}
Method & SR$\uparrow$ & SPL$\uparrow$ & SR$\uparrow$ & SPL$\uparrow$ &
SR$\uparrow$ & SPL$\uparrow$ & SR$\uparrow$ & SPL$\uparrow$\\
\midrule
\executionband{Goal prediction}{Simulator pathfinding to the selected goal}\\[2pt]
\geometryheading{Ground-truth depth and pose available}\\[1pt]
AstraNav-Memory~\cite{astranavmemory} & 46.02 & \goatAstraSubsetSPL{} & \multicolumn{2}{c}{---} & 13.75 & 9.51 & \multicolumn{2}{c}{---}\\
MTU3D~\cite{mtu3d} & 47.11 & 26.42 & \multicolumn{2}{c}{---} & 9.17 & 3.01 & \multicolumn{2}{c}{---}\\
MSGNav / GPT-4o~\cite{msgnav} & \goatMsgNavSubsetSR{} & 32.24 & 51.97\textsuperscript{*} & 29.56\textsuperscript{*} & 17.22 & 9.47 & \multicolumn{2}{c}{---}\\
SCOPE / GPT-4o~\cite{scope} & 28.09 & 18.12 & \multicolumn{2}{c}{---} & 11.67 & 7.19 & \multicolumn{2}{c}{---}\\
HGR / GPT-4o~\cite{hgr} & 30.38 & 20.70 & \multicolumn{2}{c}{---} & 11.11 & 6.68 & \multicolumn{2}{c}{---}\\
3D-Mem / GPT-4o~\cite{threedmem} & 28.01 & 18.31 & \multicolumn{2}{c}{---} & 10.56 & 5.93 & \multicolumn{2}{c}{---}\\
\addlinespace[2pt]
\method & \textbf{\goatGoalSR} & \textbf{\goatGoalSubsetSPL} & \textbf{63.71} & \textbf{36.18} & \textbf{\lingbotGoalSR} & \textbf{26.35} & \textbf{\lingbotGoalFullSR} & \textbf{27.58}\\
\cmidrule[.35pt](l{.65em}){1-9}
\geometryheading{Reconstructed geometry}\\[1pt]
\method\ (LingBot-Map 1.0) & 61.74 & 34.04 & 60.27 & 34.02 & 39.31 & 22.84 & 42.32 & 24.75\\
\method\ (LingBot-Map 2.0) & \textbf{67.93} & \textbf{38.84} & \textbf{61.28} & \textbf{35.20} & \textbf{46.25} & \textbf{26.24} & \textbf{45.78} & \textbf{26.99}\\
\midrule
\executionband{Action generation}{The method controls movement to the goal}\\[2pt]
\geometryheading{Ground-truth depth or pose}\\[1pt]
SenseAct-NN Monolithic~\cite{goatbench} & \multicolumn{2}{c}{---} & 12.3\textsuperscript{*} & 6.8\textsuperscript{*} & \multicolumn{2}{c}{---} & \multicolumn{2}{c}{---}\\
SenseAct-NN Skill Chain~\cite{goatbench} & \multicolumn{2}{c}{---} & 29.5\textsuperscript{*} & 11.3\textsuperscript{*} & \multicolumn{2}{c}{---} & \multicolumn{2}{c}{---}\\
Modular GOAT~\cite{goat,goatbench} & \multicolumn{2}{c}{---} & 24.9\textsuperscript{*} & 17.2\textsuperscript{*} & 8.47 & 5.49 & 10.03 & 6.16\\
GPT-6 Astra (NavHarness)~\cite{navharness} & \goatHarnessSubsetSR & \textbf{\goatHarnessSubsetSPL{}} & \multicolumn{2}{c}{---} & \lingbotHarnessSubsetSR & \textbf{\lingbotHarnessSubsetSPL{}} & \multicolumn{2}{c}{---}\\
\addlinespace[2pt]
\method & \textbf{56.09} & 28.98 & \textbf{52.92} & \textbf{26.98} & \textbf{32.36} & 17.06 & \textbf{30.65} & \textbf{15.73}\\
\cmidrule[.35pt](l{.65em}){1-9}
\geometryheading{Reconstructed geometry}\\[1pt]
\method\ (LingBot-Map 1.0) & 38.02 & 18.27 & 36.61 & 18.98 & 21.39 & 9.01 & 23.93 & 12.06\\
\method\ (LingBot-Map 2.0) & \textbf{\goatMapTwoSubsetSR{}} & \textbf{\goatMapTwoSubsetSPL{}} & \textbf{\goatAssistedMapTwoFullSR{}} & \textbf{20.87} & \textbf{\lingbotMapTwoSubsetSR{}} & \textbf{\lingbotMapTwoSubsetSPL{}} & \textbf{\lingbotAssistedMapTwoFullSR{}} & \textbf{14.03}\\
\bottomrule
\end{tabular*}
\end{table}

\Cref{tab:standard_navigation} groups methods by how they move and
which geometry they use. In \emph{goal prediction}, the method chooses
a destination and the simulator plans and follows a path there. In
\emph{action generation}, the method controls the robot's movements.
Within each mode, we distinguish systems with access to simulator depth
or pose from systems that use reconstructed geometry for their memory.
For \method, simulator depth and pose place observations in a common
frame. With LingBot-Map, images supply local geometry, simulator
odometry supplies metric scale, and a camera pose anchors the memory.

All six \method configurations share the same initial exploration policy
within each benchmark (\cref{sec:cold_start}). It builds the occupancy
map from simulator depth and camera poses. A pathfinder guides the robot
between selected viewpoints on Habitat's navigation mesh in GOAT-Bench
or the reference occupancy grid in \datasetname{}. The execution modes
in the table apply from the first request onward, including during
goal-directed exploration. On hardware, a color-and-depth (RGB-D) camera or calibrated stereo pair
could supply metric depth, with onboard odometry for
mapping~\cite{orbslamthree}.

Both benchmarks report subtask success rate (SR) and success weighted
by path length (SPL). SPL gives zero credit to a failed request and
less credit for success when the robot takes a longer path.

The stopping thresholds reported for GOAT-Bench refer to two different
locations. The original paper~\cite{goatbench} describes a stop within
1\,m of the target object in Euclidean distance. The authors' viewpoint
generator samples candidate positions within 1\,m of the object's
bounding box and checks navigability and visibility. The released
evaluator then measures the distance from the robot's stop to the nearest
annotated goal viewpoint along the shortest navigable path. Success
requires that distance to be strictly below 0.25\,m, the value also set
in the authors' original SenseAct-NN evaluation scripts. The published
NavHarness configuration~\cite{navharness} retains the viewpoint as
the reference but increases this stopping tolerance to 1\,m. We keep
the original released 0.25\,m rule for every GOAT-Bench rerun, including
GPT-6 Astra, so that all methods are scored against the same viewpoints
at the same tolerance (Appendix~\ref{sec:benchmark_scoring}).
On \datasetname{}, both motion settings use the target visibility masks
defined in \cref{sec:real_capture}.

\paragraph{Goal prediction.}
AstraNav-Memory~\cite{astranavmemory} and MTU3D~\cite{mtu3d} use pretrained
checkpoints. MSGNav~\cite{msgnav}, SCOPE~\cite{scope}, HGR~\cite{hgr}, and
3D-Mem~\cite{threedmem} query GPT-4o. These six methods supply search
or object destinations to simulator pathfinding. For \method, the
destination of a matched object is its predicted observation position.
If that point is unreachable, the controller can try an earlier
predicted waypoint. The simulator plans the path to either point.

We reran the six released implementations on the same 36 GOAT-Bench
episodes (278 requests), retaining each method's native camera and scan
policy and scoring the first 5,000 recorded actions per episode.
The common subset limits API costs. We average subtask scores within
each episode and give every episode equal weight. Among these six
methods, MSGNav has the highest SR (\goatMsgNavSubsetSR{}\%) and
AstraNav-Memory the highest SPL (\goatAstraSubsetSPL{}\%).
With simulator geometry, \method reaches \goatGoalSR{}\% SR and
38.04\% SPL. Its SR exceeds MSGNav's by 10.36 percentage points on
this subset. On full GOAT-Bench, Huang et al.~\cite{msgnav} report
51.97\% SR and 29.56\% SPL for MSGNav with GPT-4o.

On \datasetname{}, the external methods use fixed checkpoints or
APIs, with no training on the \datasetname{} split. Their subset SRs
range from 9.17\% to 17.22\% (MSGNav), and the best SPL is 9.51\%
(AstraNav-Memory). With simulator geometry,
\method reaches \lingbotGoalSR{}\% SR on the subset and
\lingbotGoalFullSR{}\% on the full split.
The full benchmark contains \releaseBenchmarkEpisodes{} episodes
(\releaseBenchmarkRequests{} requests). For all six \method configurations,
the standard subset uses the same 36 episodes and 156 requests, selected
from the v1.4 task lists (Appendix~\ref{sec:benchmark_scoring}).
The external goal predictors report averages over completed episodes
without the identifiers needed to pair their requests and stopping masks
with ours.

\paragraph{Action generation.}
The two SenseAct-NN policies~\cite{goatbench} use RGB with simulator
GPS and compass readings. Modular GOAT~\cite{goat} builds its object map
using simulator depth and pose. GPT-6 Astra selects actions through
NavHarness~\cite{navharness}. The model can inspect saved views and a map,
request a route preview, and issue forward or turning commands. NavHarness
builds the map from simulator depth and the camera poses estimated by
RGB-D ORB-SLAM3~\cite{orbslamthree}. In the same simulator-input group,
\method uses the supplied depth and pose for memory, predicts a route,
and follows it with its local planner.

On the GOAT-Bench subset, \method has higher SR than GPT-6 Astra
(56.09\% versus \goatHarnessSubsetSR{}\%), and GPT-6 Astra has higher
SPL (\goatHarnessSubsetSPL{}\% versus 28.98\%). The ordering is the
same on \datasetname{}, where \method achieves 32.36\% SR and 17.06\% SPL
and GPT-6 Astra reaches \lingbotHarnessSubsetSR{}\% SR and
\lingbotHarnessSubsetSPL{}\% SPL. Both systems report results for
36 episodes and 156 requests. The external results do not establish
request-by-request pairing, and the systems differ in how they obtain
camera poses and perform initial exploration.

Among the action-generating methods evaluated on the full
\datasetname{} split, Modular GOAT and \method both receive simulator
depth and pose. Modular GOAT reaches 10.03\% SR and 6.16\% SPL.
With the same sensor inputs, \method reaches 30.65\% SR and 15.73\% SPL.

Zhao et al.~\cite{navharness} report 83.7\% SR and 62.3\% SPL for
GPT-6 Astra in NavHarness on full GOAT-Bench, averaged over three
seeds with a 1\,m viewpoint threshold. Our GOAT-Bench rerun uses
both a smaller split and the released 0.25\,m threshold. On
\datasetname{} we can isolate the effect of tolerance by rescoring
the same 36 trajectories from our GPT-6 Astra run
(\cref{tab:navharness_tolerance}). SR rises from
\lingbotHarnessSubsetSR{}\% under the standard mask to 48.33\% when
stops up to 0.50\,m of traversable distance beyond the mask count,
and to 66.25\% at 1.00\,m. At 1.00\,m, 102 of 156 stops qualify,
compared with 41 inside the original masks. The locations of these
near misses leave open whether they arose in choosing a viewpoint,
following the route, or deciding when to stop.

\paragraph{Geometry and execution ablations.}
The difference between \method's geometry sources depends on how the
robot moves. With simulator pathfinding, SR with LingBot-Map~2.0 memory is
2.43 percentage points lower than with simulator geometry on full
GOAT-Bench and 1.98 points lower on full \datasetname{}. With the
local planner following the predicted route, the gaps are 12.16 and
4.37 points. The reconstructed-memory runs reach
\goatAssistedMapTwoFullSR{}\% and \lingbotAssistedMapTwoFullSR{}\% SR,
retaining 77\% and 86\% of their respective simulator-geometry rates.
With Map~2.0 fixed, the pathfinder leads the local planner by
20.52 and 19.50 SR points and by 14.33 and 12.96 SPL points.

Changing the geometry source also changes the depth available to the
local planner during recovery. Successive turns provide little parallax
for estimating clearance, and the four learned-motion runs with
reconstructed geometry had a known issue in how they supplied recovery
depth. On \datasetname{}, the
simulator-geometry runs also select memory blocks differently from the
reconstructed-geometry runs (Appendix~\ref{sec:benchmark_scoring}).
The reported gaps include these differences. With navigation weights
trained on Map~1.0 memories, the Map~2.0 runs nevertheless have higher
full-split SR on both benchmarks than their Map~1.0 counterparts. The gains are
4.15 points on GOAT-Bench and 2.35 on \datasetname{} with learned execution, and 1.01 and 3.46
points with simulator pathfinding.

\paragraph{Difficulty across benchmarks.}
All eight systems evaluated on both subsets have lower SR and SPL on
\datasetname{} (\cref{tab:standard_navigation}). Across the six external
goal predictors, GPT-6 Astra, and the six \method configurations,
the reported subset SR gap ranges from 16.42 percentage points for SCOPE
to 41.02 for MSGNav. Both use simulator pathfinding, so the gap
persists even when the simulator handles travel. The same pattern holds
for all six \method configurations on the full splits and for Modular GOAT
when compared with its published results.

The observed gap combines differences in sites, goal modalities, and
stopping rules. A familiar association between an object and a room may
guide search less reliably across shops, offices, and public facilities.
The frequency of repeat targets could also matter. A robot can reuse a
past goal for a repeat request, while finding an object seen only in
passing requires retaining the earlier sighting. The aggregate scores
leave the contributions of these factors unresolved.

\subsection{Memory reuse across requests}
\label{sec:memory_eval}

We first ask how often the robot reaches a relocated target under three
policies for checking its remembered location. We then examine how much
history the robot retains between requests and how far it can explore
before the first request.

\paragraph{Memory rechecks after object relocation.}
The three policies in \cref{tab:object_relocation} differ in which
targets can trigger a recheck before stopping. The robot queries its
updated memory and can retry the request after removing nearby older
observations (\cref{sec:cold_start}). The baseline allows a recheck for
any target, the second policy disables it, and the third permits it only
for targets flagged as movable. All three use the same placement changes, simulator geometry, and
pathfinding. These relocation episodes allow 2,000 actions per request. The standard benchmark uses different episodes and allows 500. Only the evaluator knows when and
where an object moves.

\begin{table}[!htb]
\centering\small
\setlength{\tabcolsep}{3pt}
\caption{Arrival rechecks on \relocationRequests{} \datasetname{}
requests under simulator depth, pose, and reference-grid pathfinding.
Only the last policy receives a target-mobility flag. There are 655
fixed-target requests and 450 movable-target requests, 225 at each
placement. Percentages average over the relevant requests within each
episode, then over episodes containing such requests. SoftSPL credits
progress weighted by path efficiency. All 150 episodes enter the overall
scores, including episodes that ended in an execution error and received
zero credit.}
\label{tab:object_relocation}
\begin{tabular*}{\linewidth}{@{\extracolsep{\fill}}lrrrrrrr@{}}
\toprule
& \multicolumn{3}{c}{All requests} & Fixed targets & \multicolumn{3}{c}{Movable targets, SR}\\
\cmidrule(lr){2-4}\cmidrule(lr){5-5}\cmidrule(l){6-8}
Targets eligible for recheck & SR & SPL & SoftSPL & SR & All & Before move & After move\\
\midrule
All targets (baseline) & 30.71 & 19.90 & 36.17 & 36.71 & 23.50 & 31.67 & 15.33\\
Recheck disabled & 30.38 & 21.77 & 40.56 & 40.20 & 16.83 & 30.33 & 3.33\\
Flagged movable targets (prior) & 33.87 & 23.84 & 40.40 & 41.14 & 24.33 & 34.33 & 14.33\\
\bottomrule
\end{tabular*}
\end{table}

On the 225 requests issued after a target moved, the all-target
baseline reached 15.33\% SR, compared with 3.33\% when rechecking
was disabled. Across all \relocationRequests{} requests the difference
was smaller (30.71\% versus 30.38\% SR), and the baseline had lower SPL (19.90\% versus 21.77\%). SoftSPL,
which credits partial progress weighted by path efficiency, was also
lower (36.17\% versus 40.56\%).
The policy with the mobility flag had the highest overall SR
(33.87\%). After a move, its SR was 14.33\%, compared with 15.33\%
when any target could trigger a recheck. Each policy ran on a
different software revision. Post-move scores cover requests for a target after its
scheduled relocation, whether or not the robot had found it at its
earlier location.

\paragraph{Memory retained across requests.}
The reference in \cref{tab:memory_carryover}a is the same GOAT-Bench
subset run as the simulator-geometry goal predictor in
\cref{tab:standard_navigation}. It scores \goatGoalSR{}\% SR with ordinary
memory carryover. Earlier runs on the same 36 episodes scored 61.73\%
with only the most recent 5\,m of travel available and 75.04\% with
prebuilt, higher-coverage memory. The runs span software revisions, and
the records do not establish that geometry and motion settings were
held fixed (Appendix~\ref{sec:paired_protocol}).

\begin{table}[!htb]
\centering\small
\setlength{\tabcolsep}{4pt}
\caption{Memory retention and initial exploration. Panel (a) covers
36 episodes and 278 requests. Panel (b) compares two initial exploration
budgets using simulator geometry and pathfinding. For each benchmark,
both budgets were evaluated on the same 360 episodes. The standard-budget
control was run separately from the full-split comparison in
\cref{tab:standard_navigation}.
Scores are percentages. Episode SR credits an episode only if every request succeeds.}
\label{tab:memory_carryover}
\begin{tabularx}{\linewidth}{@{}Lrrrr@{}}
\toprule
& \multicolumn{3}{c}{Subtask} & Episode\\
\cmidrule(lr){2-4}\cmidrule(l){5-5}
Condition & SR$\uparrow$ & SPL$\uparrow$ & SoftSPL$\uparrow$ & SR$\uparrow$\\
\midrule
\multicolumn{5}{@{}l}{\emph{(a) Memory across requests on the GOAT-Bench subset}}\\[2pt]
Carry memory across requests (\cref{tab:standard_navigation} reference) & \goatGoalSR & \goatGoalSubsetSPL & \goatGoalSubsetSoftSPL & \goatGoalSubsetEpisodeSR\\
Retain the most recent $\sim$5\,m of travel & 61.73 & 31.32 & 28.62 & 13.89\\
Supply prebuilt, high-coverage memory & 75.04 & 44.67 & 34.63 & 27.78\\
\midrule
\multicolumn{5}{@{}l}{\emph{(b) Initial exploration budget on the full benchmarks}}\\[2pt]
GOAT-Bench, 200 actions (standard) & 63.21 & 35.95 & 33.49 & 13.06\\
GOAT-Bench, 50 actions & 57.19 & 34.05 & 34.92 & 10.56\\[3pt]
\datasetname{}, 1,000 actions (standard) & 47.78 & 27.57 & 40.91 & 8.06\\
\datasetname{}, 200 actions & 39.08 & 22.22 & 37.07 & 4.17\\
\bottomrule
\end{tabularx}
\end{table}

\paragraph{Exploration before the first request.}
\Cref{tab:memory_carryover}b varies the action budget for initial
exploration. Goal-directed exploration remains available during each request.
Changing the cap can affect both the views in memory and the pose from
which the robot starts its first request. GOAT-Bench charges initial
exploration to the first request's 500-action budget. \datasetname{}
gives it a separate allowance.

On GOAT-Bench, cutting the cap from 200 to 50 actions lowered subtask
SR by 6.02 percentage points and SPL by 1.90 points. On \datasetname{},
cutting it from 1,000 to 200 actions lowered SR by 8.70 points and SPL
by 5.35 points. GOAT-Bench SoftSPL rose by 1.43 points with the shorter
cap, so average path-weighted progress improved even as fewer requests
succeeded.

\subsection{Model predictions from spatial memory}
\label{sec:model_diagnostics}

A robot can miss a goal because it chose the wrong object, misjudged
its own position, or predicted a poor route. We examine these
predictions on fixed requests. For grounding, we match predicted
centers to annotated objects. For localization and planning, we supply
a goal coordinate and evaluate both the estimated pose of the query camera
and the route to the goal. Average displacement
error (ADE) measures the distance between corresponding predicted and
reference waypoints (\cref{eq:route_ade}). We also illustrate how grounding and question answering can help
distinguish two objects that fit the same request (\cref{fig:clarification_example}).

All variants use the same HM3D requests. On \datasetname{} the offline
diagnostic cohort is separate from the \releaseEvalScenes{} navigation
sites. We rank 151 scenes by mean error in relative block translation
and retain the 136 with the smallest errors. All configurations use
the resulting 8,472 grounding and 1,669 coordinate-navigation requests. Appendix~\ref{sec:diagnostic_protocol}
gives the selection and scoring details.

The first study varies the memory at inference while holding model
weights fixed (\cref{tab:memory_inference}). The second varies the
training mixture and voxel width under a common training budget
(\cref{tab:memory_training}).

\newcommand{\navAblationMetricKey}{%
  For object grounding, we match predicted and annotated centers one to
  one within 1\,m. Precision (Prec.) gives the percentage of predictions
  matched and recall (Rec.) the percentage of annotations recovered.
  GPos is global camera position error (m). LPos (m) and LRot (degrees)
  give the position and rotation errors after correcting block placement.
  Route ADE is the mean distance between corresponding predicted and
  reference waypoints (m). Pose and route means use valid outputs for
  each metric. Bold marks the best value per column.}
\definecolor{ablationLayout}{HTML}{1452BB}
\definecolor{ablationBackend}{HTML}{007A87}
\definecolor{ablationResolution}{HTML}{307323}
\definecolor{ablationOrientation}{HTML}{762BA8}
\definecolor{ablationTrainingData}{HTML}{AF2346}
\definecolor{ablationVoxels}{HTML}{905800}
\begin{table}[!htb]
\centering\footnotesize
\setlength{\tabcolsep}{2.5pt}
\caption{Memory construction at inference on fixed HM3D requests and
requests from the 136 selected \datasetname{} scenes. Across rows,
grounding and localization use one checkpoint trained with \datasetname{}
data and 32\,cm voxels, and route prediction uses one flow-matching model.
Each colored label changes the corresponding reference setting.
In the common-frame conditions, reference block poses align points before
voxel sampling. The orientation variant changes pitch and roll during
memory capture and keeps the query images fixed. \navAblationMetricKey}
\label{tab:memory_inference}
\begin{tabular*}{\linewidth}{@{\extracolsep{\fill}}l*{6}{r}@{\hspace{8pt}}*{6}{r}@{}}
\toprule
 & \multicolumn{6}{c}{HM3D} & \multicolumn{6}{c}{\datasetname{}}\\
\cmidrule(lr){2-7}\cmidrule(l){8-13}
 & \multicolumn{2}{c}{Grounding} & \multicolumn{3}{c}{Pose error} & Route & \multicolumn{2}{c}{Grounding} & \multicolumn{3}{c}{Pose error} & Route\\
Memory configuration & Prec.$\uparrow$ & Rec.$\uparrow$ & GPos$\downarrow$ & LPos$\downarrow$ & LRot$\downarrow$ & ADE$\downarrow$ & Prec.$\uparrow$ & Rec.$\uparrow$ & GPos$\downarrow$ & LPos$\downarrow$ & LRot$\downarrow$ & ADE$\downarrow$\\
\midrule
\multicolumn{13}{@{}l}{\textbf{Reference settings}\quad \textcolor{ablationLayout}{one block in a common frame}\enspace$\cdot$\enspace \textcolor{ablationBackend}{LingBot-Map 1.0}\enspace$\cdot$\enspace \textcolor{ablationResolution}{$640{\times}480$ images}\enspace$\cdot$\enspace \textcolor{ablationOrientation}{memory views: zero pitch and roll}}\\
\addlinespace[1pt]
\textbf{Reference} & 64.90 & 53.57 & 0.259 & 0.259 & 4.805 & 0.297 & \textbf{47.70} & \textbf{38.82} & 0.538 & 0.538 & 5.666 & 0.571\\
\midrule
\textcolor{ablationLayout}{Keep blocks in local frames} & 65.07 & \textbf{53.92} & 0.377 & 0.280 & 5.140 & 0.358 & 46.74 & 38.19 & 0.597 & 0.511 & \textbf{5.491} & 0.611\\
\textcolor{ablationBackend}{Use LingBot-Map 2.0} & \textbf{65.84} & 53.43 & \textbf{0.254} & \textbf{0.254} & \textbf{4.801} & \textbf{0.294} & 47.57 & 37.69 & 0.557 & 0.557 & 5.750 & 0.591\\
\textcolor{ablationResolution}{Use $320{\times}224$ memory images} & 64.48 & 52.48 & 0.274 & 0.274 & 5.185 & 0.300 & 46.33 & 37.92 & \textbf{0.509} & \textbf{0.509} & 5.701 & \textbf{0.554}\\
\textcolor{ablationOrientation}{Vary pitch and roll in memory views} & 65.65 & 50.96 & 0.273 & 0.273 & 5.208 & 0.315 & 46.74 & 36.35 & 0.563 & 0.563 & 6.116 & 0.600\\
\bottomrule
\end{tabular*}
\end{table}

\paragraph{Memory design at inference.}
The reference memory in \cref{tab:memory_inference} uses
$640{\times}480$ LingBot-Map~1.0 input views. Reference block poses
align the reconstructed points in a common frame before sampling on a
32\,cm voxel grid. Keeping blocks in local frames instead and sampling
them separately changes grounding precision and recall by less than one
percentage point on either dataset. For the same predicted local-frame
camera poses, correcting block placement during scoring reduces position
error from 0.377 to 0.280\,m on HM3D and from 0.597 to 0.511\,m on
\datasetname{}. Comparing the two memory layouts also includes a change
in which points voxel sampling retains.

With common-frame sampling, changing the backend to LingBot-Map~2.0
raises HM3D grounding precision from 64.90\% to 65.84\% and lowers
route ADE from 0.297 to 0.294\,m. On the selected \datasetname{} scenes,
grounding recall falls from 38.82\% to 37.69\% and route ADE rises
from 0.571 to 0.591\,m. This is waypoint error on fixed requests
from 136 scenes, with reference poses aligning the memory blocks.
On the 36 held-out navigation sites, full-split SR with the local planner
is 23.93\% with Map~1.0 and 26.28\% with Map~2.0
(\cref{tab:standard_navigation}). There, the navigator reads local
blocks and each action determines its next observation.

The last two variants keep the query images fixed. Resizing and
padding the memory views
to $320{\times}224$ reduces grounding recall by about one percentage
point on each dataset. In the reference configuration the camera can
turn left or right, with pitch and roll fixed at zero. Allowing it to
point up or down and tilt the horizon during memory capture reduces recall
by 2.61 points on HM3D and 2.47 on \datasetname{}.

\begin{table}[!tb]
\centering\footnotesize
\setlength{\tabcolsep}{2.5pt}
\caption{Training data and voxel width on the same evaluation requests as
\cref{tab:memory_inference}. At 32\,cm the runs use none, roughly one-third,
or all of the \datasetname{} training scenes available to this study. The
two full-data runs differ only in voxel width, applied in training and
evaluation. All runs use separate LingBot-Map~1.0 blocks and
$640{\times}480$ memory views with zero pitch and roll. Each model
trains for 20,000 optimizer updates on 32 GPUs (640,000 GPU-updates), a
smaller budget than the grounding and pose reference in
\cref{tab:memory_inference}. Route scores use each model's flow-matching
head. \navAblationMetricKey}
\label{tab:memory_training}
\begin{tabular*}{\linewidth}{@{\extracolsep{\fill}}l*{6}{r}@{\hspace{8pt}}*{6}{r}@{}}
\toprule
 & \multicolumn{6}{c}{HM3D} & \multicolumn{6}{c}{\datasetname{}}\\
\cmidrule(lr){2-7}\cmidrule(l){8-13}
 & \multicolumn{2}{c}{Grounding} & \multicolumn{3}{c}{Pose error} & Route & \multicolumn{2}{c}{Grounding} & \multicolumn{3}{c}{Pose error} & Route\\
Training configuration & Prec.$\uparrow$ & Rec.$\uparrow$ & GPos$\downarrow$ & LPos$\downarrow$ & LRot$\downarrow$ & ADE$\downarrow$ & Prec.$\uparrow$ & Rec.$\uparrow$ & GPos$\downarrow$ & LPos$\downarrow$ & LRot$\downarrow$ & ADE$\downarrow$\\
\midrule
\multicolumn{13}{@{}l}{\textbf{Reference settings}\quad \textcolor{ablationTrainingData}{Training mixture without \datasetname{} data}\enspace$\cdot$\enspace \textcolor{ablationVoxels}{32\,cm voxels}}\\
\addlinespace[1pt]
\textbf{Reference} & \textbf{58.48} & 54.21 & 0.597 & 0.451 & \textbf{9.329} & 0.478 & 23.47 & 25.94 & 3.396 & 1.932 & 22.355 & 2.048\\
\midrule
\textcolor{ablationTrainingData}{Add $\sim$1/3 \datasetname{} training data} & 56.94 & \textbf{54.56} & 0.537 & \textbf{0.448} & 9.654 & 0.467 & 30.67 & \textbf{36.82} & 1.402 & 1.163 & 13.847 & 1.068\\
\addlinespace[2pt]
\textcolor{ablationTrainingData}{Add \datasetname{} training data} & 56.18 & 54.39 & \textbf{0.527} & 0.459 & 9.330 & \textbf{0.449} & \textbf{31.15} & 36.65 & \textbf{1.008} & \textbf{0.896} & \textbf{11.946} & \textbf{0.850}\\
\addlinespace[2pt]
\begin{tabular}[c]{@{}l@{}}\textcolor{ablationTrainingData}{Add \datasetname{} training data}\\\textcolor{ablationVoxels}{+ widen voxels to 64\,cm}\end{tabular} & 50.12 & 48.58 & 0.676 & 0.578 & 11.533 & 0.552 & 26.96 & 31.38 & 1.302 & 1.169 & 13.986 & 1.009\\
\bottomrule
\end{tabular*}
\end{table}

\paragraph{Training data and spatial sampling.}
All models in \cref{tab:memory_training} use the local-block layout from
\cref{tab:memory_inference}, and each training configuration has its own weights.
The reference training mixture starts with HM3D
scene and route examples, GOAT-Bench object descriptions, and ScanNet
question--answer pairs. Adding \datasetname{} brings in scene and
route examples, candidate-choice dialogues, and sign-text targets. We
use either the full training pool available to this ablation or a fixed
sample covering roughly one-third of its scenes. The full pool extends
beyond the \releaseTrainScenes{} publicly released training scenes.
The other sources, evaluation requests, and training budget stay fixed.

At 32\,cm voxel width, adding the full \datasetname{} pool raised
grounding recall from 25.94\% to 36.65\% on the 136 selected scenes.
Global camera position error fell from 3.396 to 1.008\,m and route ADE
from 2.048 to 0.850\,m. The gain in HM3D grounding recall was smaller
at 0.18 percentage points, and precision fell from 58.48\%
to 56.18\%. With the full mixture, the 32\,cm model has higher
precision and recall and lower pose and route errors than the 64\,cm
model on both datasets.

The roughly one-third sample yields similar grounding scores to the
full pool on \datasetname{} (30.67\% versus 31.15\% precision and
36.82\% versus 36.65\% recall). The larger pool made a clearer difference to camera localization
and route prediction. With the smaller pool, global camera position error is
1.402\,m and route ADE is 1.068\,m, compared with 1.008\,m and
0.850\,m for the full pool.

\par\medskip\noindent
\begin{minipage}[t]{.46\linewidth}
\vspace{0pt}
\paragraph{Proactive target clarification.}
A request for ``the fan'' leaves two possible destinations in
\cref{fig:clarification_example}. Multi-object grounding returns a
center and an observation viewpoint for each fan. Given the centers
and a question about how the fans differ, the decoder uses spatial
memory to identify one in the office and the other in the dining area.
Together, these outputs could support a single reply that names both
options and asks which the user meant. Choosing the office fan would
select its observation viewpoint for route planning.
Appendix~\ref{app:retrieval_prompts} gives the prompts for both operations.
\end{minipage}\hfill
\begin{minipage}[t]{.5\linewidth}
\vspace{0pt}\centering
\begingroup
\definecolor{cfink}{HTML}{203349}
\definecolor{cfblue}{HTML}{245B88}
\definecolor{cfteal}{HTML}{247B79}
\definecolor{cfline}{HTML}{CBD8E3}
\definecolor{cfgray}{HTML}{F5F7F9}
\definecolor{cfpale}{HTML}{EDF6F5}
\resizebox{\linewidth}{!}{%
\begin{tikzpicture}[x=1bp,y=-1bp,>=stealth,
  every node/.style={inner sep=0pt,outer sep=0pt,text=cfink,font=\sffamily\fontsize{13}{16}\selectfont},
  title/.style={anchor=west,font=\sffamily\bfseries\fontsize{14}{16}\selectfont},
  role/.style={anchor=west,font=\sffamily\bfseries\fontsize{12}{14}\selectfont},
  card/.style={rounded corners=4bp,draw=cfline,line width=.7bp},
  flow/.style={->,line width=1.25bp,draw=cfteal}]
\node[title,text=cfblue] at (0,9) {Illustrative conversation};
\node[title,text=cfteal] at (312,9) {Behind the reply};

\draw[draw=cfblue!50,line width=.8bp] (266.8,87.2) -- (313.2,26.2)
                                      (266.8,181.8) -- (313.2,240.8);
\draw[card,fill=cfgray] (0,25) rectangle (268,66);
\node[role] at (13,37) {User};
\node[anchor=west] at (13,54) {Take me to the fan.};

\draw[card,draw=cfblue!55,fill=cfblue!6] (0,86) rectangle (268,183);
\node[role,text=cfblue] at (13,100) {Robot};
\node[anchor=west] at (13,121) {I found two fans:};
\node[anchor=west] at (13,138) {A. The one in the office.};
\node[anchor=west] at (13,154) {B. The one in the dining area.};
\node[anchor=west] at (13,172) {Which one do you mean?};

\draw[card,fill=cfgray] (0,201) rectangle (268,242);
\node[role] at (13,213) {User};
\node[anchor=west] at (13,230) {A, please.};

\draw[card,draw=cfblue!55,fill=white] (312,25) rectangle (608,242);
\begin{scope}[shift={(321,8)}]
\draw[card,fill=white] (0,25) rectangle (278,139);
\node[title] at (12,39) {1. Multi-object grounding};
\node[anchor=north] at (73,52) {\includegraphics[viewport=86.4 40.32 374.4 256.32,clip,width=88bp]{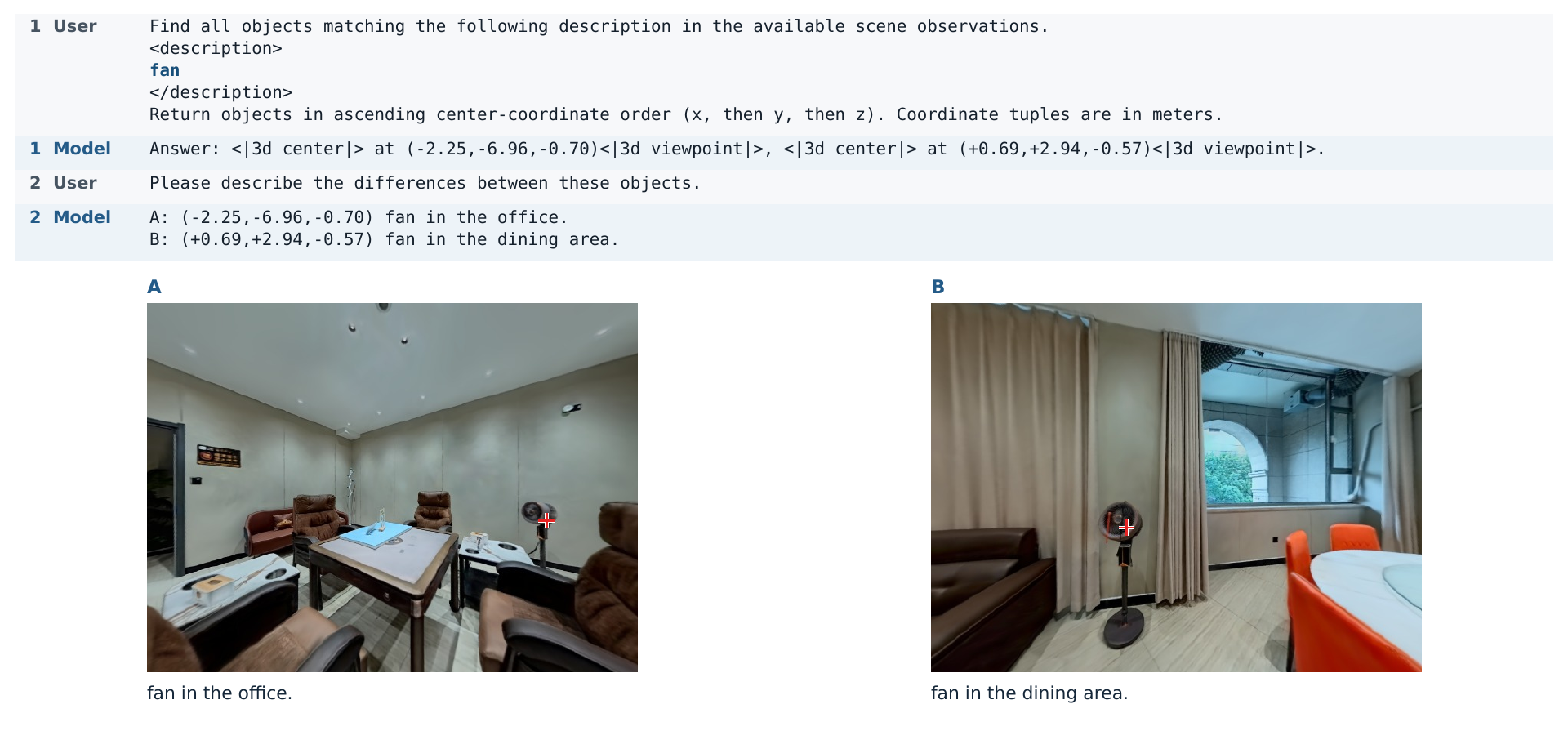}};
\node[anchor=north] at (205,52) {\includegraphics[viewport=547.2 40.32 835.2 256.32,clip,width=88bp]{figures/dialogue_fans.pdf}};
\node at (139,128) {Illustrative views: fan A (left) and fan B (right)};

\draw[card,fill=cfpale] (0,159) rectangle (278,224);
\node[title] at (12,173) {2. QA over spatial memory};
\node[anchor=west] at (12,194) {How do the two fans differ?};
\node[anchor=west,text=cfteal] at (12,212) {A: in the office\qquad B: in the dining area};
\draw[flow] (139,140) -- (139,158);
\end{scope}
\end{tikzpicture}%
}
\endgroup

\captionof{figure}{Preliminary support for proactive clarification. The left
panel illustrates a reply that offers both destinations. The inset
summarizes two recorded model outputs. Grounding locates both fans (1), and
memory-conditioned question answering distinguishes them by room (2).
The candidate views mark the predicted centers (red crosses) for the reader.
Appendix~\ref{sec:dialogue_cases} shows grounding and comparison
transcripts for sofas and jackets.}
\label{fig:clarification_example}
\end{minipage}\par

\subsection{Inference time and memory footprint}
\label{sec:resource_profile}
When successive requests use the same memory, the spatial memory
encoder (a) can compute its keys and values once. In NVIDIA TensorRT
using bfloat16 (BF16), encoding 18,846 points in three blocks took
3.002\,s on a Jetson Thor (T5000) and 1.859\,s on an RTX A6000
(\cref{tab:resource_stages}). With that encoding cached, we measure
GPU time from encoding a request's images to returning its route.
Memory updates, movement, and arrival checks fall outside this interval.

All four workloads localize the current view. An object-description
request uses the observation viewpoint paired with a single grounded
object. A coordinate request supplies the destination directly, a
reference-image request localizes a second image, and goal-directed
exploration predicts a frontier from partial memory and the unresolved
request. These are the operations in \cref{tab:task_interfaces,fig:system},
with prompts in Appendix~\ref{app:route_prompts}. The single-match
workload excludes the extra dialogue turns used for clarification.

\begin{table}[!htb]
\centering\footnotesize
\setlength{\tabcolsep}{3pt}
\caption{TensorRT inference in bfloat16 (BF16) for a memory of 18,846
points in three blocks. Module letters refer to \cref{fig:architecture}.
Each device header gives the one-time cost of encoding the memory.
Request totals include the shared vision transformer (ViT) and both
decoders, including the flow head in (c). For decoder (b), counts are
mean forward passes rounded to the nearest integer. Times include all
passes and readouts for a completed request. Peak is process GPU memory
after scene encoding. The hardware ratings give theoretical dense FP16
throughput. The Thor estimate in MAXN power mode is half the published
sparse peak.}
\label{tab:resource_stages}
\begin{tabular*}{\linewidth}{@{\extracolsep{\fill}}lrrrrrr@{}}
\toprule
\multicolumn{7}{@{}l}{\textbf{Jetson Thor (T5000)}\quad 128 GB shared LPDDR5X, $\approx$258.5 peak dense FP16 TFLOP/s (MAXN)}\\
\multicolumn{7}{@{}l}{\emph{One-time spatial memory encoding (a)}\quad 3.002\,s}\\[3pt]
& Shared ViT & \multicolumn{2}{c}{Vision-language decoder (b)} & \shortstack[r]{Trajectory\\decoder (c)} & Request total & Peak\\
\cmidrule(lr){3-4}
Task & ms & Mean forward passes & Time (ms) & ms & ms & GiB\\
\midrule
Navigation using an object description & 23.71 & 5 & 452.79 & 116.58 & 593.08 & 18.07\\
\addlinespace[4pt]
Coordinate navigation & 24.98 & 2 & 235.68 & 117.31 & 377.97 & 18.11\\
\addlinespace[4pt]
Navigation using a reference image & 48.62 & 4 & 464.62 & 115.77 & 629.01 & 18.16\\
\addlinespace[4pt]
Goal-directed exploration & 27.13 & 3 & 401.96 & 117.22 & 546.30 & 18.13\\
\addlinespace[4pt]
\midrule
\multicolumn{7}{@{}l}{\textbf{RTX A6000}\quad 48 GB GDDR6, 154.8 peak dense FP16 TFLOP/s}\\
\multicolumn{7}{@{}l}{\emph{One-time spatial memory encoding (a)}\quad 1.859\,s}\\[3pt]
& Shared ViT & \multicolumn{2}{c}{Vision-language decoder (b)} & \shortstack[r]{Trajectory\\decoder (c)} & Request total & Peak\\
\cmidrule(lr){3-4}
Task & ms & Mean forward passes & Time (ms) & ms & ms & GiB\\
\midrule
Navigation using an object description & 15.97 & 5 & 312.02 & 54.72 & 382.71 & 16.56\\
\addlinespace[4pt]
Coordinate navigation & 16.15 & 2 & 147.03 & 54.51 & 217.69 & 16.60\\
\addlinespace[4pt]
Navigation using a reference image & 32.56 & 4 & 295.32 & 55.20 & 383.07 & 16.63\\
\addlinespace[4pt]
Goal-directed exploration & 16.14 & 3 & 251.48 & 54.22 & 321.84 & 16.60\\
\bottomrule
\end{tabular*}
\end{table}

The vision-language decoder (b) accounts for the largest share of
request time on both devices for every workload. It uses multiple
forward passes to generate the response and read out the required
positions or poses. The count varies with the task and generated
response length. Once the model chooses a response template, its fixed
text can be supplied directly. For a reference-image request
on Thor, the two images take 48.62\,ms to encode, and decoder~(b)
takes 464.62\,ms of the 629.01\,ms request total. The trajectory
decoder (c) computes its query features in one backbone pass. Its
time includes prompt processing, cache transfers, and the flow head's
16 Euler updates. Each Euler update reuses the same query features
(\cref{eq:trajectory_flow}).

\begin{table}[!t]
\centering\footnotesize
\setlength{\tabcolsep}{2.5pt}
\caption{TensorRT inference on Thor using bfloat16 (BF16) and 8-bit
floating point (FP8). The navigation transformer handles a single
object-description request with a cached 18,846-point memory in three
blocks. Its BF16 row uses the object-description measurement in
\cref{tab:resource_stages}. The local planner is timed separately for one control step.
Peak is process GPU memory. Offsets measure distances between corresponding
TensorRT and PyTorch waypoints on the same device and inputs. P95 is the
95th percentile. The two transformer rows differ in both implementation
and compute precision. Both local-planner rows use native TensorRT operators.}
\label{tab:resource_scale}
\begin{tabular*}{\linewidth}{@{\extracolsep{\fill}}l*{5}{r}@{\hspace{10pt}}*{5}{r}@{}}
\toprule
& \multicolumn{5}{c}{Navigation transformer} & \multicolumn{5}{c}{Local planner}\\
\cmidrule(lr){2-6}\cmidrule(l){7-11}
TensorRT & Time & Peak & \multicolumn{3}{c}{\shortstack{Offset from PyTorch (cm)\\same flow checkpoint}}
& Time & Peak & \multicolumn{3}{c}{\shortstack{Offset from PyTorch (cm)\\BF16 vision encoder, FP32 planning head}}\\
precision & (ms) & (GiB) & Mean & P95 & Max & (ms) & (GiB) & Mean & P95 & Max\\
\midrule
BF16 & 593.08 & 18.07 & 0.00 & 0.00 & 0.00 & 94.98 & 0.93 & 0.40 & 0.87 & 1.68\\
FP8 & 361.23 & 14.28 & 10.86 & 28.19 & 32.15 & 66.62 & 0.66 & 1.61 & 5.88 & 11.63\\
\bottomrule
\end{tabular*}
\end{table}

\Cref{tab:resource_scale} evaluates 8-bit floating-point (FP8) inference
on Thor against the BF16 configuration. BF16 and FP8 name the precision used for the main computations, while
some operations run at higher precision.
For an object-description request, the BF16 transformer in the table
takes 593.08\,ms. The BF16 build uses custom TensorRT operators and
recomputes some decoder states to reproduce the PyTorch waypoint
predictions. A second BF16
build uses native TensorRT operators and reuses more cached decoder
state. It takes 466.42\,ms with a mean waypoint offset of 3.42\,cm from
PyTorch. Running this second build in FP8 takes 361.23\,ms with a mean
offset of 10.86\,cm. The 39\% reduction in navigation-transformer time therefore reflects
changes in both implementation and precision.

For one local-planner control step, BF16 and FP8 take
94.98 and 66.62\,ms and yield mean offsets of 0.40 and 1.61\,cm,
respectively. All offsets in \cref{tab:resource_scale} compare TensorRT
waypoints with PyTorch predictions for the same inputs. Route ADE in
\cref{tab:memory_inference,tab:memory_training} measures distance to
reference routes on other requests. Measuring the effect of FP8 on ADE
would require scoring both precisions against the same reference routes.

On the GOAT-Bench subset, calls to GPT-6 Astra through
NavHarness~\cite{navharness} averaged \goatHarnessTokensPerSubtask{}
tokens per subtask, including 33,257 uncached input tokens, 348,761 cached input
tokens, and 2,007 output tokens. At the rates recorded for the run,
the API cost was about \$\goatHarnessCostPerSubtask{} per subtask.
Across 278 arrival checks, the time from the start of the model turn
to the complete reply averaged \goatHarnessArrivalSeconds{}\,s
(median 14.4\,s, 95th percentile 24.7\,s). This wall-clock measurement
covers a hosted arrival decision. The local GPU measurements in
\cref{tab:resource_stages} cover route generation. \method performs
both navigation and arrival decisions without a hosted model call.

\subsection{Failure modes}
\label{sec:navigation_failure_modes}

The examples in Appendix~\ref{sec:case_analysis} trace five ways a
request can fail, from placing observations in memory to choosing where
the robot should explore next.

\paragraph{Misregistered memory blocks.}
Registration errors range from a misplaced block to a distorted layout.
With local points held fixed, the predicted poses displace a room-sized
block in the first scene, shift a block less in the second, and compress
a distant wing of the shopping mall in the third
(\cref{fig:block_alignment_bad}). Changing the mall's reference block
shifts which part of its reconstructed layout aligns most closely with
the reference (\cref{fig:block_alignment_anchors}).
For blocks captured in sequence, odometry could constrain their relative
displacement and gyroscope readings from an inertial measurement unit
(IMU) could constrain their relative rotation. Visual matches on revisits
could then help correct accumulated drift, as in visual--inertial
simultaneous localization and mapping~\cite{orbslamthree}.

\paragraph{Incorrect or missing camera pose.}
Recognizing a scene feature can leave the camera position or orientation
unresolved. The chair backs and the television wall in
\cref{fig:localization_bad} are recognizable from the predicted views,
yet both positions are displaced. Among repeated columns, the estimate
also faces nearly the opposite direction. A dark query yields no pose
even though the memory covers its location. During live navigation,
the controller already tracks motion between reliable visual fixes
using odometry aligned to memory (\cref{sec:cold_start}). This lets it
update its pose through a poor view and check a new visual match against
the distance traveled and angle turned since the last fix.

\paragraph{Incorrect or incomplete object grounding.}
The grounding examples compare predicted centers with annotated objects. The model selects a dining table for a tea-table request,
places an ice machine far from its annotated center, and returns extra
shelf centers while missing two of three medicine shelves
(\cref{fig:grounding_bad}). Sign-text requests can yield no location, as
for the All-in sign, or an unannotated sign for the right brand while
both annotated signs are missed (\cref{fig:grounding_ocr}). An extra
prediction can also point to a genuine object outside the labeled
area, as with the sixth fan in \cref{fig:grounding_good}, where all
five annotated fans were recovered.

\paragraph{Route deviations and insufficient clearance.}
Fixed endpoints leave open which passage a route takes and how closely
it skirts obstacles. In \cref{fig:trajectory_bad}, some turns cut across
occupied cells beside furniture. Other paths depart from the reference
where the passage is difficult to distinguish in the point cloud. The
last row of \cref{fig:trajectory_good} illustrates a disagreement
between representations. The route passes through an apparent doorway
in memory but grazes an area marked occupied on the map. These overlays
assess the proposed route before the local planner acts. At execution
time, a check using live depth and the robot's footprint could trigger
replanning if the next turn leaves too little room to pass.

\paragraph{Choosing a different frontier.}
A frontier error changes where the next search begins. In the dark-curtain example in \cref{fig:exploration_bad},
both routes start in the same corridor. The model heads in one
direction, while the reference heads the other way and then turns into
the room on the right. The endpoint error here reflects the choice
of where to search before the local planner begins to move.

\section{Future Work}
\label{sec:future_work}

Our navigator builds spatial memory, queries it to resolve a request,
and follows the resulting route. Longer deployments raise questions
at each stage, from keeping memory accurate as a building changes to
deciding which requests warrant help from a person or a cloud model.

\paragraph{Learning geometry from language-aligned features.}
The navigator trained with LingBot-Map~1.0 memories already benefits in
closed-loop navigation from Map~2.0 reconstructions at inference
(\cref{tab:standard_navigation}).
Training on Map~2.0 memories as well could show whether matching the
training and inference geometry improves navigation further. A broader question is whether the features used for
grounding could also match surfaces across views. A spatial foundation model trained for both tasks would need to
recognize the same sign from an oblique or shadowed view while retaining
the single word that distinguishes one destination from the next.

\paragraph{Deciding when to revise a memory.}
An unobstructed view of the floor where a chair once stood gives the robot
a reason to reconsider its memory. A view blocked by a passing person
leaves the chair's location unresolved. In our relocation experiment, the robot can
recheck any target at its remembered location and refresh nearby memory (\cref{sec:memory_eval}). A more selective policy could weigh the evidence in the arrival
view when deciding whether to keep searching or revise the stored
location. Changes during a request raise a related problem. A person crossing a
doorway may warrant an immediate change of route while the doorway itself
remains a useful landmark for later tasks.

\paragraph{Learning when and how to clarify intent.}
When two objects fit a request, \method can describe their differences
and let the user select one by letter. Training examples for freer dialogue could record the plausible
matches, the question that helped a user distinguish them, and the
instance they confirmed. These exchanges could help train the model to choose questions that
resolve the ambiguity with little effort from the user. A distinction visible on the robot's map
can help only if the user recognizes it too.

\noindent
\begin{minipage}[t]{0.51\linewidth}
\vspace{0pt}
\paragraph{Encoding and querying large spatial memories efficiently.}
Recognizing a destination may take only a few surface tokens, while
routing to it can require evidence from every room along the way.
The number of active points affects both the one-time cost of
encoding the memory and the cost of each request. On Jetson Thor,
encoding three times as many points across three blocks took
5.2 times as long (\cref{tab:memory_size_thor}). Even with that
encoding cached, answering an object-description request took
47\% longer.
\end{minipage}\hfill
\begin{minipage}[t]{0.46\linewidth}
\vspace{0pt}
\centering\footnotesize
\captionof{table}{GPU time on Jetson Thor as the number of points in three
memory blocks increases. The first row is the BF16 object-description
workload in \cref{tab:resource_stages}. Times are shown separately for
encoding the memory once (s) and answering a request with the cached
encoding (ms).}
\label{tab:memory_size_thor}
\begin{tabular*}{\linewidth}{@{\extracolsep{\fill}}rrr@{}}
\toprule
Points & Encode once (s) & Request (ms)\\
\midrule
18,846 & 3.002 & 593.08\\
37,692 & 8.014 & 729.61\\
56,538 & 15.708 & 874.47\\
\bottomrule
\end{tabular*}
\end{minipage}\par

A scene graph could index rooms and their connections so the robot loads only
the blocks a request needs. Such an index would need to preserve the passages
leading to a goal as well as visual details that lack an object label.

\paragraph{Collaboration between onboard and cloud models.}
A robot can afford to deliberate over the order of several errands.
Steering around a person who steps into its path leaves far less time.
A more capable but slower cloud model could help with the first
decision while the onboard system retains control of motion.
The robot could share candidate destinations and the passages connecting
them, keeping camera images and detailed geometry on board. By the time
a reply arrives, a passage may have closed or a target may have moved.
The onboard model could ground the suggested goal in its current spatial
memory before acting. This would leave routine navigation available
offline and reserve cloud assistance for decisions worth the wait.

\section{Conclusion}
\label{sec:conclusion}

A detail seen in passing can matter long after the robot has left the room.
\method preserves that detail as a language-aligned visual feature at a
3D surface location. This gives the robot two uses for the same observation.
It can recognize what a user means and use the surrounding geometry to
find a way there. Successive requests in the \datasetname{} benchmark
put both abilities to the test. When an object moves, its old location
gives the robot a useful lead, and what it sees on arrival can change
where it should search next.

Navigation offers a test of what a future spatial foundation model
remembers. A robot can go to a predicted location and compare the view
it expects with the one it finds. If the features it uses to interpret
that view also help reconstruct the scene, a correction made on the
way to a destination could inform a later answer about the room, even
when no one asks the robot to move. The longer a robot works in a place, the less often people should
have to explain that place to it.

\FloatBarrier

\section{Contributors}
\label{sec:contributors}
\begingroup
\setlength{\parskip}{3pt}
\noindent\textbf{Model training:} Huanhang Hu, Xiao Li, Tianlong Sun, Jie Ding,
Yuzheng Liu, Zihan Wang, and Shizhe Zang.\par
\noindent\textbf{Scene asset pipeline:} Lingyuan Yang, Huanhang Hu, Bo Chen,
Ge Mu, and Zhenhang Guo.\par
\noindent\textbf{Navigation data annotation:} Tong Wang, Jie Ding, and Xiao Li.\par
\noindent\textbf{Model evaluation:} Jiawei Wang and Zixuan Wang.\par
\noindent\textbf{Deployment and applications:} Tong Wang and Zhenhang Guo.\par
\smallskip
\noindent Xing Zhu and Yujun Shen sponsored the project. Yangyan Li is the
corresponding author, and Xinghe Li led the project.\par
\endgroup

\section*{Acknowledgments}
We thank Yihang Chen, Lin-Zhuo Chen, Jian Gao, Jiacheng Dong, Nan Xue,
and Yinghao Xu for their contributions to LingBot-Map 1.0 and 2.0.
These models reconstruct scene geometry from short image sequences,
providing the 3D locations that \methodname associates with visual
features in its spatial memory.

We thank Jingmei Zhao and Qiaoying Chen for their help with data
procurement. Yibo Lu, Bo Pang, Yu Liu, Nan Wang, Mingkang Shi, and
Ruonan Wang contributed to the data pipeline. Pengtao Song, Ruonan Zhang,
and Zhi Ling helped optimize deployment. Liutao Li and Jialong Zhang
contributed to applications and validation.

{
\small
\bibliographystyle{plain}
\bibliography{references.bib}
}

\appendix
\section*{Appendix}
\section{Training data generation and supervision}
\label{sec:training_data}
A \emph{mapping trajectory} is a sequence of camera poses through a
reconstructed scene. Two trajectories through the same region can reveal
different objects or opposite sides of one object. We use their rendered
views to build training memories, determine which descriptions those
memories support, and generate localization and navigation examples.

\Cref{tab:data} lists the records eligible for sampling in the
20,000-update training run of November~12. They describe the available
pool for that run. The training comparisons in \cref{tab:memory_training}
vary the pool and sample from it during training. HM3D~\cite{hm3d} and \datasetname{} supply the
mapping trajectories. GOAT-Bench~\cite{goatbench} contributes object
descriptions, and ScanNet~\cite{scannet} supplies spatial question--answer
pairs. Rescans of the same room stay in the same ScanNet split. We count each camera pose before cropping and include grounding
instances even when only their centers are labeled.

\begin{table}[H]
\centering\small
\setlength{\tabcolsep}{4pt}
\caption{Eligible records before augmentation and sampling. Panel~(a)
gives the count for each regional task and scene source. Localization and
navigation share the same candidate poses, counted before route
feasibility checks. Panel~(b) counts additional object instances,
sign annotations, and question--answer pairs.}
\label{tab:data}
\begin{tabularx}{\linewidth}{@{}lLrr@{}}
\toprule
\multicolumn{4}{@{}l}{\textbf{(a) Tasks using HM3D and \datasetname{} regional memories}}\\[2pt]
Task & Unit counted and use during training & HM3D & \datasetname{}\\
\midrule
Grounding & Object instances matching at least one request. A category, a combination of attributes, or a free-form description identifies the centers to predict. & 121,031 & 53,855\\[3pt]
Localization & Candidate camera poses. A randomly cropped and resized trajectory frame provides the query image. & 7,730,840 & 7,396,283\\[3pt]
Navigation & Candidate start poses from the same pool as localization. The sampler pairs a start with a reachable endpoint and plans a route online. & 7,730,840 & 7,396,283\\[3pt]
Exploration & Stored start poses. Removing memory blocks yields a frontier and a route to it for a language or image target. & 181,437 & 139,663\\[3pt]
Dialogue & Distinct initial questions per region. An exchange retrieves the candidates, describes their differences, and identifies the one selected by letter. & 31,100 & 7,443\\
\bottomrule
\end{tabularx}
\par\smallskip
\begin{tabularx}{\linewidth}{@{}lLr@{}}
\toprule
\multicolumn{3}{@{}l}{\textbf{(b) Additional sources of supervision}}\\[2pt]
Source and task & Unit counted and use during training or validation & Count\\
\midrule
GOAT-Bench grounding & Distinct target instances across positive goal descriptions. Each description identifies the centers to predict. & 4,752\\[3pt]
\datasetname{} sign text & Retained text-target annotations. A positive target must be readable from the selected memory. & 1,485\\[3pt]
ScanNet QA (train) & Question--answer pairs from 1,030 scans of 505 rooms. & 163,626\\[3pt]
ScanNet QA (validation) & Question--answer pairs from 143 scans of 58 rooms. & 23,290\\
\bottomrule
\end{tabularx}
\end{table}

\subsection{Building the training memories}
\label{app:training_memories}
In each region we plan mapping trajectories from several starts and
divide them into short segments with overlapping views where feasible.
For \datasetname{} we render color from the 3D Gaussians and reference
depth from the mesh. For HM3D both come from the textured scene mesh.

LingBot-Map~1.0~\cite{lingbotmap} estimates camera poses and depths from
the color frames of each segment. To recover metric scale, we back-project
corresponding pixels using the predicted and reference geometry and fit
a similarity transform between the two point sets. Its scale factor
applies to both predicted depths and camera translations.

We pair each retained patch feature from the frozen Qwen vision
encoder with its 3D surface point, obtained by back-projecting the patch center
using the calibrated depth and camera pose. Each segment stores one
observation per occupied 4\,cm voxel, including the point in the first
camera's frame, its visual feature, and the source camera pose. This
fine grid lets the training sampler use coarser voxels and fewer input
tokens from the same observations.

\subsection{Labeling objects and visible text}
\label{app:object_labels}
\begin{figure}[htbp]
\centering
\includegraphics[width=\linewidth]{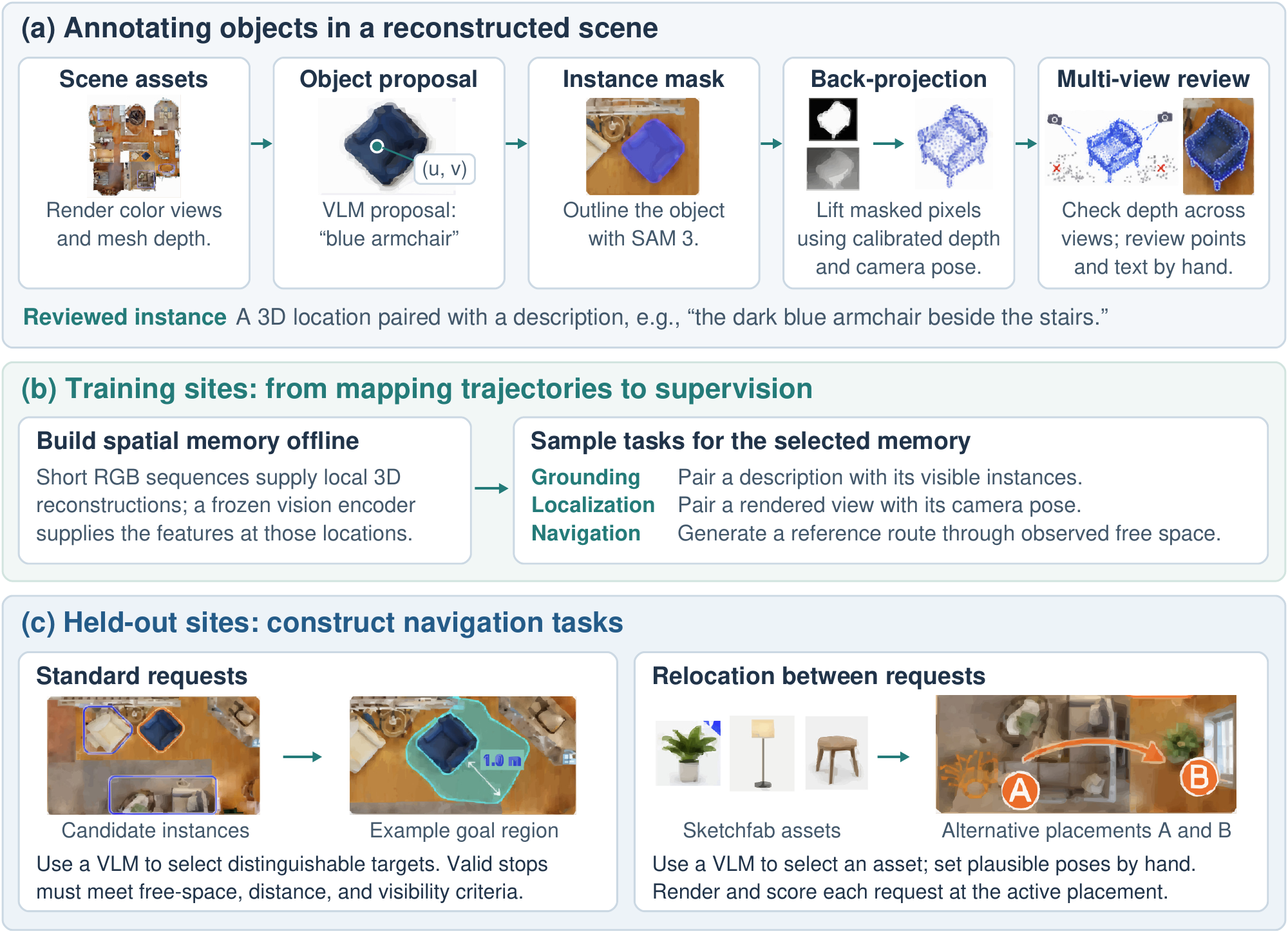}
\caption{From reconstructed \datasetname{} scenes to supervision and
navigation tasks. \textbf{(a)} Reviewers check the projected surface
points and description of a chair across views. \textbf{(b)} On
training sites, each selected memory supplies visible instances for
grounding, views with known camera poses for localization, and observed
free space for generating routes. \textbf{(c)} On held-out sites, the
shaded region illustrates proximity to a candidate target. A valid
stop also requires free space and visibility. The relocation example
shows two alternative placements of the same asset. Each request is
scored at the active placement.}
\label{fig:asset_supervision_pipeline}
\end{figure}

We prompt a VLM for the object categories visible in each view and group
synonymous names. Segment Anything Model~3 (SAM~3)~\cite{sam3} takes
these names as prompts and predicts an instance mask for each detection.
After checking its category and extent against the image, we lift the
mask into 3D using camera pose and aligned depth. Geometric overlap
links detections from different views. For the chair in
\cref{fig:asset_supervision_pipeline}a, other views help reject
background pixels in its mask and identify instance attributes such
as its subtype, color, or material. Comparing instances within a category
helps identify attributes that distinguish them in a request.

\suppressfloats[t]

To assign instances to a region, we project their sampled surfaces into
the recorded RGB-D views. A surface point counts as visible when its
projected depth agrees with the observation. We pool visible points
over all recorded starts in the region. The default filter retains a
confirmed instance if at least 30\% of its sampled surface is visible
across these views. All mapping trajectories that start in the region
share this object list.

A grounding request combines a category with sampled attributes or
uses a free-form description. We also use separately prepared long and
short English descriptions after removing annotation artifacts and
viewpoint-dependent wording. Multiple requests can refer to one counted
instance. Attribute queries target the centers of the 3D bounding boxes
of all matching instances. We discard queries with unresolved matches
or more than ten targets.

For a no-match example, we use either a confirmed object with no visible
surface samples in the region or a description from another scene in the
same split whose category family is absent from the region's visibility
records. GOAT-Bench queries retain their source descriptions and use
scene-level absence checks. We keep the GOAT-Bench validation regions
fixed when adding positive GOAT-Bench descriptions and sign-text examples
to the training mixture.
These sources supervise grounding. Viewpoint labels come from matched
regional HM3D or \datasetname{} instances.

For an observation-viewpoint label, we match the regional object by
instance ID to its rendered views. We sample one view and trace a
horizontal ray from the object's center toward the camera. If the
center lies in an obstacle cell, we place the viewpoint 0.25\,m
beyond the first exit from that obstacle. Otherwise, we place it
0.25\,m from the center along the ray. The label retains the object's
height and the sampled camera's horizontal heading. This construction
indicates which side to approach. The offset can land in an occupied
cell, so the executor still checks traversability. All mentions of the
same object in a dialogue share one sampled label. If the view or
occupancy map is unavailable, we retain the grounding example and
mask its viewpoint loss.

For shop and room signs, a separate pipeline proposes candidates, fits
a sign surface from depth, and renders close-ups to confirm and transcribe
the text. Position and name link repeated sightings. The resulting
annotation contains the text, a 3D anchor, and supporting surface points.

A close-up can verify the transcription even if the robot could not
read the sign from its mapping trajectory. A positive example must
therefore contain a view in the selected memory where the lettering
is unobstructed and passes the applicable text-size and readability
checks. Each retained annotation counts once across eligible starts.
A negative example requires separate evidence that the named text is
absent from the scene.

Dialogue data use requests that match two to ten objects, provided
annotated attributes distinguish every candidate. The three scripted
turns supervise all matching center--viewpoint pairs, descriptions of
the candidates' differences, and the pair identified by a sampled
letter choice. Each training example supplies the letter and coordinates
as the prefix for each candidate's description. The model learns to complete
the line. Each distinct first question counts once per region, across
all eligible memory starts and later-turn variations.

\subsection{Rendering start views and generating routes}
\label{app:training_routes}
The offline pipeline plans a \emph{source path} from a navigable start
to a reachable viewpoint of an annotated object. The path supplies
camera poses for rendering. Views of the target can also support
instructions about its appearance, its relation to nearby objects, or
a user's purpose. After selecting a memory, the training sampler generates
a \emph{reference route}, the waypoint sequence the model learns to predict.

To vary the view of the same ground path, we keep each source pose's
horizontal position and heading and vary the camera height from 0.4 to
2.0\,m above the local ground, pitch from $30^\circ$ downward to
$15^\circ$ upward, and roll by up to $30^\circ$ in either direction.
We store each rendered RGB frame with its augmented pose.

When preparing navigation records, we check the source path endpoints
against the recorded camera positions and associate the episode with
eligible region memories. For localization, we render
source frames at $1080{\times}640$ pixels (width by height). During
training, we select a frame at random and draw landscape or portrait
crops at different aspect ratios and output resolutions.
Cropping changes the field of view while preserving the camera pose
used as the localization target. We adjust the camera intrinsics for
the crop and any subsequent resizing. To generate an absent-view example,
we keep the memory fixed and draw an image from another scene, or from
a different region in the same scene if its estimated overlap with the
memory is below 20\%.

For coordinate navigation, the sampler selects a recorded start view and
samples reachable goals in several path-length ranges. It searches an
eight-neighbor grid of free space observed in the selected memory,
requiring at least 0.30\,m obstacle clearance and disallowing diagonal
corner cutting. The sampler shortens the path wherever a direct segment
is collision-free, resamples it to the $n+1$ waypoints of
\cref{eq:scene_planner}, and checks the resulting segments again.
If it exhausts the candidate starts and goals, it uses the original
start and its archived path together.

To train for goal-directed exploration with partial memory (\cref{sec:cold_start}),
we remove the blocks that cover the goal and retain a
connected set of blocks that covers the start when possible. We trace
the source path toward the goal to find where it first leaves observed
free space and continues through unknown space for the required minimum
distance. A nearby frontier with sufficient clearance becomes the
exploration target. A paired coordinate query uses the same partial
memory and starting view to supervise a route to that frontier.

\subsection{From records to training batches}
\label{app:training_batches}
In the earlier joint-training reference configuration, we sample memory
on a 0.32\,m voxel grid. Each retained point stays paired with its
visual feature and any available geometric target. For each patch that
can be matched to a source pixel with valid reference depth, we
back-project that pixel with the camera intrinsics and pose. The
resulting 3D point supervises the surface-position head
in \cref{sec:memory_construction}.

Training varies the memory-image resolution. For each draw, the sampler
keeps the blocks separate with probability 40\%, merges connected groups
with probability 40\%, or merges the full memory with probability 20\%.
We center each block on its retained points, rotate its local
coordinates, and shuffle the block order. The encoder receives these
local coordinates and their visual features. The first block defines
$F_1$. Recorded transforms express the center, observation-viewpoint,
camera-pose, surface-point, and route targets in that frame.

For the decoder adapters described in \cref{sec:scene_planner}, the
LoRA rank is 16 for vision-language tasks and 32 for trajectory
prediction. The language loss supervises both words and the special
tokens that mark geometric outputs. Position and orientation losses
supervise object centers, camera poses, relative block poses, and
surface points wherever the corresponding labels are available.

For each valid observation viewpoint, we express the target position
in the reference frame. To express the heading, we transform its
world-horizontal direction into that frame, project it onto the local
horizontal plane, and normalize the projection. We use axis-weighted smooth L1 for
position and squared error for the heading's cosine and sine.

Several queries can use one memory, so we count the block-pose and
surface-point losses once per memory. In each group of eight parallel
training workers, seven draw 16 queries per memory from the main task
mixture. The eighth processes two batches of paired frontier and route
examples per parameter update. We normalize each group's losses by its
valid targets before combining them with the training weights.

The flow head projects the decoder's query features $H$ to width 256.
In each of four transformer layers, eight attention heads let the
waypoint states attend to one another and cross-attend to the query
features. Each state also receives its waypoint index, a flow-time
embedding, and an embedding of the start, goal, and their displacement.

For training, we use the waypoint parameterization of \cref{eq:trajectory_flow}
to express the reference route as dimensionless offsets $Y$ from the
straight line joining the start and goal. The $k$th row is
$[Y]_k=(\mathbf p^\star_k-\mathbf c_k)/\ell$ for
$k=1,\ldots,n-1$. We draw an initial offset sequence $Z$ of the same
shape, with a smooth component and a small independent perturbation,
tapered toward both endpoints. At flow time $t$ sampled uniformly from
$[0,1]$, the head takes the interpolated offsets $(1-t)Z+tY$ and learns
their velocity $Y-Z$. We multiply the mean squared velocity error by
$\ell^2$ to express the loss in squared meters. At inference, the
Euler updates in \cref{eq:trajectory_flow} start from zero offsets.

We train the local planner separately on windows from reference
trajectories. For each eligible path, we sample one memory and cache
the route predicted by a frozen navigation transformer from an earlier
50,000-update checkpoint. At each decision step, we pair a window of
the predicted route with the RGB observations available through that
step. We keep windows where the prediction and reference agree in local
position and direction. The planner thus encounters route prediction
errors while receiving an action label consistent with its route input.

Future poses along each reference trajectory supervise the local planner's
ten-step horizon in the robot's frame. Forward
motion is sampled at 0.25\,m intervals, retaining a shorter final
interval, and in-place turns receive their own action labels. An
arrival label is valid only for source trajectories whose endpoint is a
known task goal. We train on valid waypoint, movement, and arrival targets with
the losses in \cref{eq:local_loss}. A separate training variant pairs a
language prompt with examples recorded before and after a recovery
maneuver.

\section{Evaluation protocols}
\label{sec:eval_protocols}
We first describe the sites on which we evaluate navigation, then give
the scoring rules for the two benchmarks, the protocols for varying
memory and exploration, and the offline tests of individual model outputs.

\Cref{tab:capture_statistics} records the composition and physical extent
of \datasetname{}. Its 129 institutional sites include 93 library and
education sites and 36 cultural exhibition sites. Its 142 commercial
sites include 81 retail and consumer service venues and 61 dining and
banquet venues. Each of the five evaluation sites larger than 500\,m$^2$
contributed 4,836--9,910 registered color views. Across all
\releaseEvalScenes{} evaluation sites, 199,917 views were used for fitting
and 10,539 for checking the reconstructions. The area totals in panel~(c)
use each collection's own definition, which ranges from the area of free
floor cells to the convex hull of navigable positions.

\suppressfloats[t]
\begin{table}[htbp]
\centering\small
\caption{\datasetname{} scene coverage and evaluation inventory. The
evaluation sites in (a) are held out from the \releaseTrainScenes{}
public training scenes. The RGB-view counts in (b) include views
from each camera separately. A repeat names a target requested
earlier in the episode, regardless of whether the robot saw it.
Panel~(c) uses each source's definition of area.}
\label{tab:capture_statistics}
\begin{tabular*}{\linewidth}{@{\extracolsep{\fill}}lrrrrrrrr@{}}
\toprule
\multicolumn{9}{@{}l}{\textbf{(a) Scene records by domain}}\\[2pt]
Inventory & Institutional & Commercial & Public & Office & Industrial & Recreation & Other & Total\\
\midrule
Public release & 129 & 142 & 86 & 67 & 58 & 28 & 1 & \releaseScenes{}\\
Held-out sites & 3 & 11 & 11 & 7 & 0 & 4 & 0 & \releaseEvalScenes{}\\
\bottomrule
\end{tabular*}
\par\smallskip
\begin{tabularx}{\linewidth}{@{}LLr@{}}
\toprule
\multicolumn{3}{@{}l}{\textbf{(b) \datasetname{} evaluation inventory}}\\[2pt]
Property & Reporting unit & Value\\
\midrule
Scene inventory & Scenes / reconstructions & \releaseEvalScenes{} / 36\\
Independent physical sites & Distinct sites after grouping rescans & 36\\
Capture effort & Source recordings / registered captured RGB views & 36 / 210,456\\
Views per reconstruction & Median (range), counting each camera separately & 5,785 (970--9,910)\\
Benchmark tasks & Episodes / requests & \releaseBenchmarkEpisodes\,/\,\releaseBenchmarkRequests\\
Relocation extension & Episodes / requests & 150\,/\,\relocationRequests\\
Repeat requests & Relocation requests naming an earlier target / all relocation requests & 225\,/\,\relocationRequests\\
Goal modalities & Category / description / reference-image requests across both evaluations & 0 / 2,602 / 0\\
\bottomrule
\end{tabularx}
\par\smallskip
\begin{tabularx}{\linewidth}{@{}>{\raggedright\arraybackslash}p{.25\linewidth}Lr@{}}
\toprule
\multicolumn{3}{@{}l}{\textbf{(c) Physical coverage of real captures, by source definition}}\\[2pt]
Collection & Definition of area & Total over scene records\\
\midrule
ScanNet++~\cite{scannetpp} (ICCV 2023 study) & Captured floor area (460 scenes) & $\sim$15,000\,m$^2$\\
HM3D~\cite{hm3d} & Per-floor convex hull of navigable positions, summed over floors (1,000 scans) & 365,420\,m$^2$\\
& Area of the navigation mesh for a 0.10\,m radius, 1.50\,m tall robot (same scans) & 112,500\,m$^2$\\
\datasetname{} (public release) & Free cells on the final occupancy maps, projected horizontally (\releaseScenes{} scenes) & 103,523\,m$^2$\\
\datasetname{} (evaluation) & Cells marked free on the 0.15\,m occupancy grid before allowing for robot clearance (36 reconstructions) & 9,007\,m$^2$\\
\bottomrule
\end{tabularx}
\end{table}

\subsection{Benchmark scoring and execution}
\label{sec:benchmark_scoring}
For a successful request, success weighted by path length (SPL) divides
the reference shortest-path distance by the larger of that distance
and the distance traveled. The reference path ends at an annotated goal
viewpoint on GOAT-Bench and at a reachable target-mask cell on
\datasetname{}. A failed request
earns zero. For the standard \method runs in \cref{tab:standard_navigation},
we average subtask scores within each episode, then give scored episodes
equal weight. Each subset contains one episode per scene. An episode
terminated by an execution error stays in the denominator, with every
request in that episode scored as zero.

\paragraph{Execution shared by the two benchmarks.}
During initial exploration, \method integrates rendered depth at the
simulator camera poses to choose frontier viewpoints. The benchmark
pathfinder checks reachability and guides travel to the selected viewpoint. For
an object request, the model retrieves center--viewpoint pairs. In
the absence of a user reply, the controller selects the candidate with
the most memory points near its center, or the first model prediction
if point counts are unavailable. During replanning, the center nearest
the previously selected center in the common frame takes priority,
even when all candidates are more than 2\,m away. The chosen
viewpoint's position supplies the destination. Its heading does not
constrain the final camera orientation, and a missing viewpoint leaves
the pair without a usable destination.

For goal prediction, the controller expresses the destination relative
to its estimated start and places that displacement at the robot's
current position and heading. The reference pathfinder plans a path
to the resulting point. If the point is unreachable or the robot
stops making progress, the controller can search backward along the
predicted route for a reachable endpoint and then replan. The pathfinder
still determines the path traveled. In the action-generation setting,
the local planner follows the predicted route. A failed retrieval can
trigger goal-directed exploration in either setting. During periodic
replanning, rendered depth can rank viewing directions for localization,
independently of the depth source for collision recovery.

On arrival, \method turns through $360^\circ$ to collect fresh views
and then stops in the standard evaluations. Relocation episodes can
trigger the pre-stop recheck described below. A consecutive request
identical in type and content normally stops in place. On relocation
episodes this shortcut requires the same active scene state, and a
memory-triggered retry always bypasses it. A malformed model output
also prompts a stop where the robot stands. If localization attempts
cannot initialize the odometry-to-memory alignment, the controller
submits stops for the current and remaining requests. The evaluator
scores all such stops at their actual locations.

\paragraph{GOAT-Bench.}
We use the released
\href{https://github.com/Ram81/goat-bench/blob/74c41d19d4a4c3608d1575b512087b5a529aee0e/config/tasks/goat_stretch_hm3d.yaml\#L49-L59}{task configuration},
which sets the success distance to 0.25\,m and specifies annotated
viewpoints as the goals. The
\href{https://github.com/Ram81/goat-bench/blob/74c41d19d4a4c3608d1575b512087b5a529aee0e/goat_bench/measurements/nav.py\#L195-L203}{distance measure}
computes the shortest navigable distance to those viewpoints. The
\href{https://github.com/Ram81/goat-bench/blob/74c41d19d4a4c3608d1575b512087b5a529aee0e/goat_bench/measurements/nav.py\#L285-L297}{success evaluator}
requires the robot to declare a subtask stop strictly below that
threshold. The authors' original
\href{https://github.com/Ram81/goat-bench/blob/18ed15903350f8450de8151e7c80b19b9798fa67/scripts/eval/2-goat-eval.sh\#L53-L61}{monolithic}
and \href{https://github.com/Ram81/goat-bench/blob/18ed15903350f8450de8151e7c80b19b9798fa67/scripts/eval/1-goat-individual-policy-chaining.sh\#L50-L58}{skill-chain}
evaluation scripts explicitly set the same tolerance. The
\href{https://github.com/Ram81/goat-bench/blob/05f225fedb0b9d485d169623e1fdb9a6cfccced5/goat/dataset/objectnav_generator.py\#L968-L982}{viewpoint generator}
uses a separate 1\,m limit when
\href{https://github.com/Ram81/goat-bench/blob/05f225fedb0b9d485d169623e1fdb9a6cfccced5/goat/dataset/objectnav_generator.py\#L229-L266}{screening candidates around an object}.
For the full GOAT-Bench split, we quote the SenseAct-NN and Modular
GOAT results from Table~2 of Khanna et al.~\cite{goatbench}. The
published MSGNav scores come from the GPT-4o row of supplementary
Table~4 in Huang et al.~\cite{msgnav}, covering 2,669 requests.
The shared subset selects episode ID
zero from each of the 36 Val-Unseen scenes, with 278 requests (99
category, 91 description, and 88 reference-image goals). The full split
has 360 episodes and 2,669 requests. The six open-source agents retain
their native cameras, scan policies, and internal limits. We score the
first 5,000 recorded actions per episode at most and average SR and SPL
over episodes with equal weight.

The GOAT-Bench rerun for AstraNav-Memory~\cite{astranavmemory}
configures $720{\times}640$ color and $360{\times}640$ depth images
(width by height), with a $110^\circ$ horizontal field of view at
1.31\,m above the floor. Its released policy builds memory from RGB
views and their simulator poses.
At the same height, MTU3D~\cite{mtu3d} uses $360{\times}640$
color and depth images with a $42^\circ$ horizontal field of view and
scans in twelve $30^\circ$ increments. The four GPT-4o methods render
$1280{\times}1280$ color and depth images with a $120^\circ$ horizontal
field of view at 1.5\,m. MSGNav~\cite{msgnav} resizes the model input
to $512{\times}512$. SCOPE~\cite{scope}, HGR~\cite{hgr}, and
3D-Mem~\cite{threedmem} resize it to $360{\times}360$.

The published SenseAct-NN policies use RGB and simulator pose, and
Modular GOAT uses simulator depth and pose. For GPT-6 Astra through
NavHarness~\cite{navharness}, our shared-subset rerun uses the released
0.25\,m viewpoint criterion.
The framework estimates pose with RGB-D ORB-SLAM3 and executes the
movements the model selects. For the published full-split result, the NavHarness
\href{https://github.com/billzhao1030/NavHarness/blob/9a4c12f820631b6a17576651180a5cbddd079264/exp_workspace/lhnav_es/configs/std_goat_slam_es_lhnav.yaml\#L56-L62}{configuration}
sets the goal-viewpoint threshold to 1\,m and its
\href{https://github.com/billzhao1030/NavHarness/blob/9a4c12f820631b6a17576651180a5cbddd079264/thirdparty/EmbodiedScore-envs/embodiedscore_envs/benchmarks/env/metrics.py\#L184-L193}{evaluator}
applies it at the declared stop.

The three \method goal-prediction runs use $640{\times}480$ RGB-D views
with a $90^\circ$ horizontal field of view. During initial exploration,
Habitat's NavMesh supplies reachability and travel between selected
frontiers, with up to 200 exploration actions charged to the first
subtask's 500-step budget. During goal prediction, Habitat projects each
requested endpoint onto its reachable mesh and uses its pathfinder and
greedy follower to move there. The motion filter can shorten or reject
a step at the mesh boundary without sliding the robot sideways.

For an image goal, the simulator renders a $512{\times}512$ reference
view at the benchmark-specified pose and horizontal field of view. The
model localizes this photograph in memory. A successful localization
places the goal 0.75\,m from the estimated camera position along its
horizontal viewing direction.

On GOAT-Bench, all three \method geometry configurations use the active
block selection and odometry alignment in \cref{sec:cold_start}. On
\datasetname{}, the simulator-geometry runs use the same block
selection. With either LingBot-Map backend the controller ordinarily
keeps the first block from initial exploration as its reference and reads
the full episode memory. The backend can therefore change which
observations the model reads on \datasetname{}.

\paragraph{\datasetname{} tasks.}
The benchmark robot has a radius of 0.17\,m, a body height of 1.41\,m,
and a camera height of 1.25\,m. For each target, a two-dimensional mask
marks the free cells in which the robot can stop successfully. Its
construction combines the distance
criterion with target visibility sampled at heights from 0 to 3\,m in
0.05\,m increments. At an explicit stop the evaluator tests the cell
containing the robot's horizontal position. The mask determines success
independently of the robot's final camera height and heading. The
shortest path for SPL ends at the nearest reachable mask cell on the
reference occupancy grid. The oracle success rate uses the same mask
and counts a visit anywhere along the recorded subtask path, including
its starting position.

The six external goal predictors use pretrained checkpoints or
off-the-shelf models without training on the \datasetname{} split. The
occupancy-grid follower executes their chosen destinations.
For the GPT-6 Astra subset run in \cref{tab:standard_navigation},
NavHarness provides a map and poses derived from color and
benchmark-rendered depth. GPT-6 Astra uses those tools to select
movements.

For \method the reference occupancy grid supports initial exploration,
which has its own 1,000-action budget. Subsequent standard requests allow
500 actions each. During goal prediction the same grid supplies the
path to each selected endpoint. In action generation the local planner
follows the predicted route. If a recorded start lies outside free
space, the loader moves its horizontal position to the nearest free cell.

For the standard \datasetname{} subset, we select the first episode
listed at each site in version 1.4 of the task list. All six \method
configurations therefore cover the same 36 episodes and 156 requests.
The full-split scores in \cref{tab:standard_navigation,tab:secondary_navigation_metrics}
come from the earlier 360-episode evaluation.

\Cref{tab:secondary_navigation_metrics} reports SoftSPL and whole-episode
success for the standard \method runs. SoftSPL weights the fractional
reduction in distance to the goal (clipped at zero) by the SPL
path-efficiency factor. A timeout earns zero. An episode succeeds
only if every request succeeds. With LingBot-Map~2.0 and the local planner, full-split subtask and
episode SR are \goatAssistedMapTwoFullSR{}\% and 5.28\% on GOAT-Bench,
and \lingbotAssistedMapTwoFullSR{}\% and 0.56\% on \datasetname{}.

\begin{table}[!t]
\centering\small
\caption{SoftSPL and whole-episode success for the standard \method
runs in \cref{tab:standard_navigation}. SoftSPL credits progress toward
each goal, weighted by path efficiency. Episode SR is the percentage of
episodes in which every request succeeds. Each full run covers 360 episodes,
and each subset contains one episode per scene (36 episodes). On
\datasetname{}, the subset scores use the v1.4 runs and the full-split
scores use the earlier runs in \cref{tab:standard_navigation}. All scores are percentages.}
\label{tab:secondary_navigation_metrics}
\begin{tabular*}{\linewidth}{@{\extracolsep{\fill}}llrrrr@{}}
\toprule
& & \multicolumn{2}{c}{Subset} & \multicolumn{2}{c}{Full split}\\
\cmidrule(lr){3-4}\cmidrule(l){5-6}
Motion during requests & Memory geometry & SoftSPL & Episode SR & SoftSPL & Episode SR\\
\midrule
\multicolumn{6}{@{}l}{\textit{GOAT-Bench, initial exploration with Habitat pathfinding}}\\
Simulator pathfinding & Simulator depth and pose & \goatGoalSubsetSoftSPL & \goatGoalSubsetEpisodeSR & 33.64 & 13.33\\
& LingBot-Map 1.0 & 31.85 & 11.11 & 32.62 & 11.94\\
& LingBot-Map 2.0 & 31.90 & 16.67 & 33.60 & 10.83\\
\addlinespace[2pt]
Local planner & Simulator depth and pose & 23.94 & 13.89 & 25.12 & 7.50\\
& LingBot-Map 1.0 & 20.17 & 5.56 & 21.29 & 3.06\\
& LingBot-Map 2.0 & 21.27 & 8.33 & 21.90 & 5.28\\
\midrule
\multicolumn{6}{@{}l}{\textit{\datasetname{}, initial exploration with reference-grid pathfinding}}\\
Simulator pathfinding & Simulator depth and pose & 38.87 & 11.11 & 40.96 & 8.06\\
& LingBot-Map 1.0 & 36.65 & 5.56 & 38.57 & 7.22\\
& LingBot-Map 2.0 & 38.78 & 5.56 & 40.17 & 7.50\\
\addlinespace[2pt]
Local planner & Simulator depth and pose & 26.34 & 0.00 & 27.25 & 1.11\\
& LingBot-Map 1.0 & 18.80 & 2.78 & 22.09 & 0.56\\
& LingBot-Map 2.0 & 20.86 & 0.00 & 23.75 & 0.56\\
\bottomrule
\end{tabular*}
\end{table}

For the NavHarness tolerance sweep, the percentages and pooled success
counts are reported separately (\cref{tab:navharness_tolerance}). The
aggregate results do not give the episode breakdown needed to reproduce
the percentages from those counts.

\begin{table}[!t]
\centering\small
\caption{Stopping tolerance for GPT-6 Astra (NavHarness) on the
\datasetname{} subset. All rows rescore the same trajectories from
36 episodes and 156 requests. At zero extra tolerance, the standard target mask gives the SR and SPL
reported in \cref{tab:standard_navigation}. Positive tolerances extend its boundary by the indicated
shortest-path distance through free grid cells. Oracle SR credits a
visit anywhere along the path. The last column pools successful
requests across episodes.}
\label{tab:navharness_tolerance}
\begin{tabular}{@{}rrrrr@{}}
\toprule
Extra tolerance (m) & SR (\%) & SPL (\%) & Oracle SR (\%) & Successful requests\\
\midrule
0.00 & 26.67 & 17.74 & 29.17 & 41/156\\
0.25 & 33.33 & 22.11 & 35.83 & 51/156\\
0.50 & 48.33 & 31.12 & 50.14 & 74/156\\
1.00 & 66.25 & 39.57 & 68.61 & 102/156\\
1.50 & 72.78 & 40.14 & 75.14 & 113/156\\
\bottomrule
\end{tabular}
\end{table}

\paragraph{Object relocation between requests.}
Each relocation episode supplies an ordered list of requests and
placement changes. When the designated request ends by an explicit
stop or timeout, the evaluator scores it and commits its observations
to memory before applying the change. Color and depth rendering then
use the new object placements. Collision checking updates the occupied
cells using each object's footprint and the robot's clearance. A change
that would place an object over the robot is rejected. The robot
retains its pose and memory into the next request. The placement
schedule and evaluation masks remain private to the evaluator.
Relocation subsets use a fixed ranking of the episode identifiers.

A stop succeeds if the robot occupies a free cell in the applicable
target mask. For movable targets the evaluator uses the published mask
for the active placement. Fixed targets use their static masks, also
restricted to cells that remain free after the move. Distances for SPL
are computed on the active occupancy grid to valid goal cells. Oracle
success counts any recorded visit to such a cell during that request.

All three relocation policies in \cref{tab:object_relocation} use the
same target identities, placement schedule, simulator geometry and
occupancy-grid pathfinding. Disabling arrival rechecks leaves ordinary
memory collection in place. When a recheck is enabled, a qualifying
nearby match can trigger pruning and one retry (\cref{sec:cold_start}).
Initial exploration has a budget of 1,000 actions, and each request has
a budget of 2,000. A retry retains both the step count and the distance traveled.

The 150 episodes contain \relocationRequests{} requests, including
655 for fixed targets and 225 at each placement for movable targets.
A request counts as after-move once the evaluator changes its target's
placement, regardless of whether the robot previously saw it. For each
group we average within episodes and then across episodes containing
at least one request in that group.
All three policies have scores for 150 episodes. The all-target run
had one execution error, scored as zero, and the disabled-recheck run
completed all 150 normally. The updated mobility-prior scorecard
reports the number scored without a separate count of normal completions. No policy completed every request in any episode. The
three policies were evaluated with different software revisions.

\subsection{Memory retention and initial exploration}
\label{sec:paired_protocol}

\Cref{tab:memory_carryover}a uses the first episode in
each GOAT-Bench scene. The carryover reference is the goal-prediction
run with simulator depth and pose in \cref{tab:standard_navigation}.
The short-history and prebuilt-memory runs cover the same 36 episodes
and 278 requests. They use different software revisions, and the
available records do not establish that their geometry and motion
settings match the reference.

Panel (b) uses the standard-budget controls from the full-split
exploration experiments. Both controls were run separately from the
full-split results in \cref{tab:standard_navigation}. On \datasetname{},
the 36-episode subset in that table was drawn from the same run as
the 360-episode standard-budget control here. Both round to 47.78\% SR. Their SPLs are 26.35\% and 27.57\%, respectively. GOAT-Bench contains 2,669 requests and
\datasetname{} v1.4 contains 1,497. Both experiments use simulator
depth and pose to build memory, reference pathfinding to execute
motion, and a 500-action budget per request. The separate limit on initial
exploration applies once at episode start. Exploration can finish before
reaching that limit. GOAT-Bench counts
initial exploration toward the first request's budget. On \datasetname{}
it has a separate allowance, so the first request still has all 500
actions available. In both cases the robot retains the observations
it acquires and can continue to explore during later requests.

\subsection{Model diagnostics}
\label{sec:diagnostic_protocol}

\paragraph{Grounding, localization, and route errors.}
The offline ablations hold requests, target annotations, and query images
fixed within each dataset. For grounding, we put each predicted and annotated
object center in world coordinates using the reference pose of its
memory frame. A prediction can match an annotation within a 3D distance of
1\,m. We maximize the number of one-to-one matches,
then minimize their total distance. We pool the matches across questions.
Precision (Prec.) is the percentage of predicted centers that match a target,
and recall (Rec.) is the percentage of target centers recovered. Questions with no annotated target remain
in the evaluation and contribute false positives if the model predicts
an object. This metric scores instance centers. The location and visibility
of a proposed observation viewpoint are separate quantities.

For localization, the first turn of a coordinate-navigation request
returns a camera pose. Global position error (GPos) is the distance between the estimated and
reference camera centers in a common metric frame. For the block-corrected errors,
we express the predicted camera pose in its associated reconstruction
block and map it back using the reference pose of that block. Comparing
this corrected pose with the annotated pose gives the position error
(LPos, meters) and rotation error (LRot, degrees).

For each predicted route, we express the reference waypoints in the same
metric frame and pair them by index as in \cref{eq:scene_planner}.
Average displacement error (ADE) is the mean 3D Euclidean distance
between paired waypoints,
\begin{equation}
 \operatorname{ADE} =
 \frac{1}{n+1}\sum_{k=0}^{n}
 \bigl\|\widehat{\mathbf p}_k-\mathbf p^\star_k\bigr\|_2 .
 \label{eq:route_ade}
\end{equation}
This metric includes both endpoints. In the flow-matching formulation
(\cref{sec:scene_planner}), the start and goal stay fixed while the
head updates the $n-1$ interior waypoints.
Each pose or route mean includes only requests that produced a valid
output for that metric. A valid pose counts toward the localization scores
even if the request yields no valid route.

\paragraph{Scene selection.}
For the \datasetname{} ablations, we rank the 151 available scenes by the
mean pairwise block-translation error of the offline reference model using separate
LingBot-Map 1.0 blocks. We retain the 136 scenes with the smallest errors
and use their requests for every configuration, including LingBot-Map 2.0.
Each scene contributes one value to the ranking regardless of its number
of requests. The averages in \cref{tab:memory_inference,tab:memory_training}
therefore describe performance on a subset selected for relatively good
baseline registration. This set contains 8,472 grounding requests and
1,669 coordinate-navigation requests. The closed-loop \datasetname{}
benchmark uses a separate set of \releaseEvalScenes{} held-out sites.
For HM3D, all configurations use the full set of offline diagnostic
requests without registration-based filtering. GOAT-Bench navigation
episodes on HM3D are evaluated separately (\cref{sec:benchmark_comparisons}).

\section{Navigation model prompt templates}
\label{app:prompts}
The controller translates a user's request into the task prompts in
\cref{tab:task_interfaces}. The first user turn includes the instruction
\texttt{Coordinates are in the first memory block frame.} This is the
frame $F_1$ of the reference block (\cref{sec:memory_construction}).
For localization, the query image comes before this instruction.
Monospaced text gives the prompt wording. Italicized fields in braces
stand for request-specific values.

\begingroup
\newcommand{\promptfield}[1]{\textnormal{\itshape\{#1\}}}
\newcommand{\taskprompt}[1]{\par{\small\ttfamily\raggedright\hyphenpenalty=10000\relax #1\par}}
\subsection{Retrieval, candidate choice, and spatial questions}
\label{app:retrieval_prompts}
A category name, a referring expression, and the text on a sign all use
the same retrieval prompt. Asking for all matches preserves the alternatives
that a later turn may need to distinguish.
\taskprompt{Find all objects matching the following description in the available scene observations.\par\smallskip
<description>\\
\promptfield{description}\\
</description>\\
Return objects in ascending center-coordinate order (x, then y, then z). Coordinate tuples are in meters.}
For sign text, the description takes the form
\texttt{A target labeled with the text "\promptfield{sign text}"}.
For each match, the reply emits a center token immediately followed
by its observation-viewpoint token. The pairs appear in the coordinate
order specified above. A no-match reply contains neither token.

When retrieval returns several matches, the controller issues the
following comparison prompt on the user's behalf.
\taskprompt{Please describe the differences between these objects.}
For each retrieved center, the system supplies a letter and its coordinates
as a prefix. The model completes each line with a description drawn from spatial memory.
\Cref{fig:clarification_example} illustrates how these descriptions could
be presented together. The user's letter in the final turn identifies
the center and observation viewpoint that the decoder should return.
\taskprompt{I choose \promptfield{letter}.}

For ScanNet spatial question answering, the dataset question follows
the shared frame instruction directly. The decoder answers in text.

\subsection{Localization, exploration, and route prediction}
\label{app:route_prompts}
Localization uses the same prompt for two purposes. A live image gives
a visual pose estimate. The navigation controller uses reliable estimates
to align odometry with the active memory frame (\cref{sec:cold_start}).
For a reference photograph, the controller uses the estimated camera
pose to choose an approach point for the image goal.
\taskprompt{Where was this image taken in the available spatial memory? Return its pose.}

The controller uses the frontier prompt for goal-directed search after
initial exploration has built a memory (\cref{sec:cold_start}). If
retrieval finds no match, it retains the unresolved request and the
no-match reply. The follow-up gives the robot's current position from
odometry, expressed in $F_1$, and asks where to search.
\taskprompt{My current position is (\promptfield{x}, \promptfield{y}, \promptfield{z}). Which frontier should I explore next to find \promptfield{target}? Return its position.}
For a language request, the target field reads \texttt{the described target}.
For an image goal, it reads \texttt{the location shown in the image}, and
the model sees the reference photograph from the failed localization turn.
The position is written in meters to two decimal places. The reply gives
one frontier coordinate.

The route prompt follows the live-view localization turn. In offline
evaluation, the camera position predicted from the image supplies the
start. During navigation, the controller can fill the same pose in the
preceding turn using odometry aligned to memory. For an object goal,
the destination is the position of the selected observation viewpoint.
An approach point derived from a reference image, a predicted frontier,
or a supplied coordinate can also serve as the destination.
\taskprompt{Plan a route to (\promptfield{x}, \promptfield{y}, \promptfield{z}) meters in the first memory block coordinate frame.}
The controller writes the destination in meters to six decimal places.
The trajectory decoder returns the waypoint sequence in $F_1$
(\cref{eq:trajectory_flow}).
\endgroup

\FloatBarrier
\section{Qualitative case studies}
\label{sec:case_analysis}

These examples examine the predictions that precede a robot's next
movement. We first inspect the assembled memory, then the camera pose,
object matches, and route inferred from it. The last two cases show
what the model proposes when a request yields no match or several
plausible matches. \Cref{tab:memory_inference,tab:memory_training}
give aggregate results for grounding, localization, and route prediction.

\subsection{Aligning independently reconstructed memory blocks}
\label{sec:block_alignment_cases}
\begin{figure}[H]
\centering
\includegraphics[width=.90\linewidth,height=.69\textheight,keepaspectratio]{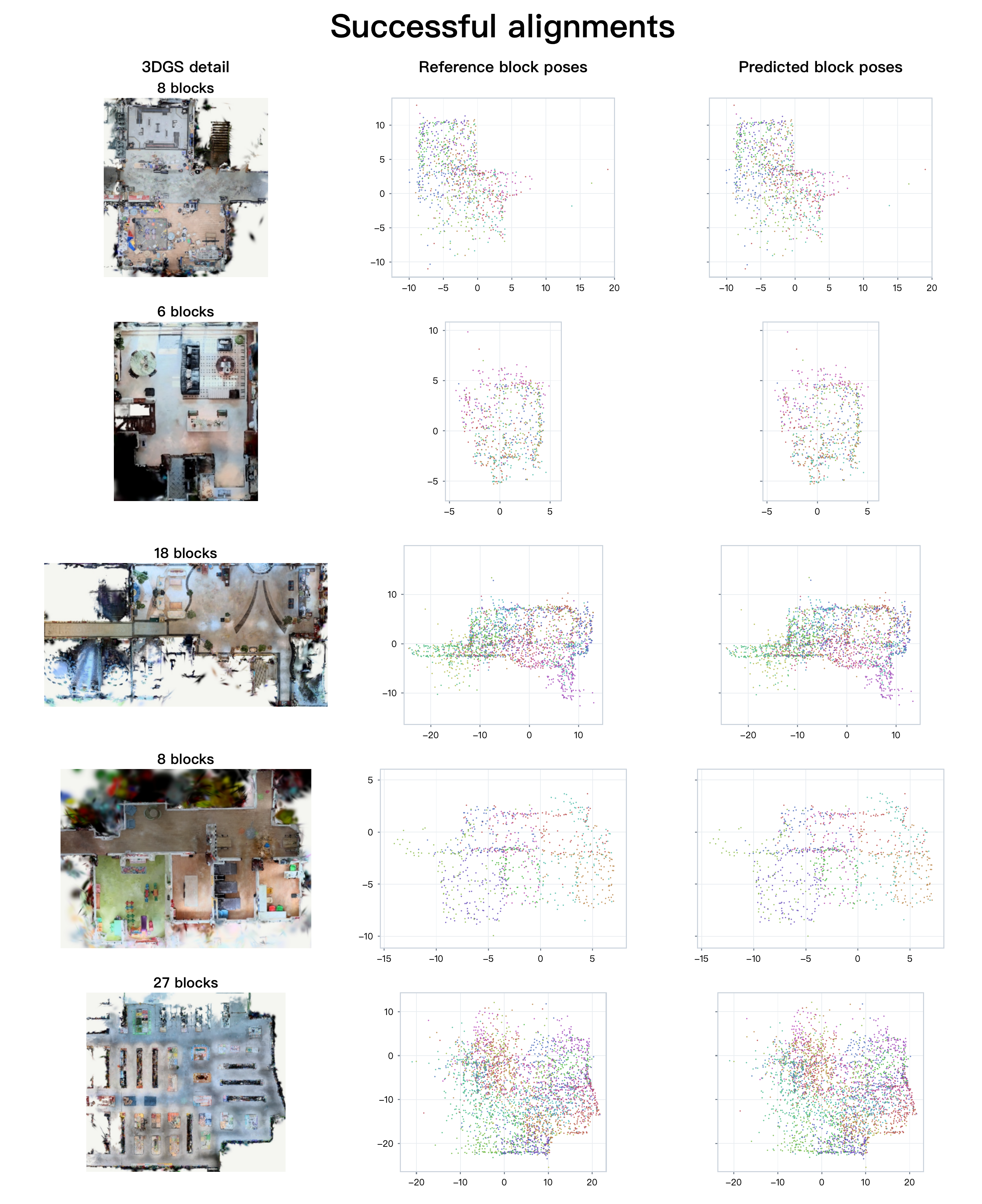}
\caption{Block alignment in five held-out \datasetname{} memories. Each row shows a
Gaussian rendering followed by maps assembled with the reference and
predicted poses. Point color identifies the block. Both maps use the
same scale and coordinate axes relative to the first block.}
\label{fig:block_alignment_good}
\end{figure}

In \cref{fig:block_alignment_good}, the reference and predicted maps
contain the same surface points. Only the positions and orientations
of the blocks differ. Room boundaries and connecting passages remain
recognizable even in the largest example, assembled from 27 blocks.

A room can look coherent in isolation and still be misplaced in the
assembled map. In the first row of \cref{fig:block_alignment_bad}, a
room-sized block is displaced relative to the rest of the scene. A
narrow overlap at a doorway would offer little shared surface on which
to estimate a relative pose, although the image alone does not establish
the cause. In the 74-block mall, the error changes the shape of a larger
region: the prediction preserves much of the layout near the reference
block but compresses part of the distant wing.

\begingroup
\setlength{\intextsep}{6pt}
\begin{figure}[H]
\centering
\includegraphics[width=.90\linewidth]{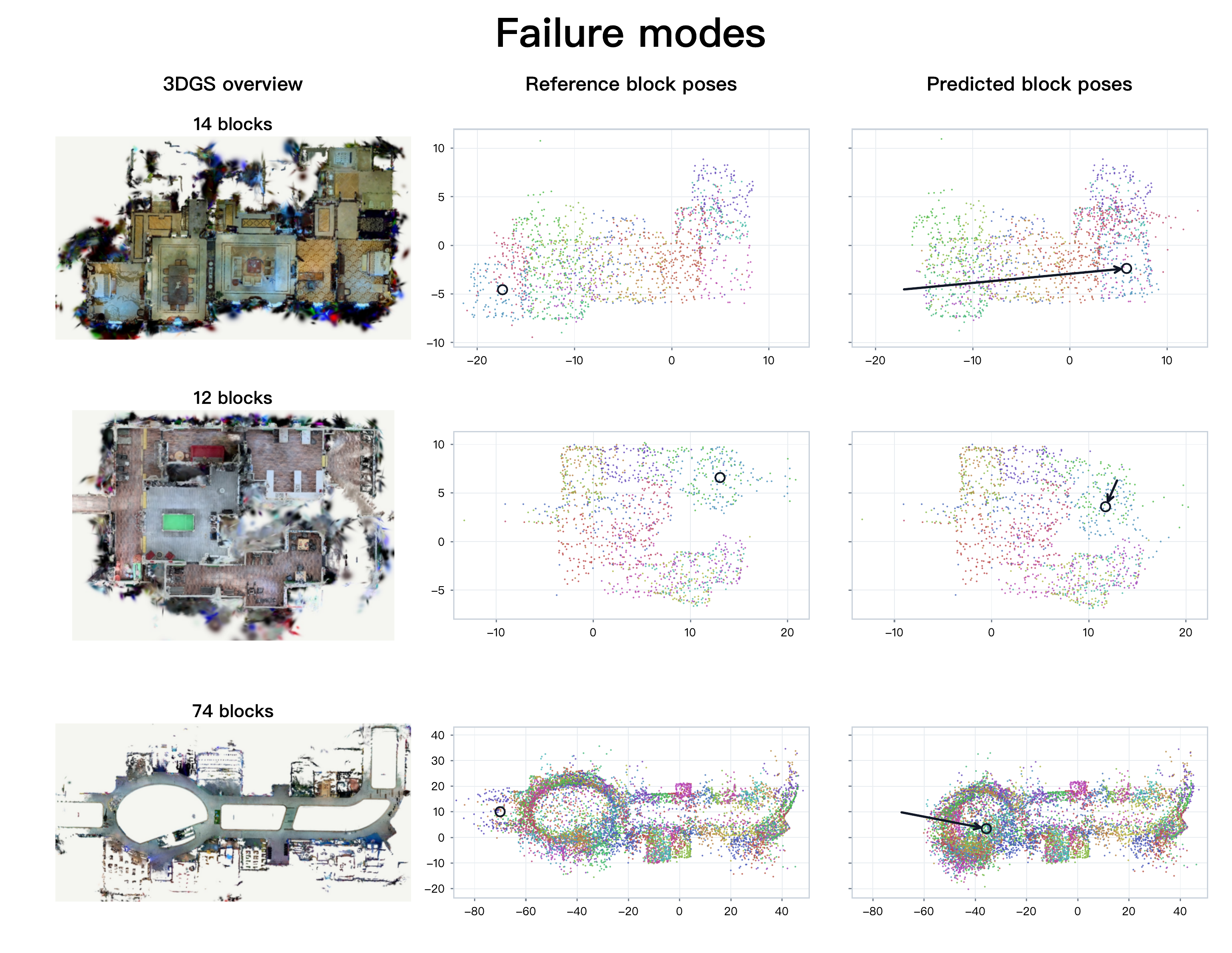}
\caption{Alignment errors in three held-out memories. In each row the
3DGS overview is followed by the reference and predicted layouts.
The circles mark the origins of the block with the largest 3D origin
error in each scene. The arrow in the predicted layout shows its
horizontal displacement from the reference. The mall is rotated to show
its long axis across the page.}
\label{fig:block_alignment_bad}
\end{figure}

\begin{figure}[H]
\centering
\includegraphics[width=.96\linewidth]{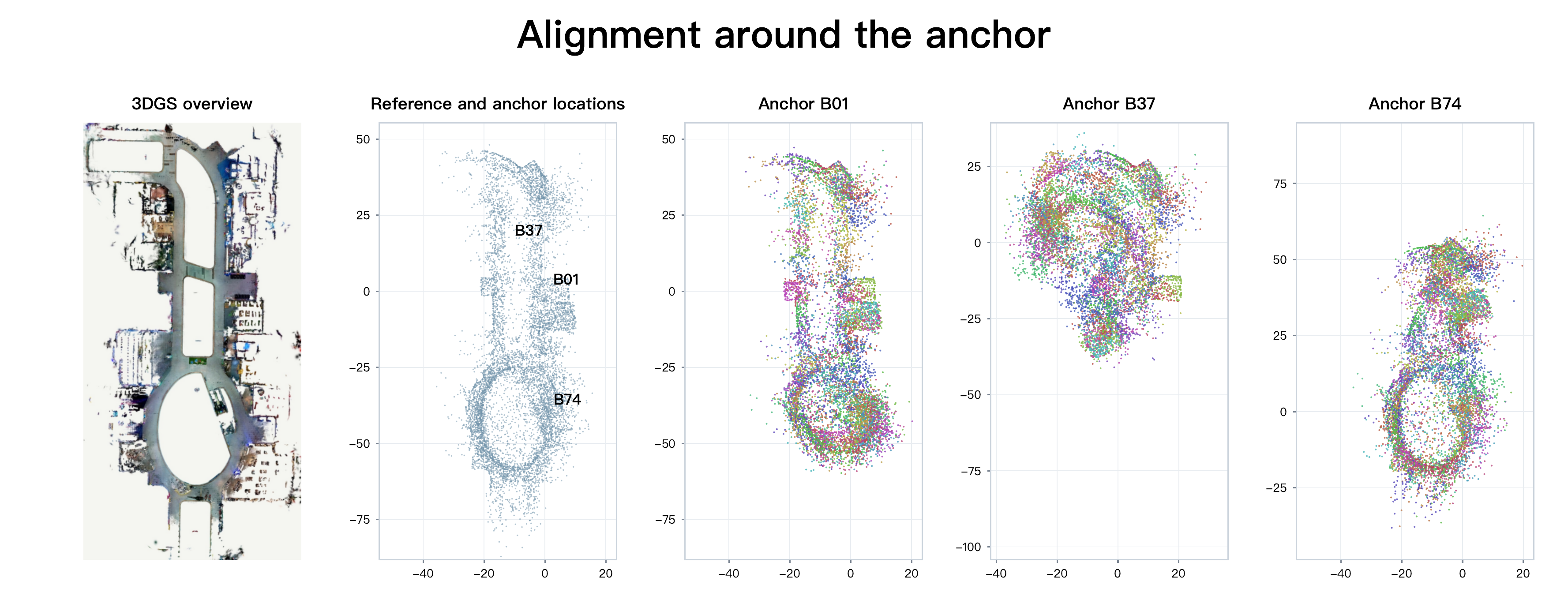}
\caption{Changing the reference block in one 74-block mall. From left to
right, the panels show the 3DGS overview, the reference map marking
B01, B37, and B74, and the three predicted layouts anchored at B01,
B37, and B74, respectively. Each predicted layout uses its anchor's
annotated pose. Axes show distances in meters from that anchor.
Block colors stay the same across predictions.}
\label{fig:block_alignment_anchors}
\end{figure}
\endgroup

Each marked block takes a turn as the first input block
(\cref{fig:block_alignment_anchors}). We rank the other 73 by distance
from the anchor in the reference map. In all three runs, the nearest
third has lower point-alignment error than the farthest.
\FloatBarrier

\subsection{Localizing a camera from an image}
\label{sec:localization_cases}

\begin{figure}[H]
\centering
\includegraphics[width=.90\linewidth]{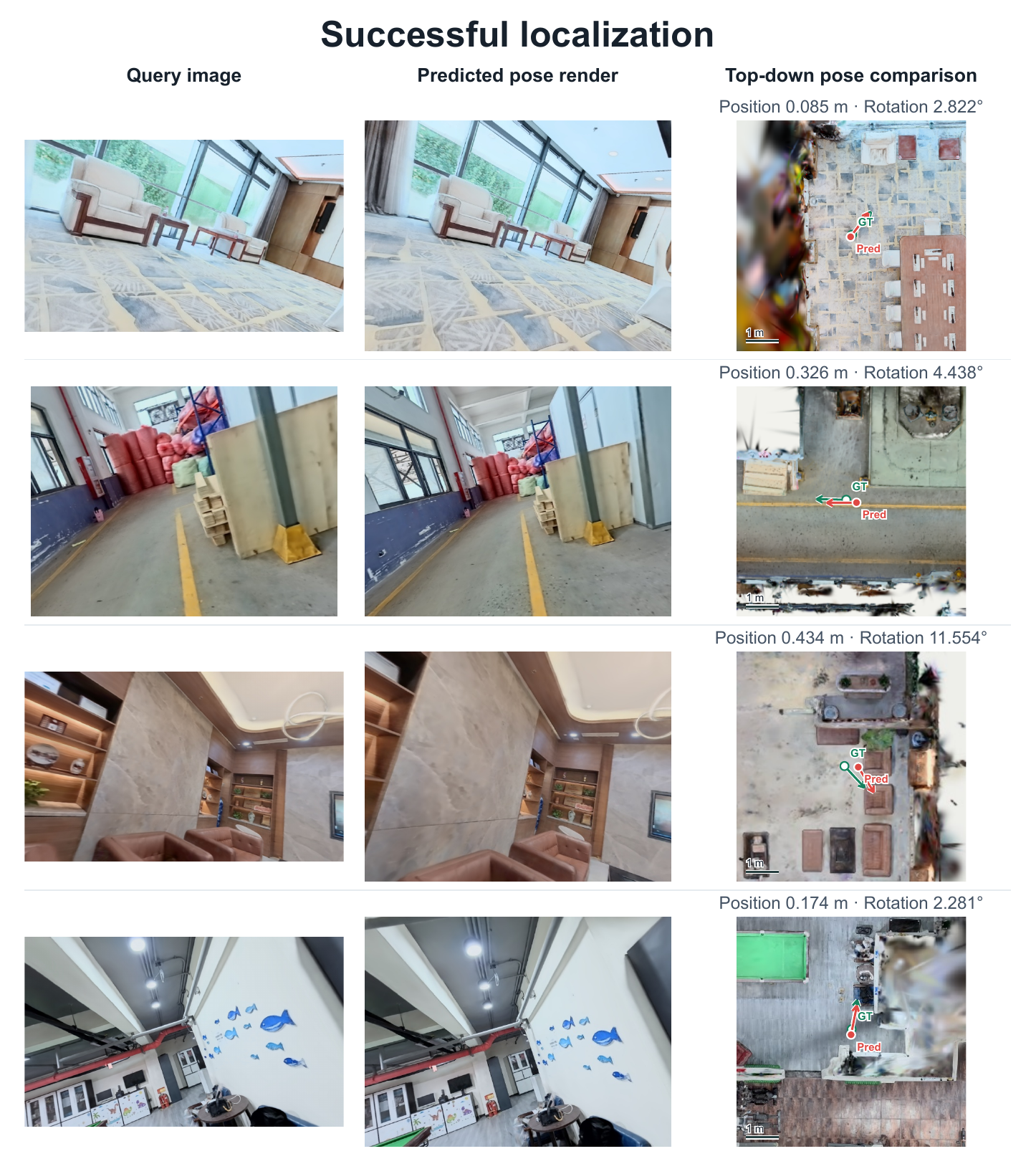}
\caption{Camera poses recovered from four query images. Each row shows the query, a view rendered at the predicted pose, and a top-down map. Green marks the ground-truth pose (GT) and red the predicted pose (Pred). Arrows show horizontal viewing directions. The labels report position and rotation errors.}
\label{fig:localization_good}
\end{figure}

Across the four queries in \cref{fig:localization_good}, camera
position error ranges from 0.085 to 0.434\,m. In the third example
this close positional agreement accompanies an orientation error of
about $12^\circ$.

\clearpage
\begin{figure}[H]
\centering
\includegraphics[width=.90\linewidth]{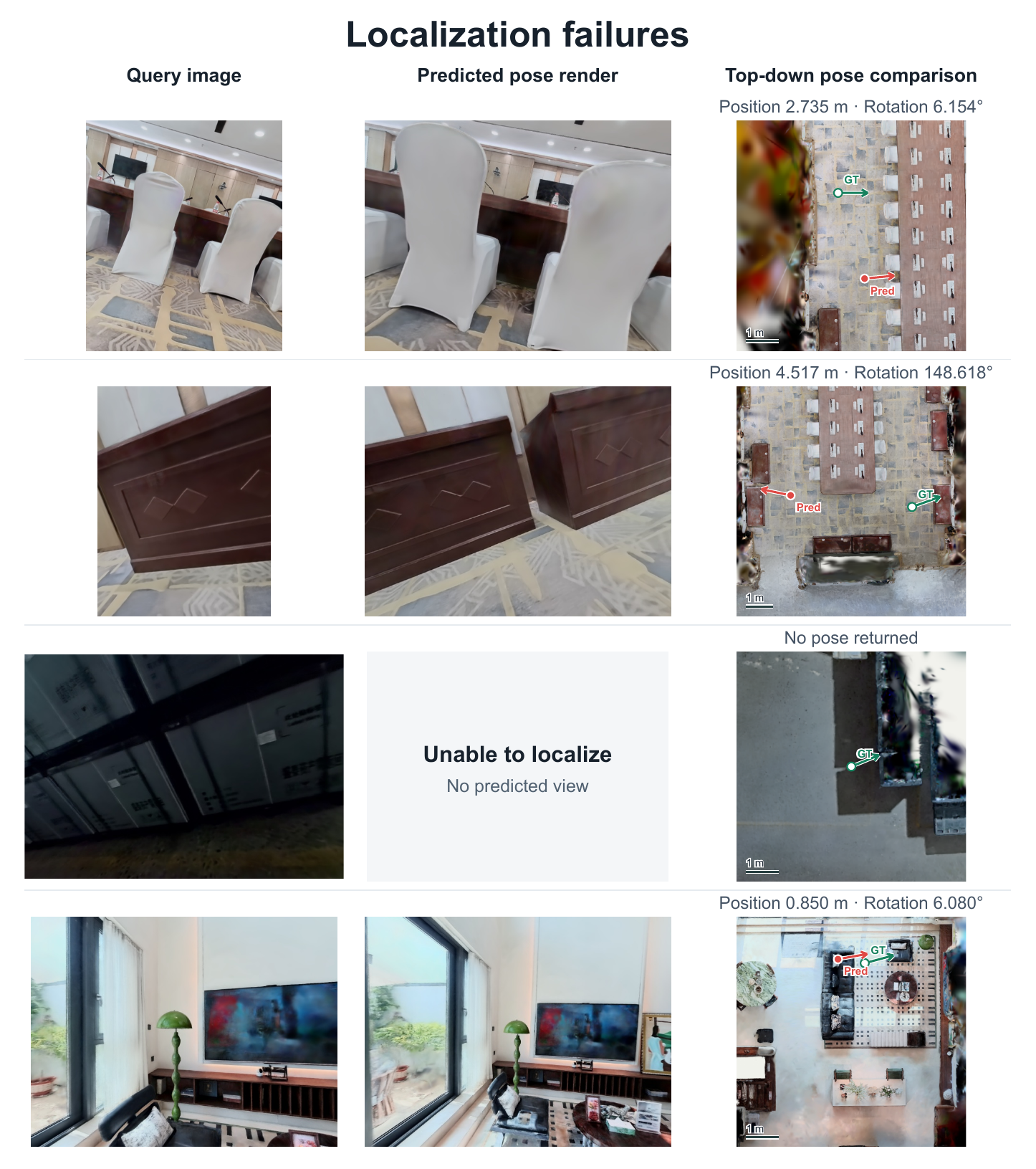}
\caption{Localization errors and a missed localization. The columns and
pose colors follow \cref{fig:localization_good}. In the third row the
model returns no pose, so the middle panel has no rendered view and the
map shows only the reference pose.}
\label{fig:localization_bad}
\end{figure}

A plausible render is not sufficient to establish position
(\cref{fig:localization_bad}). In the first row, the predicted view
looks along a row of chair backs much like the one in the query.
The orientation error is only $6.15^\circ$, yet the camera is misplaced
by 2.74\,m. The wood-paneled columns in the second row admit a similar
view from across the room: here the position error is 4.52\,m and the
rotation error is about $149^\circ$. In the dark third query, the model returns no pose even though the
memory covers that location. In the well-lit final query, it recovers
the direction of the television wall but places the camera 0.85\,m
from its reference position.
\FloatBarrier

\clearpage
\subsection{Grounding objects and sign text}
\label{sec:grounding_cases}

\begin{figure}[H]
\centering
\includegraphics[width=.78\linewidth]{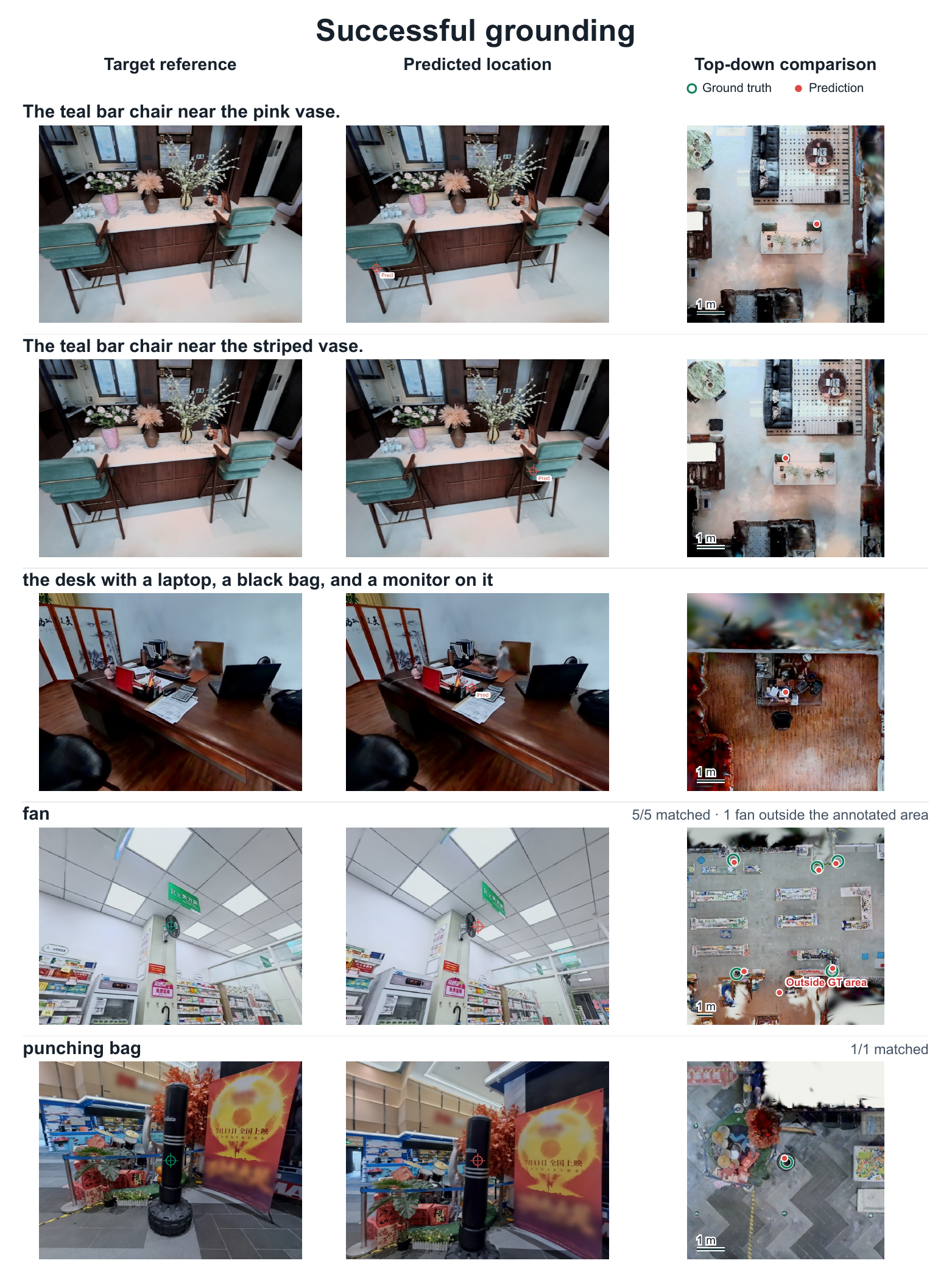}
\caption{Grounding from a category or a description. The first two image columns show the reference object and a view of a predicted location. The maps show predicted centers (red) and annotated centers where available (green). Bars indicate map scale. In all examples, a match requires the centers to be within 1\,m in 3D.}
\label{fig:grounding_good}
\end{figure}

The requests in the first two rows of \cref{fig:grounding_good} differ by a single cue:
both request a teal bar chair, but one names a pink vase and the other
a striped vase. Changing the landmark changes which chair the model
retrieves. Other rows show an object identified by items on its surface
(the desk), multiple instances of one category (the fans), and a
single instance in an unusual setting (the punching bag). Five predictions match the five annotated fans. A sixth marks a fan
outside the annotated area. These cases score object centers. Reaching a
suitable viewpoint is a separate task.

\begin{figure}[H]
\centering
\includegraphics[width=.9\linewidth]{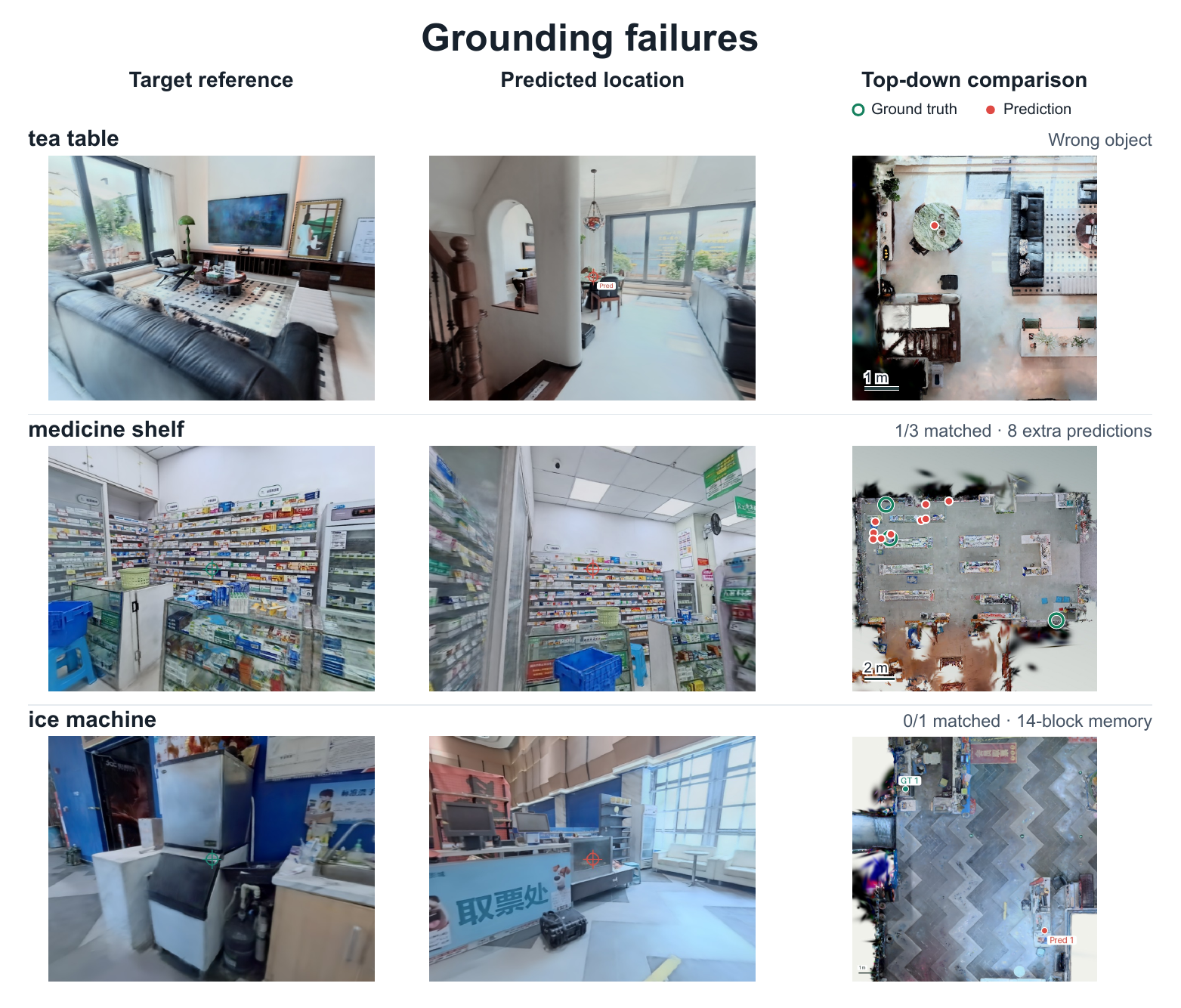}
\caption{Three kinds of object-grounding error: selecting a similar object, returning extra centers while missing annotated shelves, and placing an ice machine far from its reference location. The last example uses a 14-block memory.}
\label{fig:grounding_bad}
\end{figure}

The tea-table request in \cref{fig:grounding_bad} retrieves a round
dining table. The pharmacy example shows why both precision and recall
matter. The model returns nine medicine-shelf centers, mostly in one
part of the store. Only one center matches an annotation, leaving two
of the three annotated shelves undiscovered. For the ice machine, the
predicted center is 9.65\,m from the annotated location.

\begin{figure}[H]
\centering
\includegraphics[width=.85\linewidth]{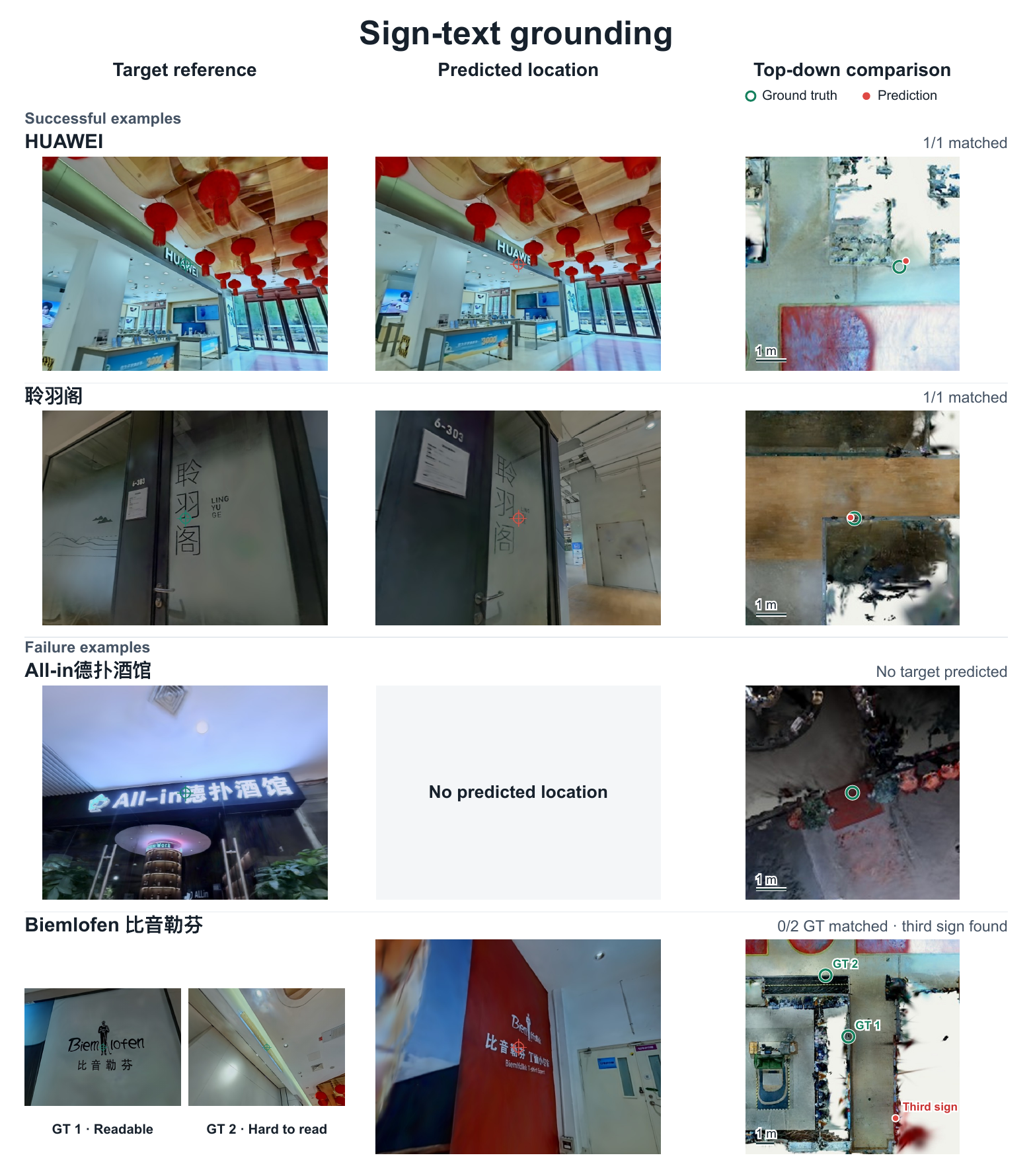}
\caption{Grounding signs from their text. Each row shows the annotated
sign or signs, the model's prediction when available, and their positions
on the map. In the last row, GT~1 and GT~2 denote distinct instances
of the requested brand.}
\label{fig:grounding_ocr}
\end{figure}

\clearpage
\paragraph{Finding a sign from its text.}
The model locates the HUAWEI sign and a vertical Chinese sign, but
returns no location for the All-in sign (\cref{fig:grounding_ocr}).
In the last example it locates another sign bearing the requested
brand. The name matches, yet both annotated instances remain unfound.

\FloatBarrier

\clearpage
\subsection{Predicting trajectories through passages}
\label{sec:trajectory_cases}

\begin{figure}[H]
\centering
\includegraphics[width=\linewidth,height=.70\textheight,keepaspectratio]{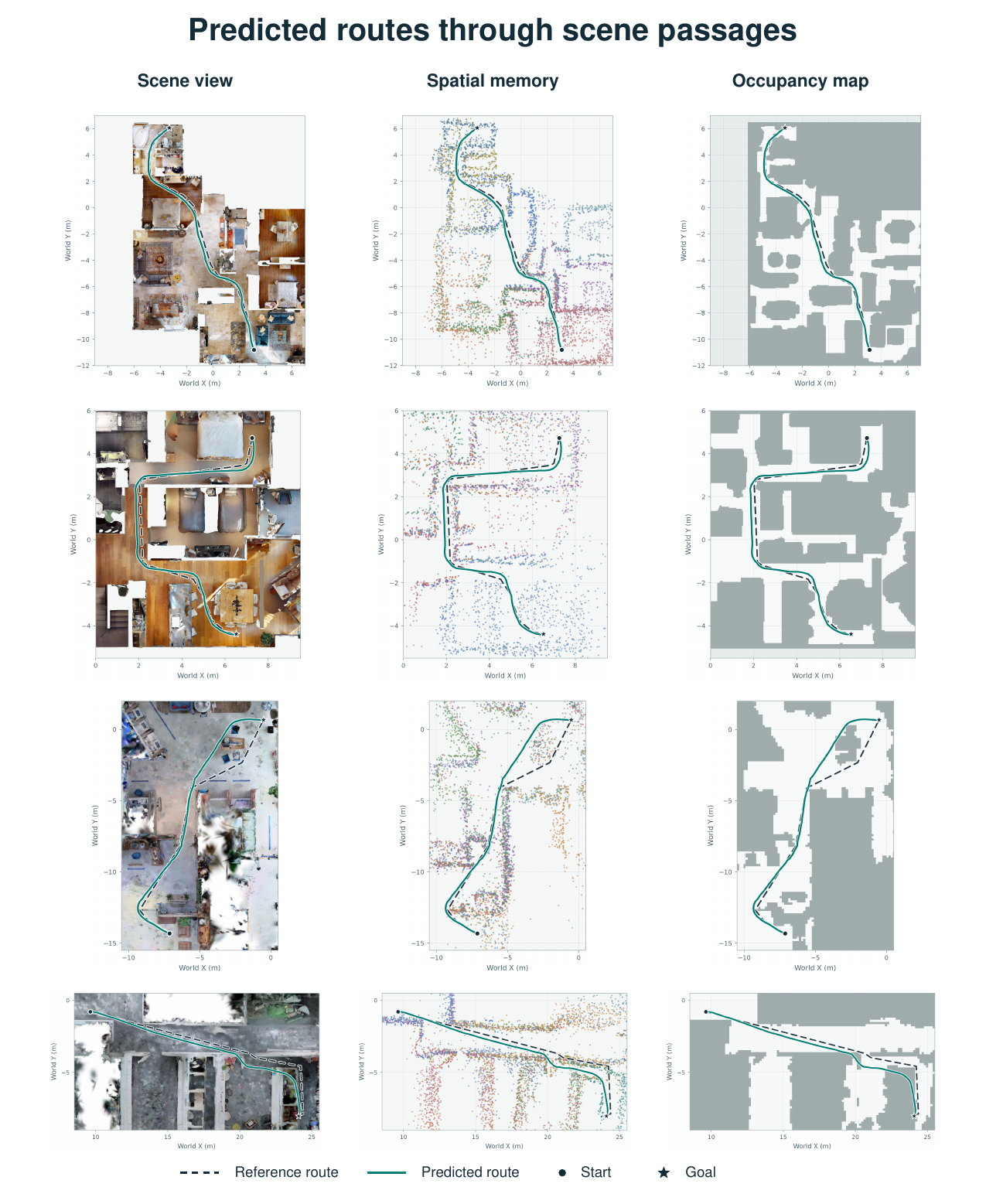}
\caption{Predicted routes through scene passages. Each row shows the
same route from above on the scene rendering, spatial memory, and
occupancy map. The solid teal line is the prediction and the dashed
line is the reference. The circle marks the start and the star the goal.
The two maps let the reader compare the predicted path through the
remembered surfaces with the free space available for travel.}
\label{fig:trajectory_good}
\end{figure}

\begin{figure}[H]
\centering
\includegraphics[width=\linewidth,height=.76\textheight,keepaspectratio]{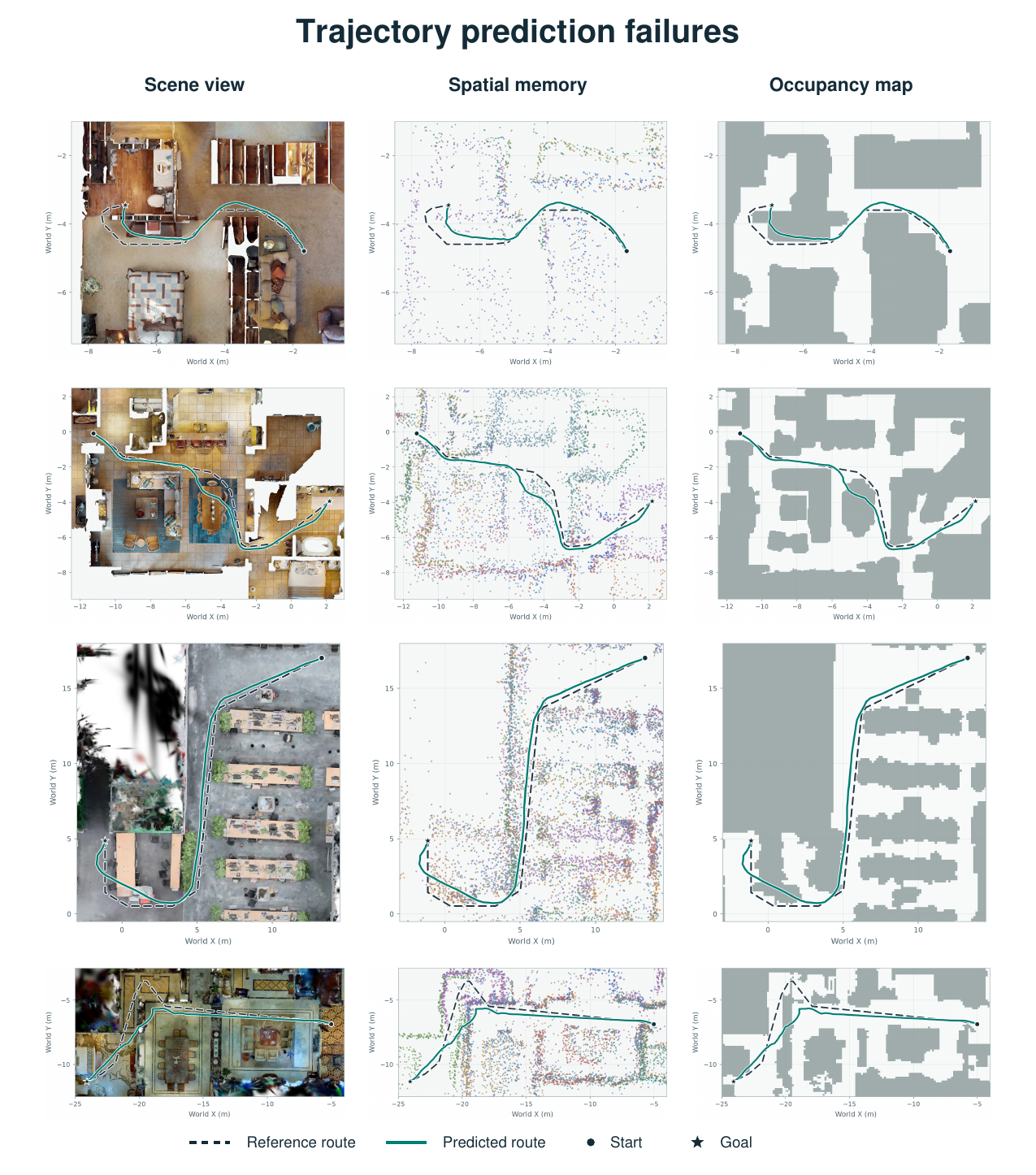}
\caption{Trajectory prediction failures before local execution. The
columns and markers follow \cref{fig:trajectory_good}. In the second
row the predicted route cuts across a gray occupied region that the
dashed reference route skirts. In the last row it takes a shallower
turn than the reference near the left end of the passage.}
\label{fig:trajectory_bad}
\end{figure}

\Cref{fig:trajectory_good,fig:trajectory_bad} compare routes for fixed
start and goal positions before the local planner acts. The flow head
refines the interior waypoints (\cref{eq:trajectory_flow}), whose
positions determine which passages the route takes and how much
clearance it leaves through a turn.

In the first two HM3D cases in \cref{fig:trajectory_good}, the model
follows the main passages through several turns. In the third, it
takes a different path around the obstacles toward the same destination. The last row shows why even a plausible route needs
a clearance check. A gap that looks like a doorway in the point cloud
is partly occupied on the map.

The predicted route can follow the overall direction of the reference
and still leave too little clearance around furniture. In
\cref{fig:trajectory_bad}, some turns cut close to tables or other
nearby obstacles and enter regions marked occupied. Other deviations
occur where the assembled point cloud gives an incomplete or unclear
view of the passage. In these cases, the endpoints alone do not distinguish a route
through the opening from one that clips its edge.
\FloatBarrier

\clearpage
\subsection{Choosing a frontier when retrieval finds no match}
\label{sec:exploration_cases}

When retrieval finds no match, the model proposes a point on the boundary
of observed space (the next frontier) and a route to it. The reference
frontier lies near the point where a path to the requested object first
leaves observed free space (Appendix~\ref{app:training_routes}). The
figures show this decision before execution. Whether the robot finds
the object depends on the views it acquires as it continues the search.

\begin{figure}[H]
\centering
\includegraphics[width=\linewidth,height=.69\textheight,keepaspectratio]{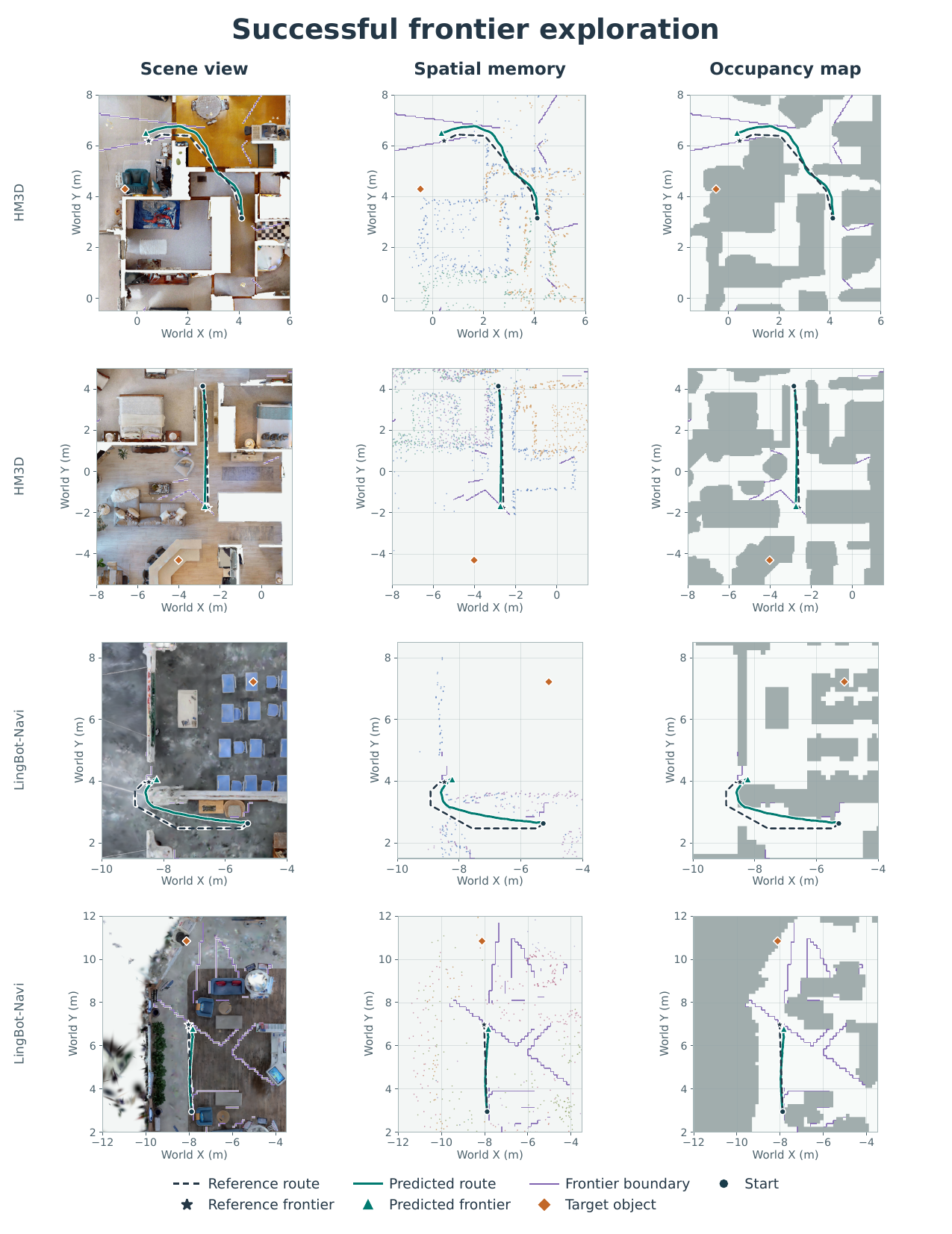}
\caption{Predicted and reference routes reach nearby frontiers.
The top two rows show HM3D scenes and the bottom two show \datasetname{}
scenes. Each row shows the same region from above as a scene rendering, a point
cloud colored by memory block, and an occupancy map with occupied cells
in gray. From the start (circle), the dashed reference leads to the star
and the solid teal prediction to the triangle. Purple segments mark
frontier boundaries. The orange diamond marks the requested object for reference.}
\label{fig:exploration_good}
\end{figure}

For the blue-topped desk at the front of the classroom (third row of
\cref{fig:exploration_good}), both routes round the end of the wall
beside the seating. They take the turn at different distances from the
wall, yet their frontier points are only 0.27\,m apart in the map plane.

\clearpage
\begin{figure}[H]
\centering
\includegraphics[width=\linewidth,height=.71\textheight,keepaspectratio]{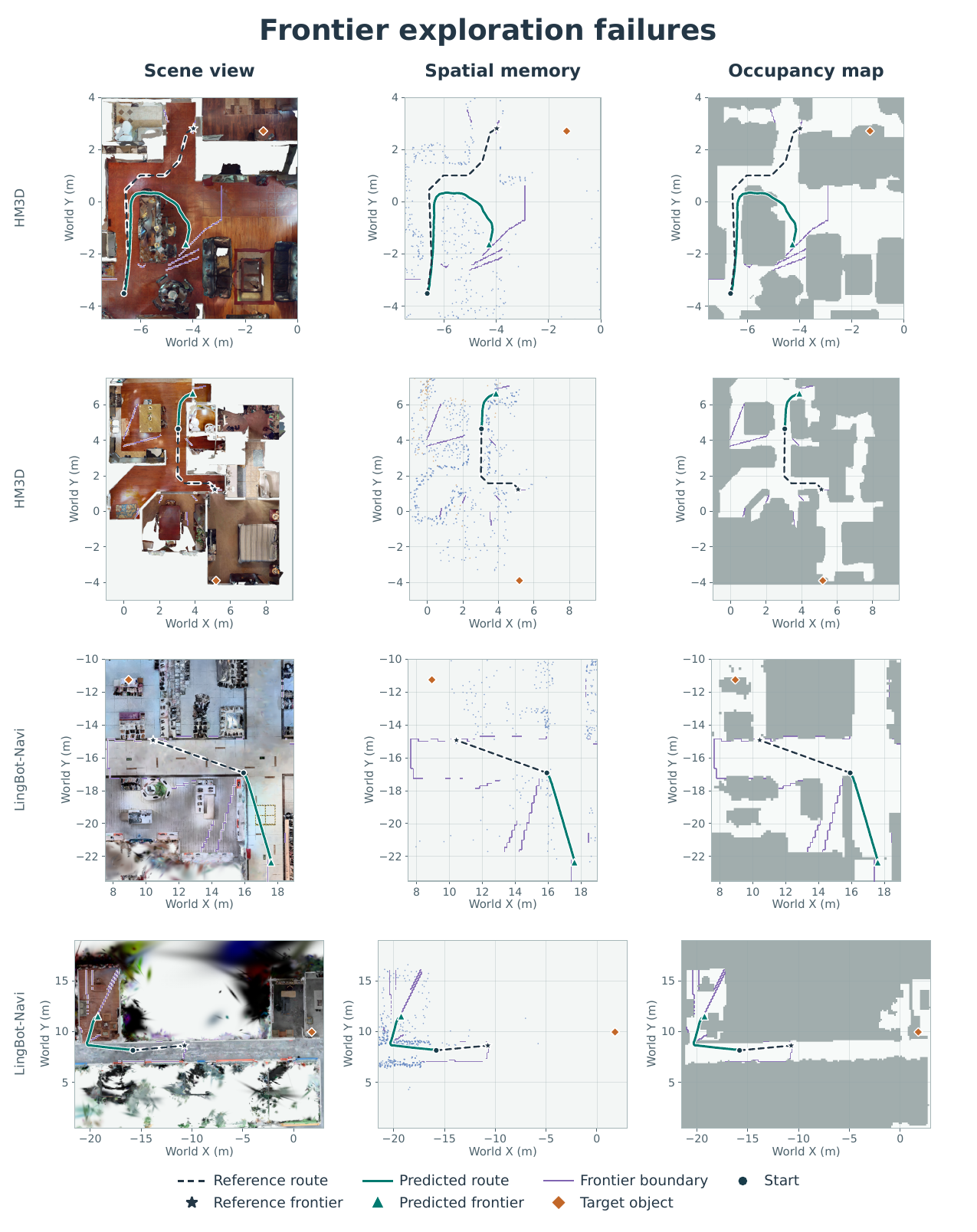}
\caption{Frontier choices that send the search away from the reference route.
As in \cref{fig:exploration_good}, the upper rows show HM3D and the lower
rows show \datasetname{}. In the first HM3D case,
the routes initially share a passage, then the prediction bends back
toward the furniture. In the second, the routes leave the start in
opposite directions along the corridor.}
\label{fig:exploration_bad}
\end{figure}

The white desk with a calculator and papers is beyond the mapped
passage in the third row of \cref{fig:exploration_bad}. The reference
route heads left from the start along the passage. The prediction
heads down and to the right. Across these four examples, the predicted
frontier is
4.44--10.33\,m from the reference in the map plane, compared with
0.17--0.33\,m in the four examples in \cref{fig:exploration_good}. Even if the robot
follows its route exactly, the frontier it chooses
determines which part of the scene it gets to observe next.
\FloatBarrier

\clearpage
\subsection{Describing the differences between candidates}
\label{sec:dialogue_cases}

\begin{figure}[H]
\centering
\includegraphics[width=.98\linewidth]{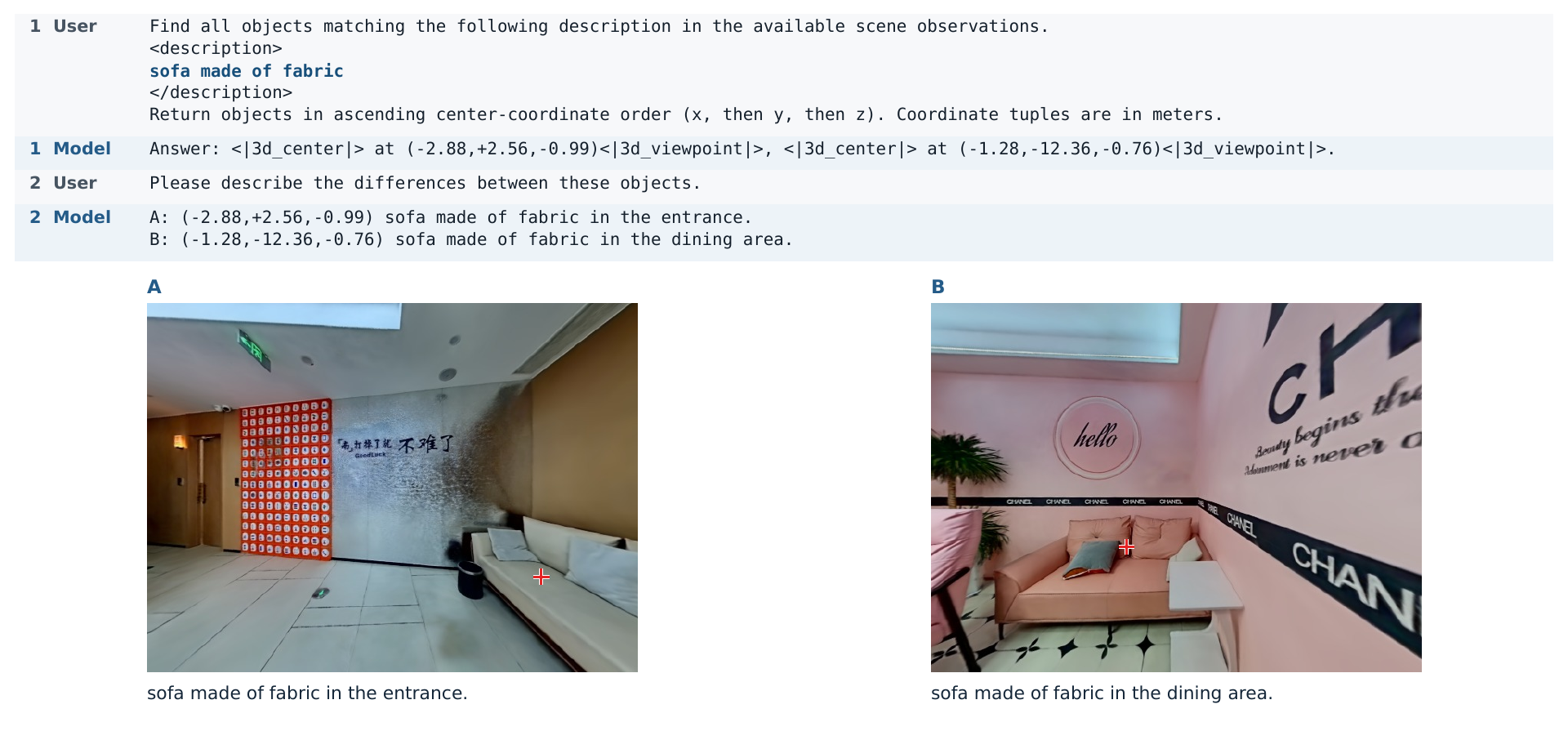}\par\medskip
\includegraphics[width=.98\linewidth]{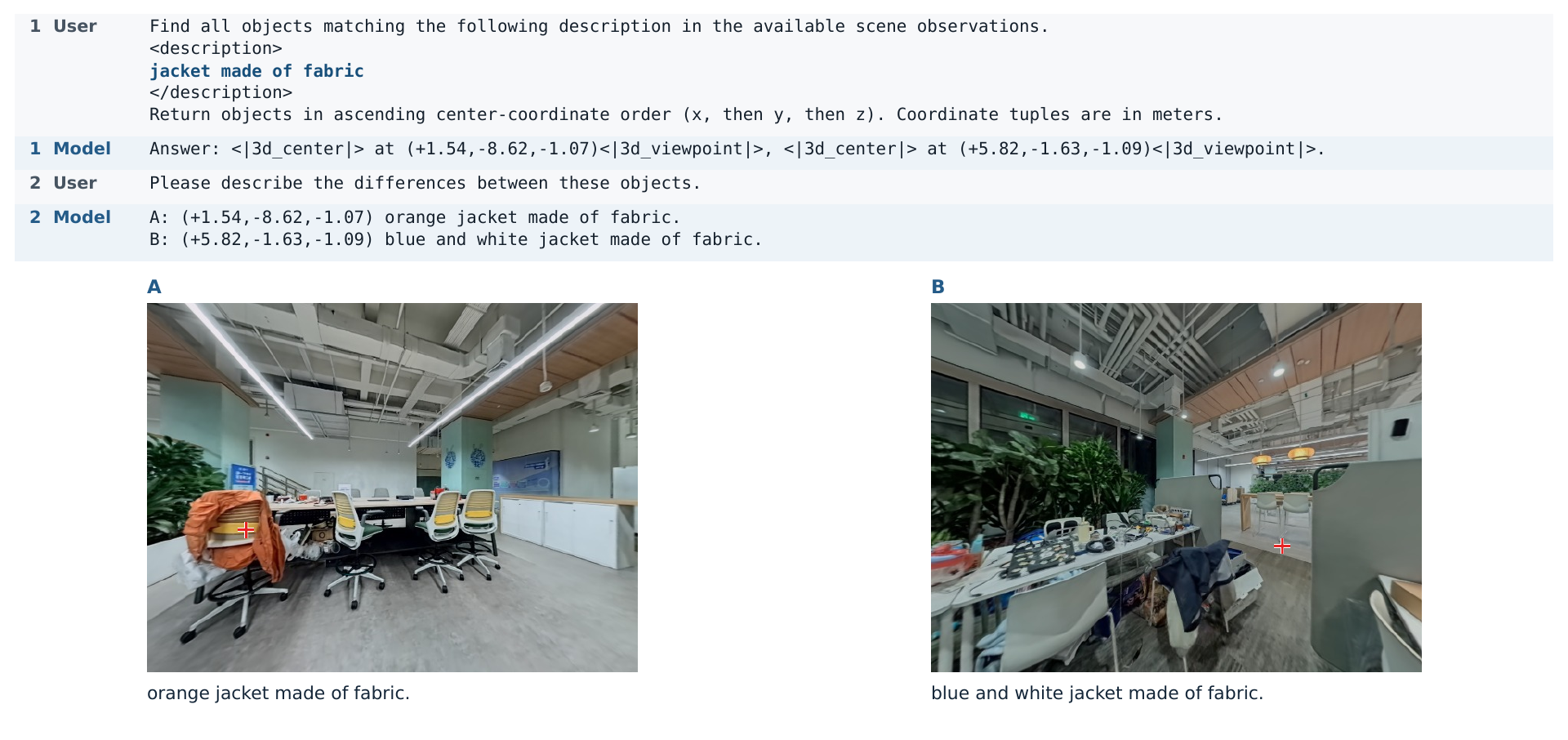}
\caption{Recorded grounding and comparison exchanges for sofas (top) and
jackets (bottom). Red crosses mark the projected centers in the candidate
views. The labels reproduce the model's answers from spatial memory.}
\label{fig:dialogue_cases}
\end{figure}

The recorded exchanges in \cref{fig:dialogue_cases} show how a robot
could compose the single clarification illustrated for the fans in
\cref{fig:clarification_example}. The first prompt retrieves the matching
objects. The second asks how they differ, and the model answers from
spatial memory. For two fabric sofas, it distinguishes one at the entrance
from the other in the dining area.
For the jackets, color is sufficient to distinguish the orange one
from the blue and white one.

\FloatBarrier

\FloatBarrier

\end{document}